\documentclass{article}

\PassOptionsToPackage{numbers, compress}{natbib}

\usepackage{microtype}
\usepackage{graphicx}
\usepackage{booktabs} 
\usepackage{makecell}
\usepackage{multirow}
\usepackage{threeparttable}
\usepackage{float}

\usepackage[final]{neurips_2024}

\usepackage[utf8]{inputenc} 
\usepackage[T1]{fontenc}    
\usepackage{hyperref}       
\usepackage{url}            
\usepackage{booktabs}       
\usepackage{amsfonts}       
\usepackage{nicefrac}       
\usepackage{microtype}      
\usepackage{xcolor}         

\usepackage{hyperref}
\usepackage{soul} 

\usepackage{amsmath}
\usepackage{amssymb}
\usepackage{mathtools}
\usepackage{amsthm}
\usepackage{bbm}
\DeclarePairedDelimiter\floor{\lfloor}{\rfloor}

\usepackage[capitalize,noabbrev]{cleveref}
\usepackage{subcaption}
\usepackage{caption}

\theoremstyle{plain}

\theoremstyle{definition}

\theoremstyle{remark}

\usepackage[textsize=tiny]{todonotes}
\usepackage{xcolor} 

\usepackage[toc,page,header]{appendix}
\usepackage{minitoc}

\usepackage[ruled,vlined,linesnumbered]{algorithm2e}
\usepackage[utf8]{inputenc}
\SetCommentSty{small}
\Crefname{algocf}{Algorithm}{Algorithms}

\title{DOFEN: Deep Oblivious Forest ENsemble}

\author{%
  Kuan-Yu Chen \\
  Sinopac Holdings\\
  \texttt{lavamore@sinopac.com} \\
  \And
  Ping-Han Chiang\thanks{Currently work at Appier Group, Inc.} \\
  Sinopac Holdings\\
  \texttt{u10000129@gmail.com} \\
  \And
  Hsin-Rung Chou \\
  Sinopac Holdings\\
  \texttt{sherry.chou@sinopac.com} \\
  \AND
  Chih-Sheng Chen \\
  Sinopac Holdings\\
  \texttt{sheng77@sinopac.com} \\
  \And
  Darby Tien-Hao Chang \\
  Sinopac Holdings\\
  National Cheng Kung University\\
  \texttt{darby@sinopac.com} \\
}

\usepackage[ruled,vlined,linesnumbered]{algorithm2e}
\usepackage[utf8]{inputenc}
\SetCommentSty{small}
\Crefname{algocf}{Algorithm}{Algorithms}

\begin{document}

\doparttoc 
\faketableofcontents 
\part{} 

\maketitle


\begin{abstract}
  Deep Neural Networks (DNNs) have revolutionized artificial intelligence, achieving impressive results on diverse data types, including images, videos, and texts.
  However, DNNs still lag behind Gradient Boosting Decision Trees (GBDT) on tabular data, a format extensively utilized across various domains.
  In this paper, we propose DOFEN, short for \textbf{D}eep \textbf{O}blivious \textbf{F}orest \textbf{EN}semble, a novel DNN architecture inspired by oblivious decision trees. DOFEN constructs relaxed oblivious decision trees (rODTs) by randomly combining conditions for each column and further enhances performance with a two-level rODT forest ensembling process. 
  By employing this approach, DOFEN achieves state-of-the-art results among DNNs and further narrows the gap between DNNs and tree-based models on the well-recognized benchmark: Tabular Benchmark \citep{grinsztajn2022tree}, which includes 73 total datasets spanning a wide array of domains. The code of DOFEN is available at:  \url{https://github.com/Sinopac-Digital-Technology-Division/DOFEN}.
\end{abstract}

\section{Introduction}
\label{introduction}

Tabular data is extensively used across various domains (e.g., finance, healthcare, government). 
For prediction tasks involving tabular data, tree-based models such as CatBoost and XGBoost \citep{prokhorenkova2018catboost,chen2016xgboost} are currently considered the state of the art \citep{grinsztajn2022tree,gorishniy2021revisiting,shwartz2022tabular,borisov2022deep,mcelfresh2023neural}.
Given the success of deep neural networks in other domains (e.g., natural language processing, computer vision), it is compelling to explore how neural networks can be leveraged to achieve improved performance on tabular data, potentially benefiting other research directions in this area (e.g., multimodal learning, self-supervised learning).

To emulate the behavior of tree-based models using deep neural networks, we observed two key points.
First, the base models for tree-based approaches (i.e., decision trees or oblivious decision trees) may exhibit crucial inductive biases that contribute to accurate predictions on tabular data. 
Second, the ensemble of base models significantly enhances predictive performance. 
For instance, bagging trees employ bootstrap sampling and bagging \citep{breiman2001random,geurts2006extremely}, while boosting trees utilize various forms of gradient boosting \citep{chen2016xgboost,ke2017lightgbm,prokhorenkova2018catboost}.

In this paper, we propose a deep neural network, named \textbf{D}eep \textbf{O}blivious \textbf{F}orest \textbf{EN}semble (DOFEN), which incorporates the two key observations mentioned earlier. 
First, we select the oblivious decision tree (ODT) \citep{kohavi1994bottom,lou2017bdt} as the base model, as it represents a decision table and is easier to model (\cref{sec:background}).
For example, the following set shows the columns with their decisive conditions from a trained ODT:
\[\{ \text{Col}_\text{B} < 7.8, \text{Col}_\text{A} > 5, \text{Col}_\text{C} = \mathrm{cat}\}\]
A key characteristic of an ODT is its disregard for the decision-making sequence, allowing us to focus on condition selection. 
Consequently, the first step in DOFEN is to randomly select column conditions from a set of generated conditions for each column.
This random selection process is repeated multiple times to derive numerous combinations of conditions.
This step primarily aims to substitute the learning process of an ODT by iterating over as many condition combinations as possible.
Next, we randomly select several condition combinations, termed relaxed oblivious decision trees (rODTs) in the following context, and assemble them into an rODT forest. 
This step is also repeated multiple times, and the final predictions are made by bagging through the predictions of these rODT forests.

\begin{figure}[t]
    \centering
    \begin{subfigure}{.25\textwidth}
        \centering
        \includegraphics[width=\linewidth]{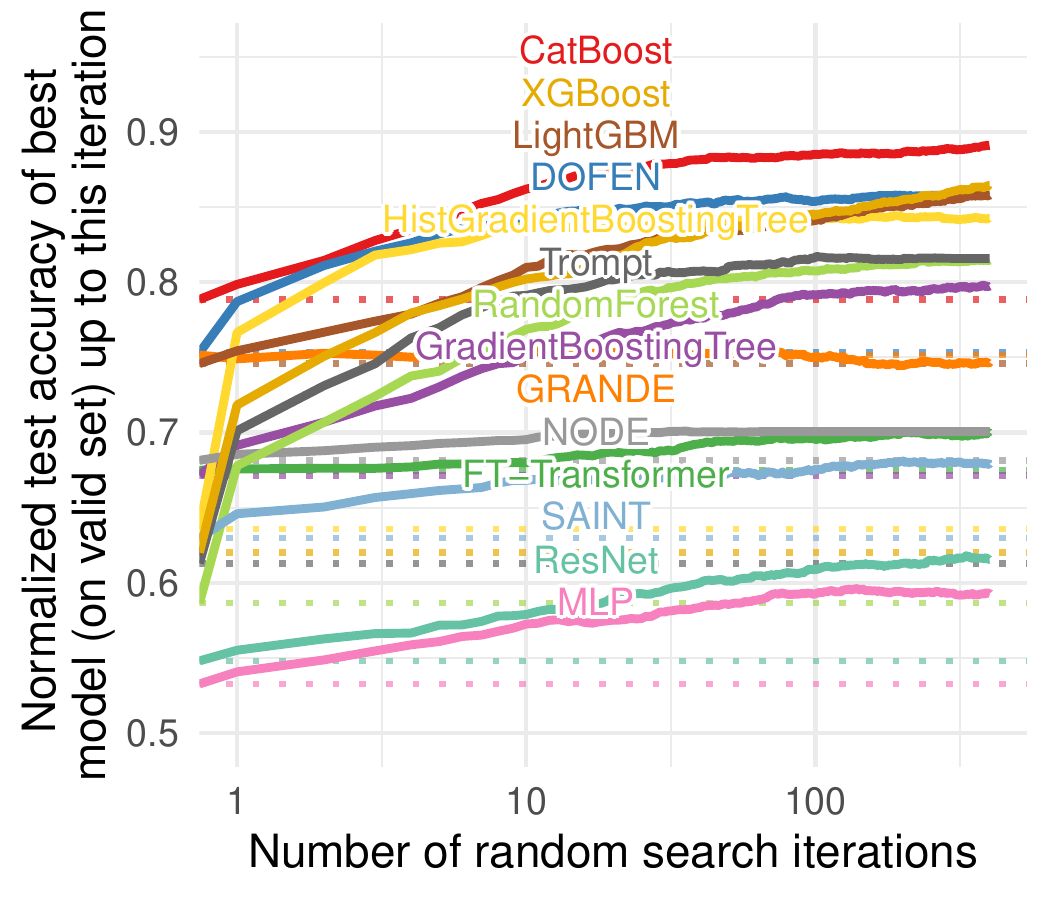}
        \caption{Medium, Classification.}
        \label{fig:overall-eval-a}
    \end{subfigure}%
    \begin{subfigure}{.24\textwidth}
        \centering
        \includegraphics[width=\linewidth]{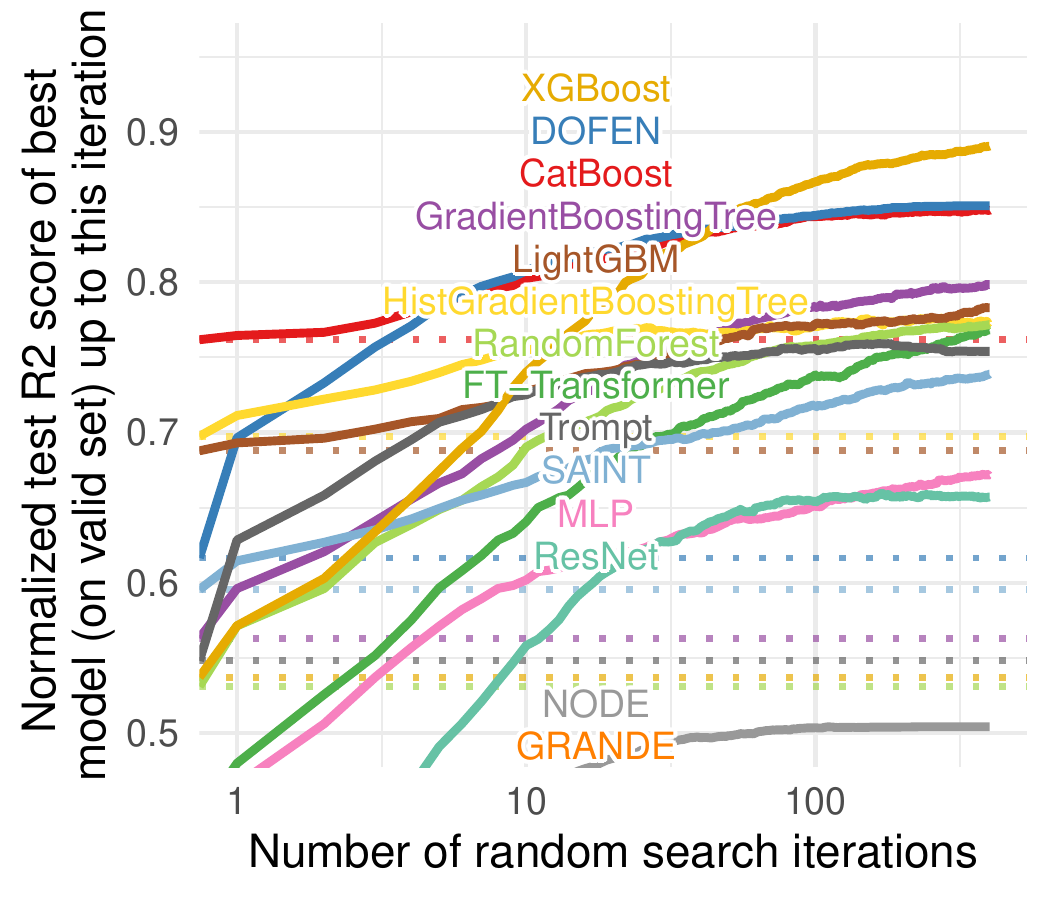}
        \caption{Medium, Regression.}
        \label{fig:overall-eval-b}
    \end{subfigure}
    \begin{subfigure}{.24\textwidth}
        \centering
        \includegraphics[width=\linewidth]{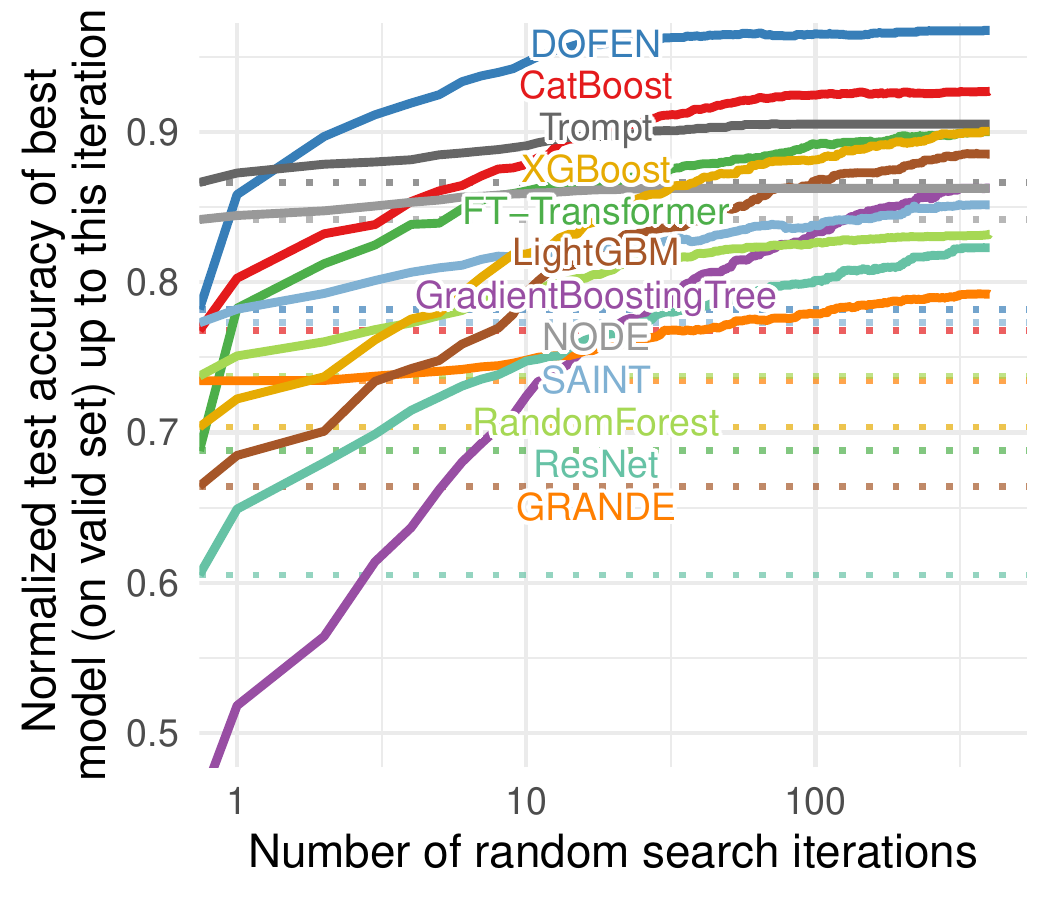}
        \caption{Large, Classification.}
        \label{fig:overall-eval-c}
    \end{subfigure}
    \begin{subfigure}{.25\textwidth}
        \centering
        \includegraphics[width=\linewidth]{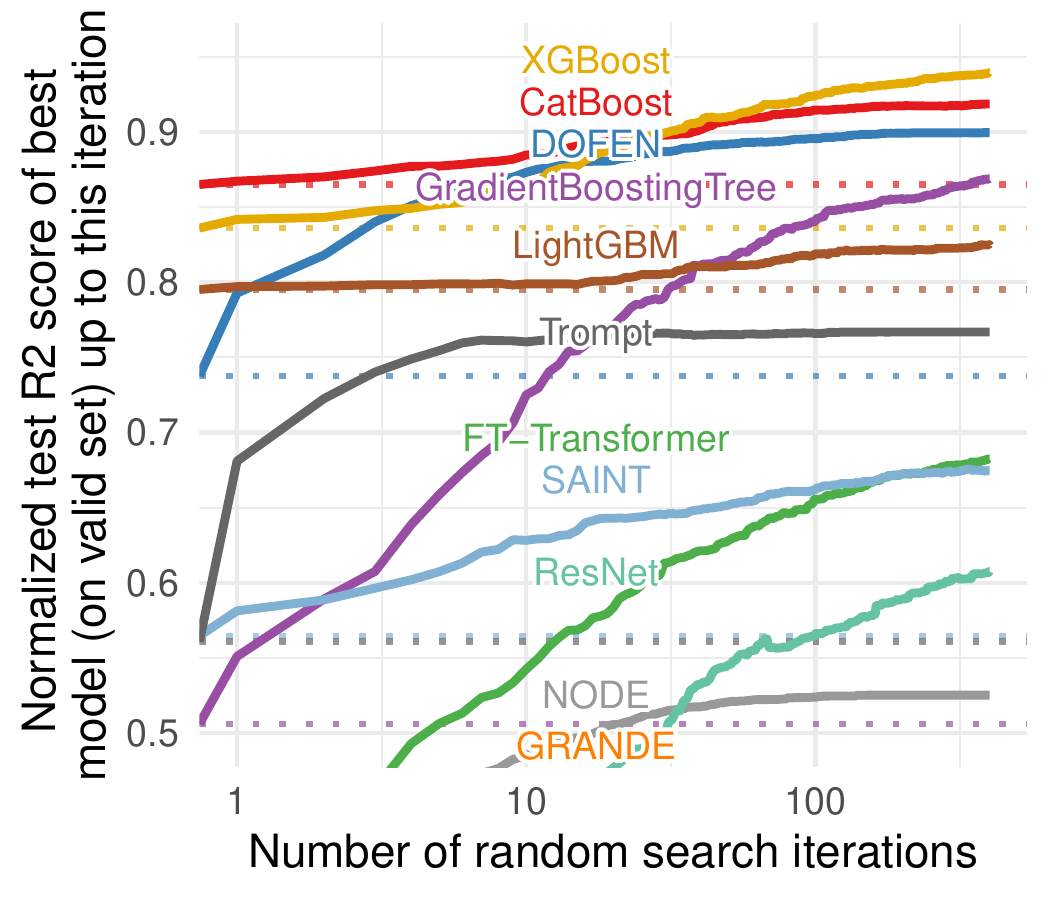}
        \caption{Large, Regression.}
        \label{fig:overall-eval-d}
    \end{subfigure}
    \caption{Evaluation results on the Tabular Benchmark. The model names are sorted by their performances at the end of the random search of hyperparameters. The result are averaged over various datasets included in each benchmark respectively, detailed number of datasets of each benchmark is provided in \cref{sec:benchmark-dataset-count}}
    \label{fig:overall-eval}
\end{figure}

Our main contributions are as follows:
\begin{enumerate}
    \item\textbf{Innovative Neural Network Architecture.} 
    DOFEN introduces a novel deep neural network architecture designed to address tabular data problems (\cref{sec:main-idea}).
    To harness the strengths of tree-based models in this domain, DOFEN integrates oblivious decision trees into the network architecture through a random condition selection process, leading to the formation of relaxed oblivious decision trees (rODTs).
    To further exploit the power of ensembling, DOFEN aggregates rODTs from the previously generated rODT pool to construct an rODT forest, and makes predictions by bagging the outputs from these rODT forests. 
    \item\textbf{State-of-the-Art Performance.} 
    To comprehensively and objectively evaluate DOFEN, we selected the recent and well-recognized Tabular Benchmark \citep{grinsztajn2022tree}. 
    This benchmark addresses the common issue of inconsistent dataset selection in deep learning research on tabular data by incorporating a variety of regression and classification datasets with standardized feature processing.
    The experimental results show that DOFEN outperforms other neural network models and competes closely with GBDTs on the Tabular Benchmark, highlighting its versatility across different tasks, as demonstrated in \cref{fig:overall-eval}.
    Additionally, we conducted detailed analyses on DOFEN's unique features, providing deeper insights into its functionalities.
    Both results can be found in \cref{sec:exp}.
\end{enumerate}

\section{Background: Oblivious Decision Tree}
\label{sec:background}
Let $\mathcal{D}=\{(\Vec{x}_i,y_i)\}_{i=1}^N$ be a dataset of size $N$ and $\mathcal{F}=\{f_j\}_{j=1}^{N_{col}}$ be its feature set of $N_{col}$ features. Namely, $\Vec{x}_i = (x_{i1}, \dots, x_{iN_{col}})$ and $y_i$ are the feature vector and label, respectively, of the $i$-th sample, where $x_{ij}$ is the value corresponding to feature $f_j$. 
Note that $y_i \in \mathbb{R}$ for regression problems and $y_i \in \{0, \dots, g\}$ for classification problems with $g$ being the number of classes.

In a trained ODT, each layer can be represented by three components: a feature of $\Vec{x}$, a threshold for that feature, and a condition based on whether a value is greater than or less than the threshold. In the following context, an ODT with a depth of $d$ consists of $d$ instances of $\Vec{z}$, $\Vec{v}$, and $\Vec{c}$, respectively, corresponding to the three components.
Let $\Vec{z} = (z_1, \dots, z_d)$, where $z_i \in \Vec{x}$ represents a feature of $\Vec{x}$.
Let $\Vec{v} = (v_1, \dots, v_d)$, where $v_i \in \mathrm{dom}(z_i)$ is a threshold for $z_i$.
Finally, let $\Vec{c} = (c_1, \dots, c_d)$, where $c_i \in \{>,<\}$ is paired with $z_i$ and $x_i$ to make decisions.
Note that we have simplified the notation by assuming only numerical features, and the training algorithm can be found in the original work \citep{kohavi1994bottom,lou2017bdt}.

ODT distinguishes itself from conventional decision tree algorithms \citep{quinlan1986induction} by restricting each layer to use only one feature.
This uniformity simplifies decision-making and improves computational efficiency through vectorized operations.
While this reduces model capacity, recent studies have shown that ensembling ODTs can improve performance \citep{prokhorenkova2018catboost,popov2019neural}.
In \cref{sec:main-idea}, we will futher discuss how we leverage ODT to design a network architecture and introduce a novel ensemble strategy to boost its performance.
\section{DOFEN: Deep Oblivious Forest Ensemble}
\label{sec:main-idea}

In this section, we first explain how DOFEN integrates ODT into the network architecture in \cref{sec:rodt-construction} followed by introducing a two-level rODT ensemble in \cref{sec:rODT-ensemble}. 
For clarity, we simplify naive sub-networks—comprising only basic neural layers such as linear layers, layer normalization, and dropout—into symbols (i.e. $\Delta_1$, $\Delta_2$, and $\Delta_3$) in the following figures and equations.
Detailed configurations of these sub-networks are provided in \cref{sec:model-layers}.

\subsection{Relaxed ODT Construction}
\label{sec:rodt-construction}

This section goes through how DOFEN transforms a raw input into soft conditions and constructs multiple relaxed ODTs by randomly combining these conditions.

Recall that a trained ODT consists of $\Vec{z}$, $\Vec{v}$ and $\Vec{c}$. Since the ODT learning algorithm —selecting features and thresholds by minimizing loss  (e.g. gini impurity or mean squared error)— is non-differentiable, deriving $\Vec{v}$ and $\Vec{c}$ using gradient-based optimization is challenging.
We bypass the non-differentiable issue by: (1) randomly selecting $\Vec{z}$ for an ODT and, (2) replacing $\Vec{v}$ and $\Vec{c}$ with a neural network, which gives a soft score (i.e. condition) measuring how a sample adheres to a decision rule.
In practice, we first generate multiple soft conditions for each column and then randomly combine several conditions to form an ODT. We called an ODT constructed by this soften procedure as a \textbf{relaxed ODT (rODT)}, and the two steps as \textbf{Condition Generation} and \textbf{Relaxed ODT Construction}, respectively. The detailed process are provided as follows.

\begin{figure}[t]
        \centering
        \includegraphics[width=0.9\linewidth]
        {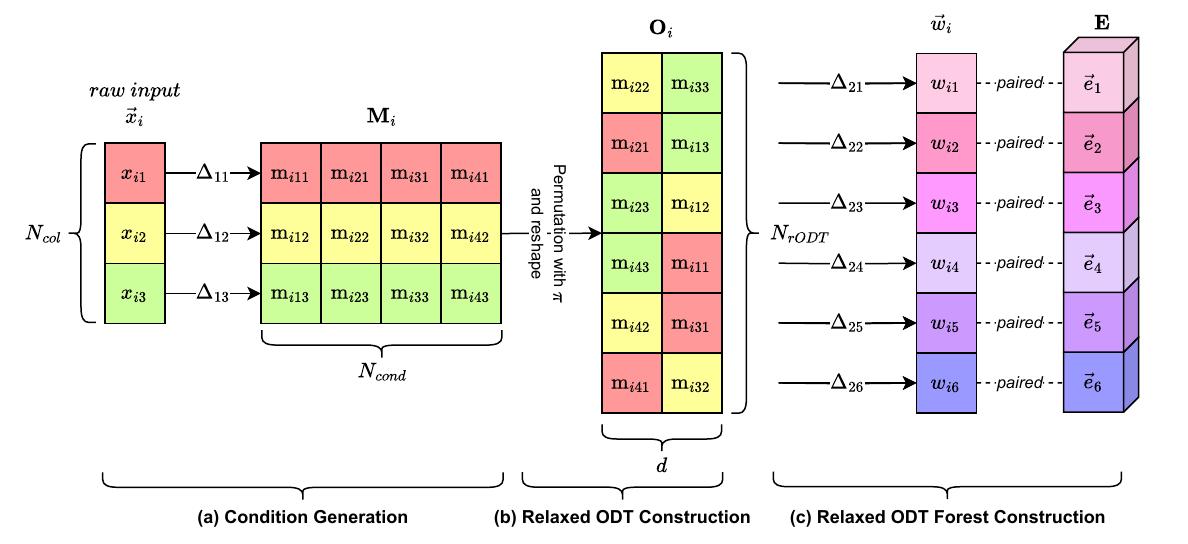}
        \caption{(a) Condition Generation: For each column, $N_\text{cond}$ conditions are generated through an individual sub-network $\Delta_1$. The aggregate of the conditions of all columns is denoted by the matrix $\mathbf{M}_i$. 
        (b) Relaxed ODT Construction: Perform permutation on $\mathbf{M}_i$ with a bijective function $\pi$ and reshape $\mathbf{M}_i$ into $\mathbf{O}_i$, a matrix representing $N_\text{rODT}$ rODTs with depth $d$.
        (c) Relaxed ODT Forest Construction: To compute the weights $w_{ij}$, an individual sub-networks $\Delta_2$ is applied to each rODT.
        In addition, each $w_{ij}$ is paired with a learnable embedding vector $\Vec{e}_j$.
        The aggregate of all weights and their corresponding embedding vectors are denoted as $\Vec{w}_i$ and $\mathbf{E}$, respectively.}
        \vspace{-10mm}
        \label{fig:dofen-part-1}

    \begin{subfigure}{0\textwidth}
        \phantomsubcaption \label{fig:condition-generation}
    \end{subfigure}

    \begin{subfigure}{0\textwidth}
        \phantomsubcaption \label{fig:rodt-construction}
    \end{subfigure}

    \begin{subfigure}{0\textwidth}
        \phantomsubcaption \label{fig:forest-construction}
    \end{subfigure}
\end{figure}

\textbf{Condition Generation.}
This process transforms a raw input $\Vec{x}_i$ into soft conditions $\mathbf{M}_i$, as shown in \cref{eq:condition-generation}. 
For each feature $x_{ij}$ of the raw input $\Vec{x}_i$, where $j \in \{1,\dots,N_\text{col}\}$, $N_\text{cond}$ conditions are generated by a sub-network $\Delta_{1j}$.
The aggregated conditions are represented by the matrix $\mathbf{M}_i$. 
This design mirrors the original ODT, where each condition involves only one feature.
As illustrated in \cref{fig:dofen-part-1}\subref{fig:condition-generation}, three instances of $\Delta_1$ generate four conditions per feature, forming a $3 \times 4$ matrix.

\begin{equation}
    \mathbf{M}_i = 
    \begin{bmatrix} 
        m_{i11}  & \dots & m_{i1N_\text{col}} \\
        \vdots & \ddots & \vdots \\
        m_{i{N_\text{cond}1}} & \dots & m_{iN_\text{cond}N_\text{col}} 
    \end{bmatrix} \in \mathbb{R}^{N_\text{cond} \times N_\text{col}}, (m_{ij1},\dots,m_{ijN_\text{cond}}) = \Delta_{1j}(x_{ij})
    \label{eq:condition-generation}
\end{equation}

\textbf{Relaxed ODT Construction.}
To build an rODT with depth $d$, we randomly select $d$ elements from the matrix $\mathbf{M}_i$.
In practice, to construct multiple rODTs, $\mathbf{M}_i$ is shuffled and reshaped into a matrix $\mathbf{O}_i$ with dimensions $N_\text{rODT} \times d$, as shown in \cref{eq:rodt-construction}. Here, we use $\pi$ to represent a bijective function that maps the index of each element in $\mathbf{M}_i$ to a unique position in $\mathbf{O}_i$ (i.e. permutation). The whole process is also illustrated in \cref{fig:dofen-part-1}\subref{fig:rodt-construction}.

In practice, we ensure $N_\text{rODT} = N_\text{cond}N_\text{col}/d$ and $N_\text{cond} = md$ hold, where $m$ and $d$ are hyper-parameters to define model capacity, in order to make the reshaping possible.
To ensure the stability during training process, this random combination is done only once during model construction.
Note that each row in $\mathbf{O}_i$ represents an rODT, which is crucial for subsequent operations.

\begin{gather}
    \mathbf{O}_i = 
    \begin{bmatrix} 
        o_{i11}  & \dots & o_{i1d} \\
        \vdots & \ddots & \vdots \\
        o_{i{N_\text{rODT}1}} & \dots & o_{iN_\text{rODT}d} 
    \end{bmatrix} \in \mathbb{R}^{N_\text{rODT} \times d}, \nonumber \\ \nonumber \\
    \left\{o_{ijk} \mid j = \left\lceil \frac{\pi(n)}{d} \right\rceil, k = \pi(n) \bmod d, n = u \times N_{\text{col}} + v \right\}
= \left\{m_{iuv}\right\} \subset \mathbf{M}_i, \nonumber \\
\text{where}\; 1 \leq u \leq N_{\text{cond}},\; 1 \leq v \leq N_{\text{col}}
    \label{eq:rodt-construction}
\end{gather}

\subsection{Two-level Relaxed ODT Ensemble}
\label{sec:rODT-ensemble}

This section integrates rODTs to construct rODT forests, then applies bagging to ensemble the predictions of the rODT forests for the final output.

\textbf{First level: Relaxed ODT Forest Construction.}
An rODT forest is constructed by aggregating randomly selected rODTs in $\mathbf{O}_i$.
Specifically, $N_\text{estimator}$ rODTs are chosen to form an rODT forest, where $N_\text{estimator}<N_\text{rODT}$.
To aggregate them, we first use $\Delta_{2j}$, where $j \in \{1,\dots,N_\text{rODT}\}$, to compute the weight $w_i$ for each rODT, as shown in \cref{eq:derive-rODT-weight}.
Additionally, each rODT is paired with an embedding $\Vec{e}_j$, as shown in \cref{eq:rodt-embedding}.
The aggregate of all weights and their corresponding embedding vectors are denoted as $\Vec{w}_i$ and $\mathbf{E}$, respectively.
This procedure is illustrated in \cref{fig:dofen-part-1}\subref{fig:forest-construction}.

\begin{equation}
    \Vec{w}_i = 
    \left(
    \begin{gathered}
        \Delta_{21}((o_{i11},\dots,o_{i1d})) \\
        \vdots \\
        \Delta_{2N_\text{rODT}}((o_{iN_\text{rODT}1},\dots,o_{iN_\text{rODT}d})) \\
    \end{gathered} 
    \right) 
    =(w_{i1},\dots,w_{iN_\text{rODT}})
    \in \mathbb{R}^{N_\text{rODT}}
\label{eq:derive-rODT-weight}
\end{equation}

\begin{equation}
    \mathbf{E} = 
    \begin{bmatrix} 
        \Vec{e}_1 \\
        \vdots \\
        \Vec{e}_{N_\text{rODT}}
    \end{bmatrix} \in \mathbb{R}^{N_\text{rODT} \times N_\text{hidden}},
    \text{where}\; \Vec{e}_j \in \mathbb{R}^{N_\text{hidden}}, j = (1,2,\dots,N_\text{rODT})
    \label{eq:rodt-embedding}
\end{equation}

\begin{figure}[t]
        \centering
        \vspace{-2mm}
        \includegraphics[width=0.9\linewidth]
        {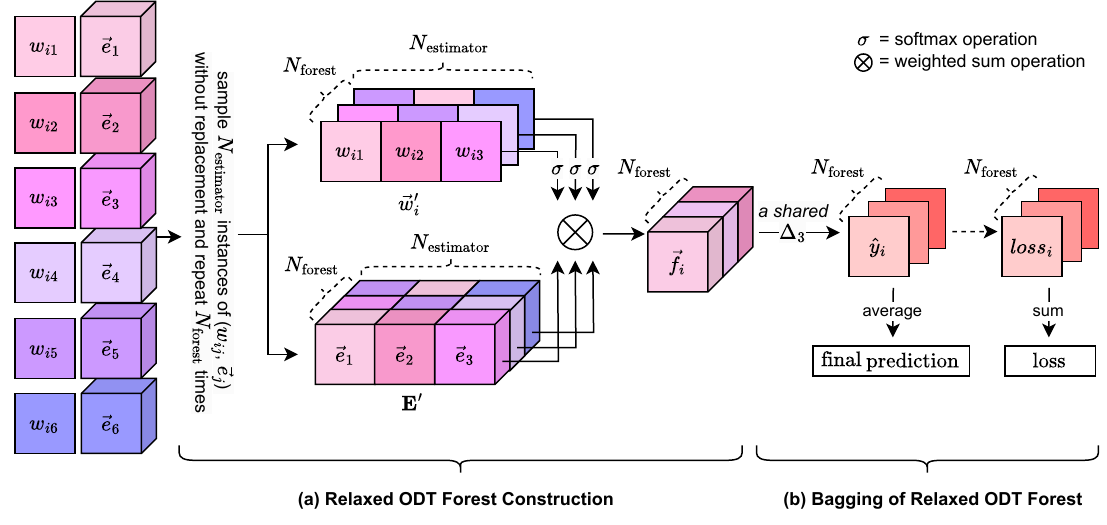}
        \caption{(a) Relaxed ODT Forest Construction: First, $N_\text{estimator}$ pairs of $(w_{ij}, \Vec{e}_j)$ are randomly sampled to form $\Vec{w}_i'$ and $\mathbf{E}'$.
        Secondly, $\Vec{w}_i'$ is transformed through a softmax function, and is used for computing the weighted sum of $\mathbf{E}'$ to form forest embedding $\Vec{f}_i$.
        (b) Baggging of Relaxed ODT Forest: a shared-weight sub-network $\Delta_3$ is employed to make a prediction $\hat{y}_i$ for each embedding.
        The final prediction is the average of all $\hat{y}_i$ values, and the total loss is the sum of their individual losses.}
        \label{fig:dofen-part-2}

        \begin{subfigure}{0\textwidth}
            \phantomsubcaption \label{fig:forest-construction-2}
        \end{subfigure}
    
        \begin{subfigure}{0\textwidth}
            \phantomsubcaption \label{fig:forest-ensemble}
        \end{subfigure}
\end{figure}

To further construct an rODT forest, $N_\text{estimator}$ of paired weights and embeddings are sampled from $\Vec{w}_i$ and $\mathbf{E}$. This process is graphically represented in \cref{fig:dofen-part-2}\subref{fig:forest-construction-2} and described in line 3 to 7 of the pseudo-code for the two-level ensemble (\cref{algo:two-level-rodt-ensemble}).
The weights are processed through a softmax function and the weighted sum of embeddings forms the embedding vector $\Vec{f}_i$ for an rODT forest.
The magnitude of these softmaxed weights indicate the importance of the selected rODTs for making predictions.
Noted that this process is repeated $N_\text{forest}$ times to form $N_\text{forest}$ instances of rODT forests.

The weighted mechanism can be extended to a multi-head version.
For simplicity, we leave the details of multi-head mechanism in \cref{sec:multihead-dofen}.

\begin{algorithm}
\caption{Two-level Relaxed ODT Ensemble}
\label{algo:two-level-rodt-ensemble}
\SetAlgoLined
\KwIn{$\Vec{w}_i$, $\mathbf{E}$, $N_\text{forest}$, $y_i$, $\mathcal{L}$}
\KwOut{$\hat{y}_i$, $loss_i$}

\textbf{Initialize} $\hat{y}_i, loss_i \gets 0, 0$\;
\For{$r \gets 1$ \KwTo $N_\text{forest}$}{
    $\Vec{w}_i', \mathbf{E}' \xleftarrow{\text{sample without replacement}} \Vec{w}_i, \mathbf{E}$\ \tcc*[r]{$N_\text{estimator}$ paired elements are sampled.}
    $\Vec{w}_i' \in \mathbb{R}^{N_\text{estimator}}$\;
    $\mathbf{E}' \in \mathbb{R}^{N_\text{estimator} \times N_\text{hidden}}$\;
    $\Vec{f}_i \gets \mathlarger{\sum}^{N_\text{estimator}}\text{softmax}(\Vec{w}_i')\circ\mathbf{E}'$ \tcc*[r]{Element-wise multiplication with broadcast.}
    $\Vec{f}_i \in \mathbb{R}^{N_\text{hidden}}$ \tcc*[r]{$\Vec{f}_i$ represents an rODT forest embedding.}
    $\hat{y}_i' \gets \Delta_3(\Vec{f}_i)$ \tcc*[r]{Give prediction with a shared $\Delta_3$.}
    $loss_i \gets loss_i + \mathcal{L}(\hat{y}_i', y_i)$ \tcc*[r]{Calculate loss with loss function $\mathcal{L}$ and aggregate.}
    $\hat{y}_i \gets \hat{y}_i + \hat{y}_i'$ \tcc*[r]{Aggregate each forest's prediction.}
}

$\hat{y}_i \gets \hat{y}_i/N_\text{forest}$\;
\Return $(\hat{y}_i, loss_i)$\;
\end{algorithm}

\textbf{Second Level: Bagging of Relaxed ODT Forest.}
To make a prediction, DOFEN applies a shared sub-network $\Delta_3$ to the embedding of each rODT forest to make individual predictions.
The predictions are then averaged for a bagging ensemble.
The process is detailed in line 1, 8, 10, and 12 in \cref{algo:two-level-rodt-ensemble} and is illustrated in \cref{fig:dofen-part-2}\subref{fig:forest-ensemble}.
Notice that the output $\hat{y}_i$ is a scalar for regression tasks and a vector for classification tasks.

During training, DOFEN updates the model parameters by aggregating the loss from each prediction, as shown in line 9 in \cref{algo:two-level-rodt-ensemble}.
The loss function $\mathcal{L}$ is cross-entropy for classification tasks and mean squared error for regression tasks.

This method of bagging over rODT forests promotes the creation of diverse rODT forests during training.
Although the randomization may seem chaotic, experimental results demonstrate that inference variance remains low even with a small $N_\text{forest}$.
Moreover, the randomization helps reduce overfitting.
Further details on these observations are provided in \cref{sec:ablation,sec:impact-of-random}.
Lastly, it is worth noting that the randomness of this process is fixed during inference stage for model to output deterministic result.
\section{Experiments}
\label{sec:exp}

This section presents a comprehensive analysis to demonstrate the effectiveness and the functionality of DOFEN. The experiments are designed to answer the following research questions (\textbf{RQ}):

\textbf{RQ1:}
How well does DOFEN perform compared to baseline and SOTA models? (\cref{sec:eval})\\
\textbf{RQ2:}
Which part of the model design contributes the most to DOFEN's performance? (\cref{sec:ablation})\\
\textbf{RQ3:}
Is the decision-making process of DOFEN interpretable? (\cref{sec:interpretability})\\
\textbf{RQ4:}
Does randomization processes involved in DOFEN cause instability issues? (\cref{sec:impact-of-random})

We also conduct additional analyses of DOFEN, including computational efficiency, scalability, and the impact of individual rODT weights. The results are presented in \cref{sec:computation-analysis,sec:long-inference-time-dofen,sec:training-time-dofen} , \cref{sec:scalability-analysis} and \cref{sec:analysis-weight,sec:more-analysis-prune}, respectively.

\subsection{Experimental Settings}
\label{sec:exp-setup}

\textbf{Datasets.} 
We strictly follow the protocols of the Tabular Benchmark as detailed in its official implementation\footnote{\href{https://github.com/LeoGrin/tabular-benchmark}{https://github.com/LeoGrin/tabular-benchmark}}. 
This includes dataset splits, preprocessing methods, hyperparameter search guidelines, and evaluation metrics. For full details, please refer to the original paper \cite{grinsztajn2022tree}. 
The Tabular Benchmark categorized datasets into classification and regression, with features being either exclusively numerical or a combination of numerical and categorical (heterogeneous). 
These datasets are further classified according to their sample size: medium-sized or large-sized. 
The dataset counts from Tabular Benchmark is provided in \cref{sec:benchmark-dataset-count}, and the detailed datasets used in Tabular Benchmark is provided in \cref{sec:id-mapping}.

\textbf{Model Selection.} 
For model comparison, Tabular Benchmark includes four tree-based models: RandomForest, GradientBoostingTree \cite{friedman2002stochastic}, HGBT \cite{scikit-learn}, and XGBoost; two generic DNN models: MLP and ResNet \cite{gorishniy2021revisiting}; and two tabular DNN models: SAINT and FT-Transformer. To ensure a comprehensive comparison, we also included two additional tree-based models: LightGBM and CatBoost, and three tabular DNN models: NODE, Trompt, and GRANDE. LightGBM and CatBoost are selected due to their widespread use across various domains; 
NODE and GRANDE both share similar motivation and high-level structure with DOFEN; Trompt represents the current state-of-the-art tabular DNNs when following the origin protocols of the Tabular Benchmark.
The default hyperparameter configuration of DOFEN and hyperparameter search space of different models are presented in \cref{sec:params,sec:search-space}, and the list of some missing model baselines from Tabular Benchmark is provided in \cref{sec:benchmark-missing-baseline}.

\subsection{Performance Evaluation}
\label{sec:eval}

In this section, we evaluate DOFEN on the medium-sized benchmark of the Tabular Benchmark for classification and regression tasks separately. The evaluation metrics adhere to the Tabular Benchmark protocols, which use accuracy for classification datasets and the R-squared score for regression datasets. We discuss the overall performance in this section and provide comprehensive results for each dataset in \cref{sec:more-eval}. 

\begin{figure}[t]
    \begin{subfigure}[t]{.25\textwidth}
        \centering
        \captionsetup{justification=centering}
        \includegraphics[width=\textwidth]{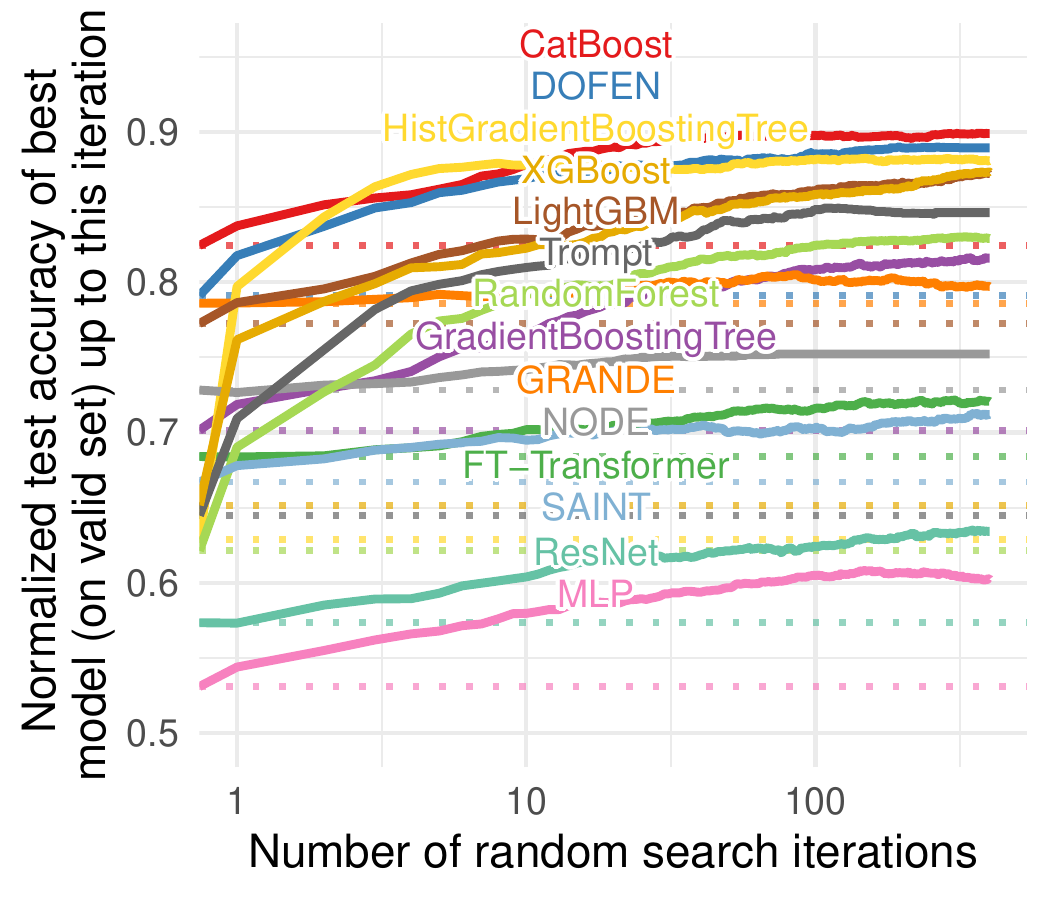}
        \caption{Exclusively Numerical,\\Classification}
        \label{fig:cls-eval-num}
    \end{subfigure}%
    \begin{subfigure}[t]{.25\textwidth}
        \centering
        \captionsetup{justification=centering}
        \includegraphics[width=\textwidth]{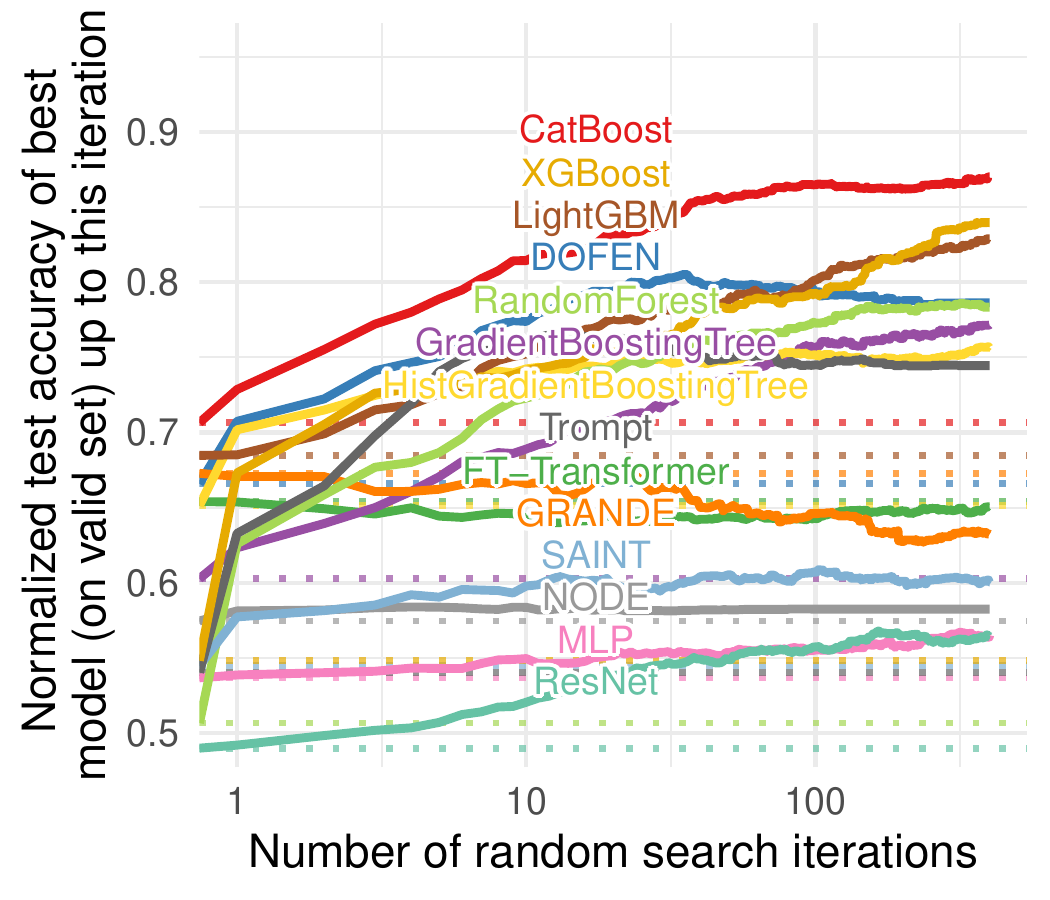}
        \caption{Heterogeneous,\\Classification}
        \label{fig:cls-eval-het}
    \end{subfigure}%
    \label{fig:cls-eval}
    \begin{subfigure}[t]{.25\textwidth}
        \centering
        \captionsetup{justification=centering}
        \includegraphics[width=\textwidth]{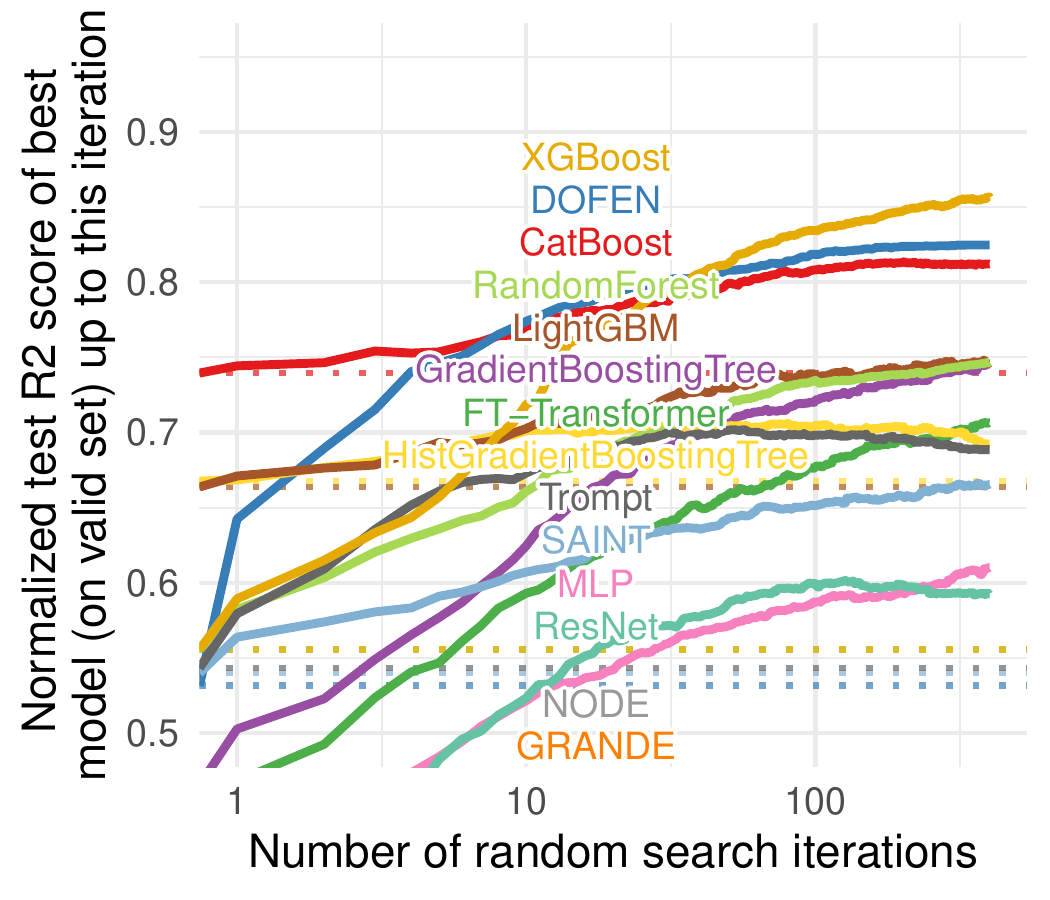}
        \caption{Exclusively Numerical,\\Regression}
        \label{fig:reg-eval-num}
    \end{subfigure}%
    \begin{subfigure}[t]{.25\textwidth}
        \centering
        \captionsetup{justification=centering}
        \includegraphics[width=\textwidth]{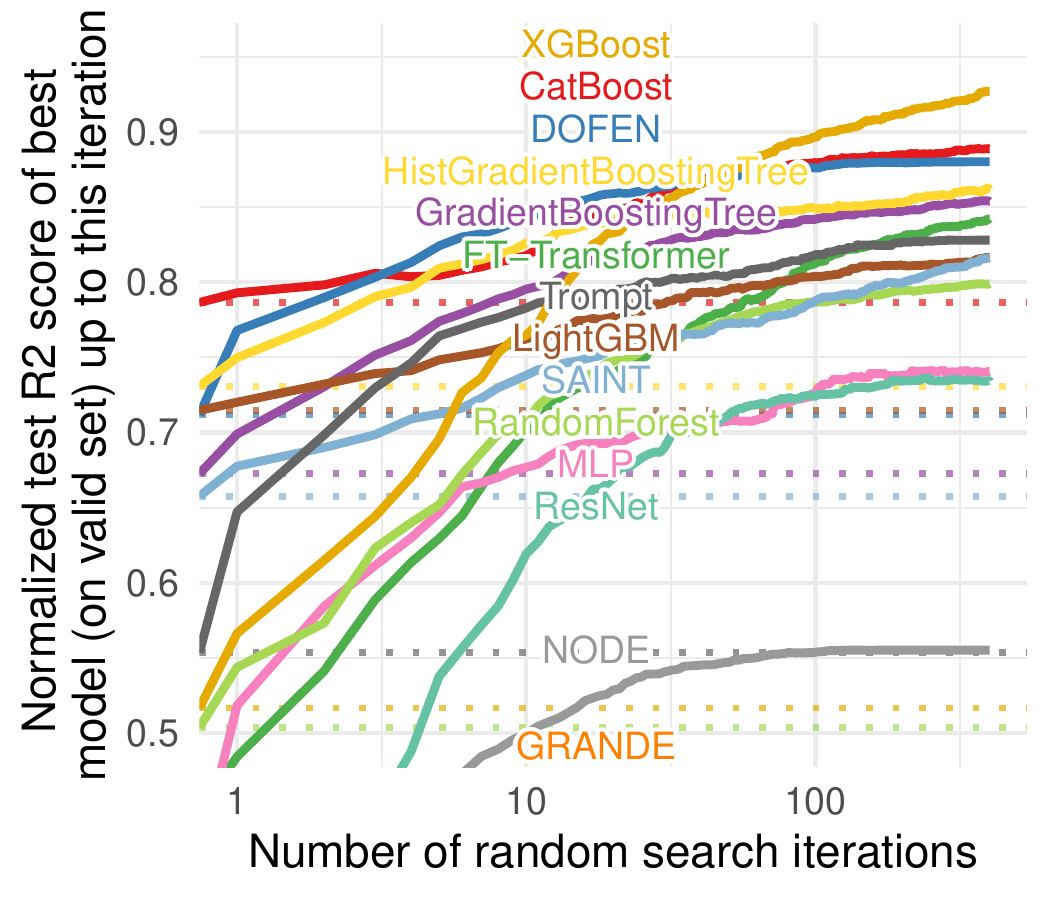}
        \caption{Heterogeneous,\\Regression}
        \label{fig:reg-eval-het}
    \end{subfigure}
    \caption{Results on medium-sized classification and regression datasets. The result are averaged over various datasets included in each benchmark respectively, detailed number of datasets of each benchmark is provided in \cref{sec:benchmark-dataset-count}}
    \label{fig:medium-eval}
\end{figure}

\textbf{Classification.}
In \cref{fig:cls-eval-num}, the models can be grouped into three categories: 1) tree-based models along with three tabular DNN models, DOFEN, Trompt and GRANDE; 2) three other tabular DNN models; and 3) the two generic DNN models. Before DOFEN, Trompt was the only DNN model comparable to tree-based models. DOFEN not only matches but surpasses the performance of most tree-based models, setting a new benchmark for DNN models in tabular data. In \cref{fig:cls-eval-het}, DOFEN and Trompt are again the only DNN models grouped with tree-based models, although they are positioned at the bottom of this group.

\textbf{Regression.}
In \cref{fig:reg-eval-num} and \cref{fig:reg-eval-het}, XGBoost, DOFEN, and CatBoost emerge as a distinct category of top performers. XGBoost consistently leads in performance after the hyperparameter search, with DOFEN and CatBoost holding the next two top positions. Notably, DOFEN is the only deep learning model that is comparable to tree-based models in these two regression benchmarks.

The analysis of \cref{fig:medium-eval} allows us to draw several conclusions. DOFEN consistently ranks first when compared to other DNN models. Additionally, DOFEN shows strong competitiveness against tree-based models, consistently placing within the top three on datasets with numerical features. However, when dealing with heterogeneous features, DOFEN's performance slightly declines, though it still ranks within the top four. This challenge in managing heterogeneous features is common among all DNN models, indicating a potential area for improvement in future tabular DNN models.

As for the results on the large-sized benchmark, we discuss the findings in \cref{sec:eval-large}, where DOFEN also shows superior performance. Beyond the Tabular Benchmark, we conduct additional experiments on datasets used in FT-Transformer and GRANDE paper in \cref{sec:eval-ft-result,sec:eval-grande-result}, respectively, where DOFEN demonstrates competitive results even when using only default hyperparameter settings.

\subsection{Ablation Study}
\label{sec:ablation}

DOFEN contains two key features: (1) randomly combine generated conditions to construct a large pool of rODTs, and (2) introduce a two-level ensemble that randomly aggregates multiple rODT forests. More importantly, both features involved randomization to increase the diversity of rODTs constructed and the rODT forests for ensemble.

To better understand the mechanisms driving DOFEN’s performance, we conduct experiments to ablate the two key features of DOFEN. 
For the first key feature, we lower the diversity of constructed rODTs by removing the operation in \cref{eq:rodt-construction} and let $\mathbf{O}_i=\mathbf{M}_i$ in the following process.
In this case, neighboring columns are always in the same group during rODT construction (\cref{eq:derive-rODT-weight}), leading to a less diverse set of condition combinations. 
For the second key feature, we remove the second level bagging ensemble.
Alternatively, we use all rODTs from \cref{eq:rodt-construction} to form a single forest without the for loop in \cref{algo:two-level-rodt-ensemble}.
In this setup, the softmax function is directly applied to weight vector $\Vec{w}_i$, and the weighted sum of corresponding embeddings $\mathbf{E}$ is calculated. 
This results in a single prediction per sample, unlike the $N_\text{forest}$ predictions described in \cref{algo:two-level-rodt-ensemble}.
The experimental results are presented in \cref{tab:ablation}. 

\textbf{The diversity of condition combinations affects DOFEN's performance.} 
As shown in \cref{tab:ablation}, DOFEN without condition shuffling have lower performance due to its less diverse condition set. 

\textbf{Forest ensemble helps mitigate overfitting and improve DOFEN's performance.} 
On the other hand, the exclusion of the forest ensemble also lowers DOFEN's performance across all datasets, regardless of the type of task or features, with a particularly strong effect on regression tasks. 
To further investigate the significant drop in performance without forest ensemble sampling, we analyzed model performance at various training checkpoints in \cref{sec:overfitting-without-sampling}. 
The result shows that the sampling process mitigates overfitting hence increase DOFEN's testing performance. 

\begin{table}[t]
\centering
\small
\caption{Ablation study of DOFEN. For evaluation metrics, accuracy is used for classification datasets, while the R-squared score for regression datasets.}
\begin{threeparttable}
\begin{tabular}{l c c c c}
\toprule
& \textbf{default} & \textbf{no condition shuffling} & \textbf{no forest ensemble} & \textbf{remove both} \\ \midrule
Classification & \underline{0.7725} & 0.7709 & 0.7362 & 0.7323 \\
 -- Numerical Only & \underline{0.7920} & 0.7904 & 0.7526 & 0.7523 \\
 -- Heterogeneous & \underline{0.7281} & 0.7264 & 0.6988 & 0.6867 \\
\midrule
Regression & \underline{0.6605} & 0.6439 & 0.3238\tnote{*} 
& 0.5441 
\\
 -- Numerical Only & \underline{0.6814} & 0.6527 & 0.1867\tnote{*} 
 & 0.6058 
 \\
 -- Heterogeneous & \underline{0.6371} & 0.6342 & 0.4770 & 0.4751 \\
\bottomrule
\end{tabular}
\begin{tablenotes}
\footnotesize
\item[*] This unexpected result is due to an outlier performance occurs when averaging R-squared scores of datasets. By removing the outlier dataset, the average score is 0.5457 and 0.6107 for all regression datasets and numerical only regression datasets, respectively.
\end{tablenotes}
\end{threeparttable}
\label{tab:ablation}
\end{table}

\subsection{Interpretability}
\label{sec:interpretability}

This section aims to demonstrate the interpretability of DOFEN. Specifically, we adopt a feature importance metric akin to the "split" or "weight" importance used in LightGBM and XGBoost, which counts how often a feature is used in the model.

To calculate DOFEN's feature importance of a specific sample, let $\mathbf{F} \in \mathbb{R}^{N_\textit{rODT} \times N_\textit{col}}$ be a matrix of feature occurrences across different rODTs. We then use the output of sub-module $\Delta_2$, a vector $\Vec{w}_i \in \mathbb{R}^{N_\textit{rODT}}$ (\cref{eq:derive-rODT-weight}), to represent the importance across all rODTs for each sample, as this weight $\Vec{w}_i$ is used for constructing rODT forest to perform prediction in DOFEN model. A softmax operation is further applied to the vector $\Vec{w}_i$ to ensure the importance sums to 1 (also done in line 6 of \cref{algo:two-level-rodt-ensemble}). Finally, we perform a weighted sum between the feature occurrences and the importance of each rODT, resulting in a single vector $\Vec{t}_i \in \mathbb{R}^{N_\text{col}}$ representing DOFEN's feature importance for specific sample. To calculate DOFEN's overall feature importance of a dataset, we simply average the feature importance of all samples in training dataset.

We tested the reliability of DOFEN's feature importance on three real-world datasets: the mushroom dataset, the red wine quality dataset, and the white wine quality dataset, following the experimental design used by Trompt. The results for the mushroom dataset are shown in \cref{tab:importance-mushroom}, with results for the wine quality datasets provided in \cref{sec:more-interpretability}. The results indicate that the top-3 important features identified by DOFEN align closely with those selected by other tree-based models, with only minor ranking differences. This demonstrates DOFEN's ability to reliably identify key features while maintaining interpretability despite its deep learning architecture.

\begin{table}[t]
    \small
    \caption{Top 3 Feature importance of DOFEN on mushroom dataset.}
    \centering
    \begin{tabular}{l l l l}
    \toprule
        ~ & \textbf{1st} & \textbf{2nd} & \textbf{3rd} \\
    \midrule
        Random Forest & odor (15.11\%) & gill-size (12.37\%) & gill-color (10.42 \%) \\
        XGBoost & spore-print-color (29.43\%) & odor (22.71\%) & cap-color (14.07\%) \\
        LightGBM & spore-print-color (22.08\%) & gill-color (14.95\%) & odor (12.96\%) \\
        CatBoost & odor (72.43\%) & spore-print-color (10.57\%) & gill-size (2.71\%) \\
        GradientBoostingTree & gill-color (31.08\%) & spore-print-color (19.89\%) & odor (17.44\%) \\
        Trompt & odor (24.93\%) & gill-size (8.13\%) & gill-color (5.73\%) \\
        DOFEN (ours) & odor (13.15\%) & spore-print-color (6.84\%) & gill-size (5.58\%) \\
    \bottomrule
    \end{tabular}
    \label{tab:importance-mushroom}
\end{table}

\subsection{Stability of DOFEN}
\label{sec:impact-of-random}

Randomness plays an important role in DOFEN, where it is incorporated at two steps: first, in the selection of conditions as shown in \cref{eq:rodt-construction} for rODT construction, and second, in the sampling of rODTs as shown in \cref{algo:two-level-rodt-ensemble} for a two-level rODT ensemble.
This section explores how randomness affects the stability of DOFEN. 

We begin by analyzing the variation in performance across four datasets where DOFEN ranks first, as shown in \cref{tab:analysis-seed}.
As shown in \cref{tab:analysis-seed}, the standard deviations are even negligible when $N_\text{forest}=1$ (about $0.1\%$ to $1\%$ to mean), except for the delays-zurich dataset. 
Moreover, with increased $N_\text{forest}$, the standard deviations become even smaller (about $0.01\%$ to $0.1\%$ to mean). 

These results suggest that the stability of DOFEN is not an issue in most cases ($N_{forest}>10$), and using the default setting of DOFEN (100 forests) ensures both adequate performance and stability for most datasets.
Furthermore, the performance improves as the $N_\text{forest}$ increases, indicating that the tree bagging of DOFEN not only mitigates instability but also enhances the model's generalizability.



\begin{table}[t]
\centering
\small
\caption{Mean ($\mu$) and standard deviation ($\sigma$) of DOFEN's performance under default hyperparameters with 15 random seeds on 4 datasets from different tasks. For evaluation metrics, accuracy is used for classification datasets, while the R-squared score for regression datasets.}

\label{tab:analysis-seed}
\begin{tabular}{l c c c c c c c}
\toprule
& $N_\text{forest}$ & \textbf{1} & \textbf{10} & \textbf{20} & \textbf{50} & \textbf{\makecell{100 (default)}} & \textbf{400} \\ \midrule
\multirow{2}{*}{\makecell[l]{jannis \\ (numerical classification)}} & $\mu$ ($\uparrow$) & 0.7382 & 0.7747 & 0.7782 & 0.7800 & 0.7808 & \underline{0.7814} \\
& $\sigma$ ($\downarrow$) & 0.0060& 0.0019 & 0.0015 & 0.0006 & 0.0007 & \underline{0.0004} \\ \midrule
\multirow{2}{*}{\makecell[l]{road-safety \\ (heterogeneous classification)}} & $\mu$ ($\uparrow$) & 0.7517 & 0.7712 & 0.7720 & 0.7728 & \underline{0.7732} & \underline{0.7732} \\
& $\sigma$ ($\downarrow$) & 0.0118 & 0.0010 & 0.0007 & 0.0004 & 0.0005 & \underline{0.0003} \\ \midrule
\multirow{2}{*}{\makecell[l]{delays-zurich \\ (numerical regression)}} & $\mu$ ($\uparrow$) & 0.0054 & 0.0248 & 0.0258 & 0.0265 & 0.0268 & \underline{0.0270}\\
& $\sigma$ ($\downarrow$) & 0.0033 & 0.0009 & 0.0005 & 0.0003 & 0.0003 & \underline{0.0002} \\ \midrule
\multirow{2}{*}{\makecell[l]{abalone \\ (heterogeneous regression)}} & $\mu$ ($\uparrow$) & 0.5469 & 0.5810& 0.5846 & 0.5862 & 0.5868 & \underline{0.5870}\\
& $\sigma$ ($\downarrow$) & 0.0181 & 0.0038 & 0.0026 & 0.0017 & 0.0010& \underline{0.0004} \\
\bottomrule
\end{tabular}
\end{table}
\section{Related Work}
\label{related-work}
In this section, we categorize deep tabular neural networks into two main streams: tree-inspired DNN architectures and novel DNN architectures. By comparing DOFEN with these established models, we aim to showcase its unique contributions and position it within the broader landscape of deep tabular network research.

\textbf{Tree-inspired DNN Architectures.}
Integrating decision tree (DT) algorithms with DNNs has become prominent for handling tabular data. Pioneering works like Deep Forest \cite{zhou2019deep}, NODE \cite{popov2019neural}, TabNet \cite{arik2021tabnet}, GradTree \cite{marton2024gradtree} and GRANDE \cite{marton2024grande} have each introduced unique methodologies.

Deep Forest adapts the random forest algorithm and incorporates multi-grained feature scanning to leverage the representation learning capabilities of DNNs.
TabNet models the sequential decision-making process of traditional decision trees using a DNN, featuring a distinct encoder-decoder architecture that enables self-supervised learning.
GradTree recognizes the importance of hard, axis-aligned splits for tabular data and uses a straight-through operator to handle the non-differentiable nature of decision trees, allowing for the end-to-end training of decision trees.
NODE and GRANDE share a similar observation and high-level structure to DOFEN, in that they ensemble multiple tree-like deep learning base models.
NODE uses ODT as a base predictor and employs a DenseNet-like multi-layer ensemble to boost performance. GRANDE, a successor to GradTree, uses DT as a base predictor and introduces advanced instance-wise weighting for ensembling each base model's prediction.

However, DOFEN distinguishes itself from NODE and GRANDE through its unique architectural design.
First, DOFEN employs a different approach to transforming tree-based models into neural networks.
Unlike NODE and GRANDE, which explicitly learn the decision paths (i.e., selecting features and thresholds for each node) and the leaf node values of a tree, DOFEN randomly selects features to form rODTs and uses a neural network to measure how well a sample aligns with the decision rule.
Additionally, the leaf node value of an rODT is replaced with an embedding vector for further ensembling.
Second, DOFEN introduces a novel two-level ensemble process to enhance model performance and stability.
Unlike NODE and GRANDE, which simply perform a weighted sum on base model predictions, DOFEN first constructs multiple rODT forests by randomly aggregating selected rODT embeddings and then applies bagging on the predictions of these rODT forests.

\textbf{Novel DNN Architectures.}
Beyond merging decision tree algorithms with DNNs, significant progress has been made in developing novel architectures for tabular data. Notable among these are TabTransformer \cite{huang2020tabtransformer}, FT-Transformer \cite{gorishniy2021revisiting}, SAINT \cite{somepalli2021saint}, TabPFN \cite{hollmann2022tabpfn}, and Trompt \cite{chen2023trompt}. These models primarily leverage the transformer architecture \cite{vaswani2017attention}, utilizing self-attention mechanisms to capture complex feature relationships.

TabTransformer applies transformer blocks specifically to numerical features, while FT-Transformer extends this approach to both numerical and categorical features. SAINT enhances the model further by applying self-attention both column-wise and sample-wise, increasing its capacity. TabPFN, a variant of the Prior Fitted Network (PFN) \cite{muller2021transformers}, is particularly effective with smaller datasets. 
Trompt introduces an innovative approach by incorporating prompt learning techniques from natural language processing \cite{radford2018improving}, aiming to extract deeper insights from the tabular data's columnar structure. 

These architectures have demonstrated impressive performance across various studies and benchmark datasets and have been chosen as baselines in our performance evaluation, offering a comprehensive view of the current state of the art in deep learning for tabular data.
\section{Limitation}
\label{sec:limitation}

Although DOFEN shows promising results, it still contains two weaknesses.
First, the inference time of DOFEN is relatively long compared to other DNN models, as shown in \cref{sec:computation-analysis}.
However, \cref{sec:computation-analysis} also shows that DOFEN possesses the fewest floating point operations (FLOPs). This inconsistency between inference time and FLOPs is mainly caused by the group convolution operation for calculating weights for each rODT (\cref{sec:long-inference-time-dofen}), which can be improved in the future implementation of DOFEN.
Second, the randomization steps involved in DOFEN result in a slower convergence speed, meaning that DOFEN requires more training steps to reach optimal performance.
This is reflected in the relatively larger number of training epochs needed for DOFEN.
Therefore, the workaround strategy of differentiable sparse selection proposed in this study is merely a starting point, demonstrating its potential. Finding more efficient strategies will be the future work.

\section{Conclusion}
\label{sec:conclusion}
In this work, we introduced DOFEN, a novel DNN architecture inspired by oblivious decision trees for tabular data. DOFEN first randomly combines conditions generated for each column to construct various relaxed oblivious decision trees (rODTs) and further enhances performance through a novel two-level rODT forest ensembling process.

We evaluate DOFEN on a well-recognized benchmark: the Tabular Benchmark, where DOFEN achieves state-of-the-art performance among DNN-based models and significantly narrows the gap between DNNs and traditional tree-based methods.
We conducted an ablation study and analysis, which shows that the novel two-level rODT forest ensembling process not only contributes the most to DOFEN's superior performance but also maintains the stability of the randomization process involved in DOFEN.
Moreover, due to DOFEN's tree-like structure, its decision-making process is interpretable, which is an important feature for deep learning models.

In summary, DOFEN shows great potential as a versatile backbone for tabular data across various scenarios, with its outstanding performance and interpretability, including self- and semi-supervised learning and multi-modal training.


\newpage
\bibliography{paper}
\bibliographystyle{achemso}


\newpage
\appendix
\addcontentsline{toc}{section}{Appendix} 
\part{Appendix} 
\parttoc 
\newpage
\section{More DOFEN Settings}

\subsection{Default Hyperparameters Settings for DOFEN}
\label{sec:params}

In this section, we describe the hyperparameters used in our DOFEN model, along with their default values, as shown in \cref{tab:default-params}.
All notations used here have been previously introduced in \cref{sec:main-idea}, except for \textit{dropout\_rate} and $N_\text{head}$.
The \textit{dropout\_rate} is applied in dropout layers, and its usage is detailed in \cref{sec:model-layers}. The $N_\text{head}$ is for the multi-head extension of rODT weighting mechanism, and its detail is provided in \cref{sec:multihead-dofen}.

The calculated $N_\text{estimator}$ for each dataset can be found in \cref{sec:n_estimator-val}.
Additionally, the hyperparameter search spaces for both the DOFEN model and all baseline models are detailed in \cref{sec:search-space}.

DOFEN is implemented in Pytorch \cite{paszke2019pytorch}. For hyperparameters used in model optimization (e.g. optimizer, learning rate, weight decay, etc.), all experiments share the same settings. Specifically, DOFEN uses AdamW optimizer \cite{loshchilov2017decoupled} with $1\mathrm{e}{-3}$ learning rate and no weight decay. The batch size of DOFEN is set to 256 and is trained for 500 epochs without using learning rate scheduling or early stopping if not specified.

\begin{table}[h]
\centering
\small
\setlength\tabcolsep{15.0pt}
\caption{The default hyperparameters of DOFEN.}
\begin{threeparttable}
\begin{tabular}{l l}
\toprule
\textbf{Hyperparameter}& \textbf{Default Value}     \\ \midrule
$N_\text{col}$                        & depends on dataset         \\ 
$d$ \tnote{1}& 4                          \\ 
$m$ \tnote{2}& 16                          \\ 
$N_\text{cond}$                       & $md$\\ 
$N_\text{rODT}$                       & $N_\text{col}N_\text{cond}/d = N_\text{col}m$\\
$N_\text{head}$ & 1\\
$N_\text{estimator}$                  & $\max\{2, \floor{\sqrt{N_\text{col}}}\}\cdot N_\text{cond}/d$ \\
$N_\text{forest}$                     & 100                        \\
$N_\text{hidden}$                     & 128                        \\
$N_\text{class}$                      & depends on dataset         \\
$\textit{dropout\_rate}$& 0.0         \\
\bottomrule
\end{tabular}
\begin{tablenotes}
\footnotesize
\item[1] depth of a rODT
\item[2] an intermediate parameter to ensure that $N_\text{rODT}$ is an integer
\end{tablenotes}
\end{threeparttable}
\label{tab:default-params}
\end{table}

\subsection{Detailed Model Configurations}
\label{sec:model-layers}

In this appendix, we elucidate the specific configurations of the neural network layer composites, denoted as $\Delta_1$, $\Delta_2$, and $\Delta_3$ in the main paper.

\begin{figure}[h]
    \centering
    \includegraphics[width=.4\linewidth]{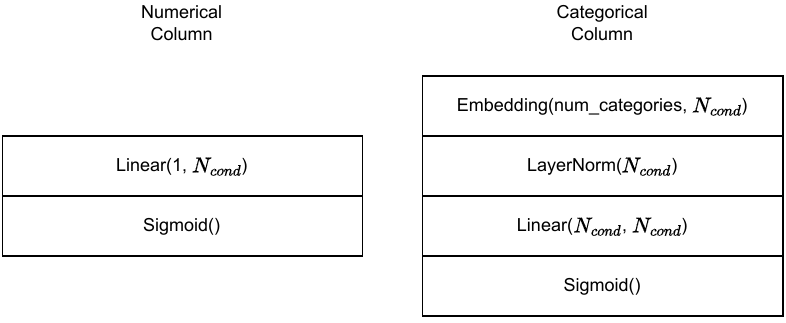}
    \caption{Detailed network layer composite for $\Delta_1$.}
    \label{fig:delta-1}
\end{figure}

\begin{figure}[h]
    \centering
    \begin{minipage}{.45\textwidth}
        \centering
        \includegraphics[width=.4\linewidth]{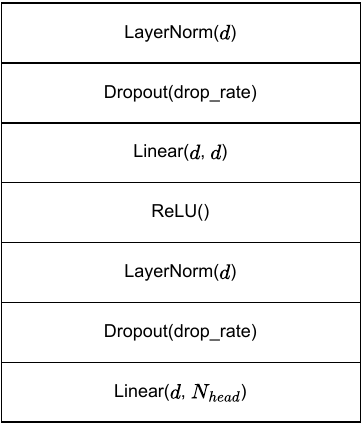}
        \caption{Detailed network layer composite for $\Delta_2$.}
        \label{fig:delta-2}
    \end{minipage}
    \hfill
    \begin{minipage}{.45\textwidth}
        \centering
        \includegraphics[width=.4\linewidth]{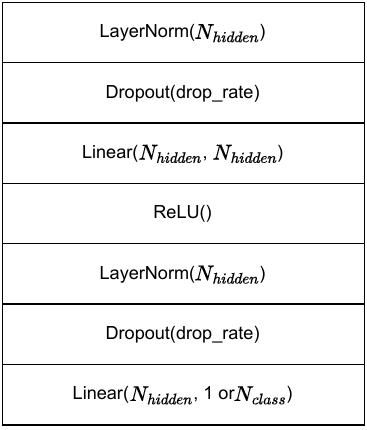}
        \caption{Detailed network layer composite for $\Delta_3$.}
        \label{fig:delta-3}
    \end{minipage}
    \hfill
\end{figure}

\begin{enumerate}
    \item $\Delta_1$ - Generate conditions for each column:
    $\Delta_1$ is designed to generate conditions for both numerical and categorical data columns, as detailed in \cref{fig:delta-1}.
    For categorical columns in particular, we employ embedding layers.
    These layers are utilized to transform categorical features into a format that the neural network can effectively process.
    \item $\Delta_2$ and $\Delta_3$ - Derive weights and make predictions:
    The layers represented by $\Delta_2$ and $\Delta_3$ are responsible for generating weights based on the combination of conditions and making predictions, respectively.
    The relevant structures and processes are illustrated in \cref{fig:delta-2} and \cref{fig:delta-3}.
    \item Most parameters and their notations used here have been defined in the main paper and \cref{sec:params}, despite \emph{num\_categories}. This parameter represents the number of distinct categories in a given categorical column.
\end{enumerate}

\subsection{Actual $N_\text{estimator}$ for each Dataset}
\label{sec:n_estimator-val}

The $N_\text{estimator}$ is calculated through a pre-defined formula as shown in \cref{tab:default-params}.
In this section, we provide the calculated $N_\text{estimator}$ for each dataset in \cref{tab:n_estimator-val} when using default hyperparameters.
Datasets are represented by their OpenML ID as described in \cref{sec:id-mapping}.

\begin{table}[H]
\centering
\small
\caption{$N_\text{estimator}$ for each dataset, as long as their $N_\text{col}$ and $N_\text{rODT}$. }
\begin{tabular}{l c c c c c c c c c}
\toprule
OpenML ID & 361086 & 361294 & 361094 & 361289 & 361293 & 361085 & 361082 & 361103 & 361080 \\ \midrule
$N_\text{col}$ & 3 & 3 & 4 & 4 & 5 & 6 & 6 & 6 & 6 \\
$N_\text{rODT}$ & 48 & 48 & 64 & 64 & 80 & 96 & 96 & 96 & 96 \\
$N_\text{estimator}$ & 32 & 32 & 32 & 32 & 32 & 32 & 32 & 32 & 32 \\
\midrule
OpenML ID & 361273 & 361066 & 361060 & 361280 & 361093 & 361110 & 361288 & 361281 & 361277 \\ \midrule
$N_\text{col}$ & 7 & 7 & 7 & 7 & 7 & 8 & 8 & 8 & 8 \\
$N_\text{rODT}$ & 112 & 112 & 112 & 112 & 112 & 128 & 128 & 128 & 128 \\
$N_\text{estimator}$ & 32 & 32 & 32 & 32 & 32 & 32 & 32 & 32 & 32 \\
\midrule
OpenML ID & 361081 & 361078 & 361104 & 361083 & 361096 & 361055 & 361061 & 361065 & 361095 \\ \midrule
$N_\text{col}$ & 8 & 8 & 9 & 9 & 9 & 10 & 10 & 10 & 10 \\
$N_\text{rODT}$ & 128 & 128 & 144 & 144 & 144 & 160 & 160 & 160 & 160 \\
$N_\text{estimator}$ & 32 & 32 & 48 & 48 & 48 & 48 & 48 & 48 & 48 \\
\midrule
OpenML ID & 361076 & 361098 & 361291 & 361099 & 361286 & 361087 & 361084 & 361063 & 361074 \\ \midrule
$N_\text{col}$ & 11 & 11 & 11 & 11 & 11 & 13 & 15 & 16 & 16 \\
$N_\text{rODT}$ & 176 & 176 & 176 & 176 & 176 & 208 & 240 & 256 & 256 \\
$N_\text{estimator}$ & 48 & 48 & 48 & 48 & 48 & 48 & 48 & 64 & 64 \\
\midrule
OpenML ID & 361079 & 361101 & 361102 & 361070 & 361275 & 361072 & 361283 & 361278 & 361111 \\ \midrule
$N_\text{col}$ & 16 & 16 & 17 & 20 & 20 & 21 & 21 & 22 & 23 \\
$N_\text{rODT}$ & 256 & 256 & 272 & 320 & 320 & 336 & 336 & 352 & 368 \\
$N_\text{estimator}$ & 64 & 64 & 64 & 64 & 64 & 64 & 64 & 64 & 64 \\
\midrule
OpenML ID & 361069 & 361062 & 361073 & 361282 & 361285 & 361077 & 361279 & 361068 & 361113 \\ \midrule
$N_\text{col}$ & 24 & 26 & 26 & 31 & 32 & 33 & 42 & 50 & 54 \\
$N_\text{rODT}$ & 384 & 416 & 416 & 496 & 512 & 528 & 672 & 800 & 864 \\
$N_\text{estimator}$ & 64 & 80 & 80 & 80 & 80 & 80 & 96 & 112 & 112 \\
\midrule
OpenML ID & 361274 & 361088 & 361091 & 361292 & 361287 & 361097 & 361276 \\ \midrule
$N_\text{col}$ & 54 & 79 & 91 & 124 & 255 & 359 & 419 \\
$N_\text{rODT}$ & 864 & 1264 & 1456 & 1984 & 4080 & 5744 & 6704 \\
$N_\text{estimator}$ & 112 & 128 & 144 & 176 & 240 & 288 & 320 \\
\bottomrule
\end{tabular}
\label{tab:n_estimator-val}
\end{table}

\subsection{Multi-head extension of DOFEN}
\label{sec:multihead-dofen}

This section shows the extended multi-head version of the weighting mechanism when constructing rODT Forests (\cref{sec:rODT-ensemble}).

In the single-head version of weighting, a weight scalar $w_{ij}$ (\cref{eq:derive-rODT-weight}) of each rODT is the corresponding weight when aggregating the paired embedding vector $\Vec{e}_i$ (\cref{eq:rodt-embedding}) to form an rODT forest.
In multi-head version of weighting, we change the weight of each rODT from a scalar into a $N_\text{head}$ dimension vector $\Vec{w}_{ij}$, as shown in \cref{eq:derive-rODT-weight-multihead}, while each head dimension of weight is responsible for weighting a part of the paired embedding vector instead of full dimensionality. 
This concept is similar to the one used in a multi-head attention \citep{vaswani2017attention}, where each head can learn different weighting, enhancing the capacity and diversity of the weighting process. 

The full process of multi-head weighting is shown in \cref{algo:multihead-two-level-rodt-ensemble}. Here, we focus on mentioning steps that are different from a single-head weighting process described in \cref{algo:two-level-rodt-ensemble}: 

\begin{enumerate}
    \item First, in line 6 and 7 of \cref{algo:multihead-two-level-rodt-ensemble}, the embedding vector should be reshaped into a matrix for an additional head dimension, while the hidden dimension is reduced for each head. 
    \item Second, in line 9 to 15 of \cref{algo:multihead-two-level-rodt-ensemble}, we iterate through different heads, each weight and embedding of corresponding head are weighted summed to form part of the embedding for an rODT forest.
    \item Lastly, in line 8, 16, and 18 of \cref{algo:multihead-two-level-rodt-ensemble}, we concatenate these embeddings from each head to form a complete embedding of an rODT forest.
\end{enumerate}

Moreover, this multi-head weighting does not increase much of computational cost, as the embedding dimension is reduced of each head.

\begin{gather}
    \mathbf{w}_i = 
    \begin{bmatrix}    
        \Delta_{21}((o_{i11},\dots,o_{i1d})) \\
        \vdots \\
        \Delta_{2N_\text{rODT}}((o_{iN_\text{rODT}1},\dots,o_{iN_\text{rODT}d})) \\
    \end{bmatrix} 
    =
    \begin{bmatrix}
    \Vec{w}_{i1} \\
    \vdots \\
    \Vec{w}_{iN_\text{rODT}}
    \end{bmatrix} 
    \in \mathbb{R}^{N_\text{rODT} \times N_\text{head}},\Vec{w}_{ij} \in \mathbb{R}^{N_\text{head}}, \nonumber \\ \nonumber \\ 
    \text{where}\;j = (1,2,\dots,N_\text{rODT})
\label{eq:derive-rODT-weight-multihead}
\end{gather}

\begin{algorithm}
\caption{Multi-head version of Two-level Relaxed ODT Ensemble}
\label{algo:multihead-two-level-rodt-ensemble}
\SetAlgoLined
\KwIn{$\mathbf{w}_i$, $\mathbf{E}$, $N_\text{forest}$, $N_\text{head}$,
$y_i$,
$\mathcal{L}$}
\KwOut{$\hat{y}_i$, $loss_i$}

\textbf{Initialize} $\hat{y}_i, loss_i \gets 0, 0$\;
\For{$r \gets 1$ \KwTo $N_\text{forest}$}{
    $\mathbf{w}_i', \mathbf{E}' \xleftarrow{\text{sample without replacement}} \mathbf{w}_i, \mathbf{E}$\ \tcc*[r]{$N_\text{estimator}$ paired elements are sampled.}

    $\mathbf{w}_i' \in \mathbb{R}^{N_\text{estimator} \times N_\text{head}}$\;
    $\mathbf{E}' \in \mathbb{R}^{N_\text{estimator} \times N_\text{hidden}}$\;

    $\mathbf{E}'' \xleftarrow{\text{reshape}} \mathbf{E}'$\
    \tcc*[r]{Reshape matrix for a $N_\text{head}$ dimension.}
    
    $\mathbf{E}'' \in \mathbb{R}^{N_\text{estimator} \times N_\text{head} \times {\frac{N_\text{hidden}}{N_\text{head}}}}$\;

    $\mathbf{F} \gets []$\tcc*{Initialize an empty list for concatenation}
    
    \For{$h \gets 1$ \KwTo $N_\text{head}$}{
        $\Vec{w}_i' \gets \mathbf{w}_i'[:,h]$\;
        $\Vec{w}_i' \in \mathbb{R}^{N_\text{estimator}}$\;

        $\mathbf{E}''' \gets \mathbf{E}''[:,h,:]$\;
        $\mathbf{E}''' \in \mathbb{R}^{N_\text{estimator} \times {\frac{N_\text{hidden}}{N_\text{head}}}}$\;

        $\Vec{f}_i \gets \mathlarger{\sum}^{N_\text{estimator}}\text{softmax}(\Vec{w}_i')\circ\mathbf{E}'''$\tcc*[r]{Element-wise multiplication with broadcast.}
        $\Vec{f}_i \in \mathbb{R}^{\frac{N_\text{hidden}}{N_\text{head}}}$\tcc*[r]{$\Vec{f}_i$ represents an rODT forest embedding.}
    
        $\mathbf{F} \gets \mathbf{F} \parallel \Vec{f}_i$\tcc*{Concatenate $\Vec{f}_i$ to $\mathbf{F}$}
    }
    $\mathbf{F} \in \mathbb{R}^{N_\text{hidden}}$\;
    $\hat{y}_i' \gets \Delta_3(\mathbf{F})$ \tcc*[r]{Give prediction with a shared $\Delta_3$.}
    $loss_i \gets loss_i + \mathcal{L}(\hat{y}_i', y_i)$ \tcc*[r]{Calculate loss with loss function $\mathcal{L}$ and aggregate.}
    $\hat{y}_i \gets \hat{y}_i + \hat{y}_i'$\tcc*[r]{Aggregate each forest's prediction.} 
}
$\hat{y}_i \gets \hat{y}_i/N_\text{forest}$\;
\Return $(\hat{y}_i, loss_i)$ \;
\end{algorithm}
\section{More Tabular Benchmark Settings}

\subsection{Dataset Counts}
\label{sec:benchmark-dataset-count}

In this section, we provide the dataset counts for each task for your reference, as presented in \cref{tab:dataset-count}.

\begin{table}[H]
\centering
\small
\caption{Dataset counts for each task.}
\begin{tabular}{c c c}
\toprule
\textbf{Task} & \textbf{Feature} & \textbf{Count} \\ \midrule
\multirow{2}{*}{medium-sized classification} & numerical & 16 \\
& heterogeneous & 7 \\
\midrule
\multirow{2}{*}{medium-sized regression} & numerical & 19 \\
& heterogeneous & 17 \\
\midrule
\multirow{2}{*}{large-sized classification} & numerical & 4 \\
& heterogeneous & 2 \\
\midrule
\multirow{2}{*}{large-sized regression} & numerical & 3 \\
& heterogeneous & 5 \\
\bottomrule
\end{tabular}
\label{tab:dataset-count}
\end{table}

\subsection{Missing Model Baselines}
\label{sec:benchmark-missing-baseline}

We found that two baselines, MLP and HGBT, are absent from the evaluation results in the large-sized classification task because they are missing from the official repository.
Furthermore, MLP, HGBT, and RandomForest are not included in the large-sized regression task for the same reason.

\subsection{Mappings of OpenML Task ID and Dataset Name}
\label{sec:id-mapping}

In this section, we introduce the mappings between OpenML Task IDs and elaborate on how to download the corresponding datasets using these IDs.

The mappings are provided in \cref{tab:mapping-num-cls,tab:mapping-het-cls,tab:mapping-num-reg,tab:mapping-het-reg}.
To access the datasets, please follow the links below, which direct you to the OpenML website for each type of dataset.
You can then search using the OpenML ID.

\begin{itemize}
    \item Classification datasets with numerical features only: \\ \url{https://www.openml.org/search?type=benchmark&study_type=task&id=337}
    \item Classification datasets with heterogeneous features: \\ \url{https://www.openml.org/search?type=benchmark&study_type=task&id=334}
    \item Regression datasets with numerical features only: \\ \url{https://www.openml.org/search?type=benchmark&study_type=task&id=336} \\ \url{https://www.openml.org/search?type=benchmark&study_type=task&id=297} (only for task ID 361091)
    \item Regression datasets with heterogeneous features: \\ \url{https://www.openml.org/search?type=benchmark&study_type=task&id=335} \\ \url{https://www.openml.org/search?type=benchmark&study_type=task&id=299} (only for task ID 361095)
\end{itemize}

\begin{table}[H]
\centering
\small
\caption{OpenML Task ID mappings for \textbf{classification} datasets with \textbf{numerical features only}.}
\begin{tabular}{c c}
\toprule
\textbf{OpenML ID} & \textbf{Dataset} \\ \midrule
361055 & credit \\
361060 & electricity \\
361061 & covertype \\
361062 & pol \\
361063 & house\_16H \\
361065 & MagicTelescope \\
361066 & bank-marketing \\
361068 & MiniBooNE \\
361069 & Higgs \\
361070 & eye\_movements \\
361273 & Diabetes130US \\
361274 & jannis \\
361275 & default-of-credit-card-clients \\
361276 & Bioresponse \\
361277 & california \\
361278 & heloc \\
\bottomrule
\end{tabular}
\label{tab:mapping-num-cls}
\end{table}

\begin{table}[H]
\centering
\small
\caption{OpenML Task ID mappings for \textbf{classification} datasets with \textbf{heterogeneous features}.}
\begin{tabular}{c c}
\toprule
\textbf{OpenML ID} & \textbf{Dataset} \\ \midrule
361110 & electricity \\
361111 & eye\_movements \\
361113 & covertype \\
361282 & albert \\
361283 & default-of-credit-card-clients \\
361285 & road-safety \\
361286 & compas-two-years \\
\bottomrule
\end{tabular}
\label{tab:mapping-het-cls}
\end{table}

\begin{table}[H]
\centering
\small
\caption{OpenML Task ID mappings for \textbf{regression} datasets with \textbf{numerical features only}.}
\begin{tabular}{c c}
\toprule
\textbf{OpenML ID} & \textbf{Dataset} \\ \midrule
361072 & cpu\_act \\
361073 & pol \\
361074 & elevators \\
361076 & wine\_quality \\
361077 & Ailerons \\
361078 & houses \\
361079 & house\_16H \\
361080 & diamonds \\
361081 & Brazilian\_houses \\
361082 & Bike\_Sharing\_Demand \\
361083 & nyc-taxi-green-dec-2016 \\
361084 & house\_sales \\
361085 & sulfur \\
361086 & medical\_charges \\
361087 & MiamiHousing2016 \\
361088 & superconduct \\
361091 & year \\
361279 & yprop\_4\_1 \\
361280 & abalone \\
361281 & delays\_zurich\_transport \\
\bottomrule
\end{tabular}
\label{tab:mapping-num-reg}
\end{table}

\begin{table}[H]
\centering
\small
\caption{OpenML Task ID mappings for \textbf{regression} datasets with \textbf{heterogeneous features}.}
\begin{tabular}{c c}
\toprule
\textbf{OpenML ID} & \textbf{Dataset} \\ \midrule
361093 & analcatdata\_supreme \\
361094 & visualizing\_soil \\
361095 & black\_friday \\
361096 & diamonds \\
361097 & Mercedes\_Benz\_Greener\_Manufacturing \\
361098 & Brazilian\_houses \\
361099 & Bike\_Sharing\_Demand \\
361101 & nyc-taxi-green-dec-2016 \\ 
361102 & house\_sales \\
361103 & particulate-matter-ukair-2017 \\
361104 & SGEMM\_GPU\_kernel\_performance \\
361287 & topo\_2\_1 \\
361288 & abalone \\
361289 & seattlecrime6 \\
361291 & delays\_zurich\_transport \\
361292 & Allstate\_Claims\_Severity \\
361293 & Airlines\_DepDelay\_1M \\
361294 & medical\_charges \\
\bottomrule
\end{tabular}
\label{tab:mapping-het-reg}
\end{table}
\section{Computational Efficiency Analysis}

\subsection{Computational Efficiency Analysis}
\label{sec:computation-analysis}

To discuss the computational efficiency, we analyzed the average floating point operations (FLOPs) \cite{fvcore}, parameter sizes, and inference time of DOFEN and other baseline models. Our analyses covered both the default and optimal hyperparameter settings, where the optimal hyperparameter delivers the best performance for each model on each dataset. The experiments involving DNN-based models were performed using an NVIDIA GeForce RTX 2080 Ti, while those for the GBDT-based models utilized an AMD EPYC 7742 64-core Processor with 16 threads.

We begin with the comparison between DNN-based and GBDT-based models. This comparison primarily focuses on inference time, as FLOPs and parameter sizes are applicable for evaluating the efficiency of DNN-based models but cannot be applied to GBDTs. Additionally, inference times under the optimal parameters are provided only when those parameters are available. As shown in Tables \cref{tab:efficency-default-cls-medium} to \cref{tab:efficency-optimal-rgr-medium}, the inference times for all DNN-based models are slower than those for GBDT-based models. This is expected due to the inherent differences between the two types of models.

When compared to other DNN baselines, DOFEN achieves the highest performance, the lowest FLOPs, and the smallest parameter sizes but exhibits the relatively long inference time among all the DNN-based models. This inconsistency between FLOPs and inference time suggests that there is still room for implementation improvements in DOFEN. 
Hence, we conduct additional experiments to analyze which part of the DOFEN model is the computational bottleneck, as discussed in \cref{sec:long-inference-time-dofen}, showing that the bottleneck of DOFEN arises from using group operations when constructing rODTs. 
Although this does not affect DOFEN's article, improvements can be made during future open-source releases.





\begin{table}[H]
\centering
\small
\caption{Computational efficiency analysis of default hyperparameters on medium-sized classification datasets.}
\begin{tabular}{l c c c c}
\toprule
\textbf{Model} & \textbf{Performance (Accuracy)} & \textbf{FLOPs (M)} & \textbf{Parameters (M)} & \textbf{Inference time (sec.)} \\
\midrule
DOFEN & 0.7725 & 0.1845 & 0.0140 & 0.0125 \\
Trompt & 0.7704 & 53.2127 & 3.8608 & 0.0225 \\
FT-Transformer & 0.7662 & 3.3147 & 0.0908 & 0.0058 \\
NODE & 0.7658 & 0.8299 & 0.7525 & 0.0041 \\
XGBoost & 0.7717 & -- & -- & 0.0015 \\
LightGBM & 0.7757 & -- & -- & 0.0016 \\
CatBoost & 0.7777 & -- & -- & 0.0029 \\
\bottomrule
\end{tabular}
\label{tab:efficency-default-cls-medium}
\end{table}

\begin{table}[H]
\centering
\small
\caption{Computational efficiency analysis of optimal hyperparameters on medium-sized classification datasets.}
\begin{threeparttable}
\begin{tabular}{l c c c c}
\toprule
\textbf{Model} & \textbf{Performance (Accuracy)} & \textbf{FLOPs (M)} & \textbf{Parameters (M)} & \textbf{Inference time (sec.)} \\
\midrule
DOFEN & 0.7805 & 0.2093 & 0.0437 & 0.0213 \\
Trompt & 0.7797 & 38.7712 & 2.0398 & 0.0202 \\
FT-Transformer & 0.7686 & 6.0696 & 0.2514 & 0.0061 \\
NODE & 0.7677 & 3.2860 & 2.6778 & 0.0033 \\
XGBoost & 0.7848 & -- & -- & 0.0014 \\
LightGBM \tnote{*}& 0.7838 & -- & -- & N/A \\
CatBoost \tnote{*}& 0.7858 & -- & -- & N/A \\
\bottomrule
\end{tabular}
\begin{tablenotes}
\footnotesize
\item[*] The evaluation results are obtained from the Trompt paper without the corresponding optimal hyperparameters. Thus, the inference time under the optimal hyperparameters is unavailable.
\end{tablenotes}
\end{threeparttable}
\label{tab:efficency-optimal-cls-medium}
\end{table}

\begin{table}[H]
\centering
\small
\caption{Computational efficiency analysis of default hyperparameters on medium-sized regression datasets.}
\begin{tabular}{l c c c c}
\toprule
\textbf{Model} & \textbf{Performance (R2 Score)} & \textbf{FLOPs (M)} & \textbf{Parameters (M)} & \textbf{Inference time (sec.)} \\
\midrule
DOFEN & 0.6611 & 0.1875 & 0.0173 & 0.0105 \\
Trompt & 0.6541 & 45.8507 & 3.8591 & 0.0224 \\
FT-Transformer & 0.6359 & 2.7795 & 0.0909 & 0.0039 \\
NODE & 0.1080 & 0.5839 & 0.5065 & 0.0039 \\
XGBoost & 0.6719 & -- & -- & 0.0012 \\
LightGBM & 0.6832 & -- & -- & 0.0014 \\
CatBoost & 0.6896 & -- & -- & 0.0030 \\
\bottomrule
\end{tabular}
\label{tab:efficency-default-rgr-medium}
\end{table}

\begin{table}[H]
\centering
\small
\caption{Computational efficiency analysis of optimal hyperparameters on medium-sized regression datasets.}
\begin{threeparttable}
\begin{tabular}{l c c c c}
\toprule
\textbf{Model} & \textbf{Performance (R2 Score)} & \textbf{FLOPs (M)} & \textbf{Parameters (M)} & \textbf{Inference time (sec.)} \\
\midrule
DOFEN & 0.6882 & 0.2030 & 0.0364 & 0.0182 \\
Trompt & 0.6830 & 17.9560 & 1.2857 & 0.0200 \\
FT-Transformer & 0.6834 & 9.0576 & 0.2965 & 0.0065 \\
NODE & 0.6631 & 2.1379 & 1.6930 & 0.0035 \\
XGBoost & 0.6985 & -- & -- & 0.0014 \\
LightGBM \tnote{*}& 0.6896 & -- & -- & N/A \\
CatBoost \tnote{*}& 0.6940 & -- & -- & N/A \\
\bottomrule
\end{tabular}
\begin{tablenotes}
\footnotesize
\item[*] The evaluation results are obtained from the Trompt paper without the corresponding optimal hyperparameters. Thus, the inference time under the optimal hyperparameters is unavailable.
\end{tablenotes}
\end{threeparttable}
\label{tab:efficency-optimal-rgr-medium}
\end{table}

\subsection{Long Inference Time of DOFEN}
\label{sec:long-inference-time-dofen}

To find out the computation bottleneck of DOFEN, 
we analyzed the inference time of each DOFEN module in proportion, as shown in \cref{tab:module-inference-time} and \cref{tab:operation-inference-time}, which is averaged across 59 medium-sized datasets with default hyperparameters. Table A1 shows that the Forest Construction module consumes the most inference time. In Table A2, more detailed operations reveal that the sub-module $\Delta_2$ in the Forest Construction module, which generates weights for each rODT, has the longest inference time.

The sub-module $\Delta_2$ is designed with multiple MLP and normalization layers, implemented using group convolution and group normalization to parallelize scoring for each rODT. However, the efficiency of group convolution in PyTorch has been problematic and remains unresolved. Specifically, the operation efficiency decreases as the number of groups increases, sometimes making it slower than separate convolutions in CUDA streams (see PyTorch issues 18631, 70954, 73764). The sub-module $\Delta_1$ also uses group convolution to parallelize condition generation across different numerical columns, resulting in slower inference times compared to other operations, though less significant than $\Delta_2$ due to fewer groups being used.

However, we mainly focus on the concept and model structure in this paper, acknowledging that model implementation can be further optimized. For example, attention operations are originally slow due to quadratic complexity, and many recent works have successfully accelerated the speed of attention operations and reduced their memory usage. Hence, we believe there will be better implementations of these group operations with much greater efficiency in the future.

\begin{table}[H]
    \small
    \setlength\tabcolsep{3.5pt}
    \caption{Average inference time proportion of each DOFEN module across 59 medium-sized datasets.}
    \centering
    \begin{tabular}{l c c c}
    \toprule
        \textbf{Module Name} & \textbf{Source} & \begin{tabular}[c]{@{}c@{}}\textbf{Inference time proportion}\\\textbf{(mean)}\end{tabular} & \begin{tabular}[c]{@{}c@{}}\textbf{Inference time proportion}\\\textbf{(std)}\end{tabular} \\
    \midrule
        Condition Generation & \cref{fig:dofen-part-1}\subref{fig:condition-generation} & 7.03 \% & 5.29 \% \\
        Relaxed ODT Construction & \cref{fig:dofen-part-1}\subref{fig:rodt-construction} & 1.64 \% & 0.90 \% \\
        Forest Construction & \cref{fig:dofen-part-1}\subref{fig:forest-construction} and \cref{fig:dofen-part-2}\subref{fig:forest-construction-2} & 87.39 \% & 7.68 \% \\
        Forest Ensemble & \cref{fig:dofen-part-2}\subref{fig:forest-ensemble} & 3.94 \% & 2.36 \% \\
    \bottomrule
    \end{tabular}
\label{tab:module-inference-time}
\end{table}

\begin{table}[H]
    \small
    \setlength\tabcolsep{2.6pt}
    \caption{Average inference time proportion of each DOFEN operation across 59 medium-sized datasets.}
    \centering
    \begin{tabular}{l c c c}
    \toprule
        \textbf{Module Name} & \textbf{Operation} & \textbf{Source} & \begin{tabular}[c]{@{}c@{}}\textbf{Inference time}\\\textbf{ proportion (mean)}\end{tabular} \\
    \midrule
        Condition Generation & $\Delta_1$ & \cref{eq:condition-generation} & 7.03 \% \\
        Relaxed ODT Construction & permutation and reshape & \cref{eq:rodt-construction} & 1.64 \% \\
        Forest Construction & $\Delta_2$ & \cref{eq:derive-rODT-weight} & 85.11 \% \\
        ~ & get rODT embedding & \cref{eq:rodt-embedding} & 0.22 \% \\
        ~ & sample rODTs to form forests & \cref{algo:two-level-rodt-ensemble}, line 3 & 1.01 \% \\
        & softmax + weighted sum & \cref{algo:two-level-rodt-ensemble}, line 6 & 1.05 \% \\
        Forest Ensemble & $\Delta_3$ & \cref{algo:two-level-rodt-ensemble}, line 8 & 3.55 \% \\
        ~ & average forest predictions & \cref{algo:two-level-rodt-ensemble}, line 10 and 12 & 0.39 \% \\
    \bottomrule
    \end{tabular}
\label{tab:operation-inference-time}
\end{table}

\subsection{Training Time of DOFEN}
\label{sec:training-time-dofen}


To know more about how the slow inference time will affect the training time of DOFEN, we also conducted an experiment to compare the training time of DOFEN with other deep learning methods included in our paper (i.e. Trompt, FT Transformer, and NODE). We measured the training time on medium-sized datasets using both default and optimal hyperparameter settings, where the optimal hyperparameters  refers to the settings that deliver the best performance for each model on each dataset.

This experiment was conducted using a single NVIDIA Tesla V100 GPU. During model training, we carefully ensured that no other computational processes were running concurrently to enable a fair comparison. Additionally, we excluded datasets that would cause OOM (Out of Memory) issues during training, resulting in the selection of 50 out of 59 medium-sized datasets.

The average training time across datasets for each model is provided in Table A7. The results show that the training time for DOFEN is relatively long, approximately twice as long as Trompt when using optimal hyperparameters. This extended training time may be due to the inefficient group operations involved in DOFEN, which consume about 80\% of the computation time during the forward pass. For more details, please refer to \cref{sec:long-inference-time-dofen}. Therefore, improving the efficiency of group operations could reduce both the training and inference time of DOFEN.

\begin{table}[H]
    \small
    \caption{Average training time of different methods using default and optimal hyperparameter settings on 50 medium-sized datasets. Numbers are in Seconds, with lower values indicating faster training speed.}
    \centering
    \begin{tabular}{l c c}
    \toprule
         \textbf{Model Name} & \textbf{Training Time (Default)} & \textbf{Training Time (Optimal)} \\
    \midrule
        DOFEN & 332.6998 +/- 125.1965 & 1143.7674 +/- 804.3809 \\
        Trompt & 552.3495 +/- 213.3278 & 535.1781 +/- 291.9933 \\
        FT-Transformer & 80.3425 +/- 57.2647 & 99.1068 +/- 79.2272 \\
        NODE & 95.0274 +/- 54.7463 & 427.8625 +/- 394.1191 \\
    \bottomrule
    \end{tabular}
\label{tab:dofen-training-time}
\end{table}
\section{Scalability of DOFEN}
\label{sec:scalability-analysis}

To discuss the scalability of DOFEN, we have conducted experiments to investigate its performance given changes in hyperparameters $m$, $d$, and the number of MLP layers ($num\_layers$). In detail, changes in $m$ and $d$ affect the number of conditions ($N_\text{cond}$), while alterations in $m$ impact both the total number of rODTs ($N_\text{rODT}$) and the number of rODTs within an rODT forest ($N_\text{estimator}$). For further details on these parameters, please refer to \cref{tab:default-params}. The $num\_layers$ hyperparameter, newly introduced, refers to the number of MLP layers in neural networks $\Delta_1$, $\Delta_2$, and $\Delta_3$. A detailed introduction to $\Delta_1$, $\Delta_2$, and $\Delta_3$ can be found in \cref{sec:model-layers}.

Due to limited computational resources, we only conducted this experiment on datasets that would not cause out-of-memory (OOM) issues on our machine across all hyperparameter settings. This selection resulted in 51 out of 59 medium-sized datasets and 10 out of 14 large-sized datasets.

Based on \cref{tab:scalability-m-medium} to \cref{tab:scalability-num-layers-large}, we observed that larger values of $m$ and $d$ enhance DOFEN’s performance. Notably, improvements are more significant with large-sized datasets than with medium-sized datasets, likely because larger datasets benefit more from increased model capacity. In contrast, \cref{tab:scalability-num-layers-medium} reveals that an increase in $num\_layers$ generally results in poorer performance. This could be attributed to the substantial growth in parameter size and FLOPs, compared to adjustments in the $m$ and $d$, potentially leading to overfitting.

\begin{table}[H]
\centering
\small
\caption{Analysis of performance and efficiency across varied settings of $m$ on medium-sized datasets.}
\begin{tabular}{l c c c c c c c}
\toprule
& $m$ & \textbf{4} & \textbf{8} & \textbf{16 (default)} & \textbf{32} & \textbf{64} \\ \midrule
\multirow{2}{*}{Classification} & Performance (Accuracy) & 0.7491 & 0.7552 & 0.7602 & 0.7601 & \underline{0.7603} \\
& Parameters (M) & \underline{0.0029} & 0.0042 & 0.0070 & 0.0134 & 0.0296 \\
& FLOPs (M) & \underline{0.1797} & 0.1802 & 0.1815 & 0.1849 & 0.1951 \\
\midrule
\multirow{2}{*}{Regression} & Performance (R2 score) & 0.6496 & 0.6488 & 0.6796 & \underline{0.6940} & 0.6603 \\
& Parameters (M) & \underline{0.0026} & 0.0035 & 0.0056 & 0.0105 & 0.0235 \\
& FLOPs (M) & \underline{0.1783} & 0.1787 & 0.1797 & 0.1825 & 0.1912 \\
\bottomrule
\end{tabular}
\label{tab:scalability-m-medium}
\end{table}

\begin{table}[H]
\centering
\small
\caption{Analysis of performance and efficiency across varied settings of $m$ on large-sized datasets.}
\begin{tabular}{l c c c c c c c}
\toprule
& $m$ & \textbf{4} & \textbf{8} & \textbf{16 (default)} & \textbf{32} & \textbf{64} \\ \midrule
\multirow{2}{*}{Classification} & Performance (Accuracy) & 0.7498 & 0.7635 & 0.7800 & 0.7922 & \underline{0.8010} \\
& Parameters (M) & \underline{0.0033} & 0.0050 & 0.0084 & 0.0159 & 0.0333 \\
& FLOPs (M) & \underline{0.1798} & 0.1804 & 0.1819 & 0.1854 & 0.1949 \\
\midrule
\multirow{2}{*}{Regression} & Performance (R2 score) & 0.7227 & 0.7521 & 0.7583 & \underline{0.7698} & 0.7697 \\
& Parameters (M) & \underline{0.0025} & 0.0034 & 0.0058 & 0.0127 & 0.0350 \\
& FLOPs (M) & \underline{0.1783} & 0.1788 & 0.1803 & 0.1856 & 0.2045 \\
\bottomrule
\end{tabular}
\label{tab:scalability-m-large}
\end{table}

\begin{table}[H]
\centering
\small
\caption{Analysis of performance and efficiency across varied settings of $d$ on medium-sized datasets.}
\begin{tabular}{l c c c c c c c}
\toprule
& $d$ & \textbf{2} & \textbf{3} & \textbf{4 (default)} & \textbf{6} & \textbf{8} \\ \midrule
\multirow{2}{*}{Classification} & Performance (Accuracy) & 0.7402 & 0.7588 & \underline{0.7602} & 0.7583 & 0.7545 \\
& Parameters (M) & \underline{0.0058} & 0.0064 & 0.0070 & 0.0087 & 0.0108 \\
& FLOPs (M) & \underline{0.1801} & 0.1807 & 0.1815 & 0.1834 & 0.1857 \\
\midrule
\multirow{2}{*}{Regression} & Performance (R2 score) & 0.5961 & 0.6699 & 0.6796 & 0.6111 & \underline{0.6914} \\
& Parameters (M) & \underline{0.0047} & 0.0051 & 0.0056 & 0.0069 & 0.0087 \\
& FLOPs (M) & \underline{0.1786} & 0.1791 & 0.1797 & 0.1812 & 0.1831 \\
\bottomrule
\end{tabular}
\label{tab:scalability-d-medium}
\end{table}

\begin{table}[H]
\centering
\small
\caption{Analysis of performance and efficiency across varied settings of $d$ on large-sized datasets.}
\begin{tabular}{l c c c c c c c}
\toprule
& $d$ & \textbf{2} & \textbf{3} & \textbf{4 (default)} & \textbf{6} & \textbf{8} \\ \midrule
\multirow{2}{*}{Classification} & Performance (Accuracy) & 0.7433 & 0.7726 & 0.7800 & 0.7853 & \underline{0.7916} \\
& Parameters (M) & \underline{0.0071} & 0.0077 & 0.0084 & 0.0102 & 0.0125 \\
& FLOPs (M) & \underline{0.1803} & 0.1810 & 0.1819 & 0.1840 & 0.1865 \\
\midrule
\multirow{2}{*}{Regression} & Performance (R2 score) & 0.6572 & 0.7443 & 0.7583 & 0.7694 & \underline{0.7704} \\
& Parameters (M) & \underline{0.0043} & 0.0050 & 0.0058 & 0.0082 & 0.0113 \\
& FLOPs (M) & \underline{0.1787} & 0.1794 & 0.1803 & 0.1828 & 0.1860 \\
\bottomrule
\end{tabular}
\label{tab:scalability-d-large}
\end{table}

\begin{table}[H]
\centering
\small
\caption{Analysis of performance and efficiency across varied settings of $num\_layers$ on medium-sized datasets.}
\begin{tabular}{l c c c c c}
\toprule
& $num\_layers$ & \textbf{Default (1, 2, 2)} & \textbf{Twice (2, 4, 4)} & \textbf{Triple (3, 6, 6)} \\ \midrule
\multirow{2}{*}{Classification} & Performance (Accuracy) & \underline{0.7602} & 0.7592 & 0.7481 \\
& Parameters (M) & \underline{0.0070} & 0.0189 & 0.0308 \\
& FLOPs (M) & \underline{0.1815} & 0.5311 & 0.8808 \\
\midrule
\multirow{2}{*}{Regression} & Performance (R2 score) & 0.6796 & 0.6595 & \underline{0.7731} \\
& Parameters (M) & \underline{0.0056} & 0.0150 & 0.0245 \\
& FLOPs (M) & \underline{0.1797} & 0.5267 & 0.8737 \\
\bottomrule
\end{tabular}
\label{tab:scalability-num-layers-medium}
\end{table}

\begin{table}[H]
\centering
\small
\caption{Analysis of performance and efficiency across varied settings of $num\_layers$ on large-sized datasets.}
\begin{tabular}{l c c c c c}
\toprule
& $num\_layers$ & \textbf{Default (1, 2, 2)} & \textbf{Twice (2, 4, 4)} & \textbf{Triple (3, 6, 6)} \\ \midrule
\multirow{2}{*}{Classification} & Performance (Accuracy) & 0.7800 & \underline{0.7959} & 0.7638 \\
& Parameters (M) & \underline{0.0084} & 0.0231 & 0.0379 \\
& FLOPs (M) & \underline{0.1819} & 0.5346 & 0.8873 \\
\midrule
\multirow{2}{*}{Regression} & Performance (R2 score) & \underline{0.7583} & 0.7575 & 0.6715 \\
& Parameters (M) & \underline{0.0058} & 0.0139 & 0.0220 \\
& FLOPs (M) & \underline{0.1803} & 0.5259 & 0.8715 \\
\bottomrule
\end{tabular}
\label{tab:scalability-num-layers-large}
\end{table}
\section{More Interpretability Result}
\label{sec:more-interpretability}

This section provide results of DOFEN's feature importance on wine quality datasets mentioned in \cref{sec:interpretability}, detailed in \cref{tab:importance-red-wine,tab:importance-white-wine}. 

The results are similar to the one observed on mushroom dataset in \cref{tab:importance-mushroom}. DOFEN scores high feature importance to the features where tree-based models also considered important. This further indicates that DOFEN may contain similar decision-making process as tree-based model does, as it is a tree-inspired deep neural network.

\begin{table}[H]
    \small
    \caption{Feature importance of DOFEN on red wine dataset.}
    \centering
    \begin{tabular}{l c c c}
    \toprule
        ~ & \textbf{1st} & \textbf{2nd} & \textbf{3rd} \\
    \midrule
        Random Forest & alcohol (27.17\%) & sulphates (15.44\%) & volatile acidity (10.92\%) \\
        XGBoost & alcohol (35.42\%) & sulphates (15.44\%) & volatile acidity (7.56\%) \\
        LightGBM & alcohol (26.08\%) & sulphates (15.75\%) & volatile acidity (10.63\%) \\
        CatBoost & sulphates (16.29\%) & alcohol (15.67\%) & volatile acidity (10.40\%) \\
        GradientBoostingTree & alcohol (26.27\%) & sulphates (16.24\%) & volatile acidity (11.12\%) \\
        Trompt & alcohol (11.83\%) & sulphates (10.94\%) & total sulfur dioxide (9.78\%) \\
        DOFEN (ours) & alcohol (11.16\%) & volatile acidity (10.77\%) & sulphates (10.17\%) \\ 
    \bottomrule
    \end{tabular}
    \label{tab:importance-red-wine}
\end{table}

\begin{table}[H]
    \small
    \setlength\tabcolsep{3.0pt}
    \caption{Feature importance of DOFEN on white wine dataset.}
    \centering
    \begin{tabular}{l c c c}
    \toprule
        ~ & \textbf{1st} & \textbf{2nd} & \textbf{3rd} \\
    \midrule
        Random Forest & alcohol (24.22\%) & volatile acidity (12.44\%) & free sulfur dioxide (11.78\%) \\
        XGBoost & alcohol (31.87\%) & free sulfur dioxide (11.38\%) & volatile acidity (10.05\%) \\
        LightGBM & alcohol (24.02\%) & volatile acidity (12.47\%) & free sulfur dioxide (11.45\%) \\
        CatBoost & alcohol (17.34\%) & volatile acidity (12.07\%) & free sulfur dioxide (11.47\%) \\
        GradientBoostingTree & alcohol (27.84\%) & volatile acidity (13.59\%) & free sulfur dioxide (12.87\%) \\
        Trompt & fixed acidity (10.91\%) & volatile acidity (10.47\%) & pH (10.37\%) \\
        DOFEN (ours) & alcohol (10.90\%) & free sulfur dioxide (10.21\%) & volatile acidity (10.01\%) \\
    \bottomrule
    \end{tabular}
    \label{tab:importance-white-wine}
\end{table}
\section{More Analysis}
\label{sec:more-analysis}

\subsection{Overfitting arises without applying Forest Ensemble}
\label{sec:overfitting-without-sampling}

This section provides detailed experiment results of \cref{sec:ablation}, where we mentioned applying forest ensemble helps mitigate the overfitting issue of DOFEN.

\begin{figure}[H]
        \centering
        \includegraphics[width=\linewidth]{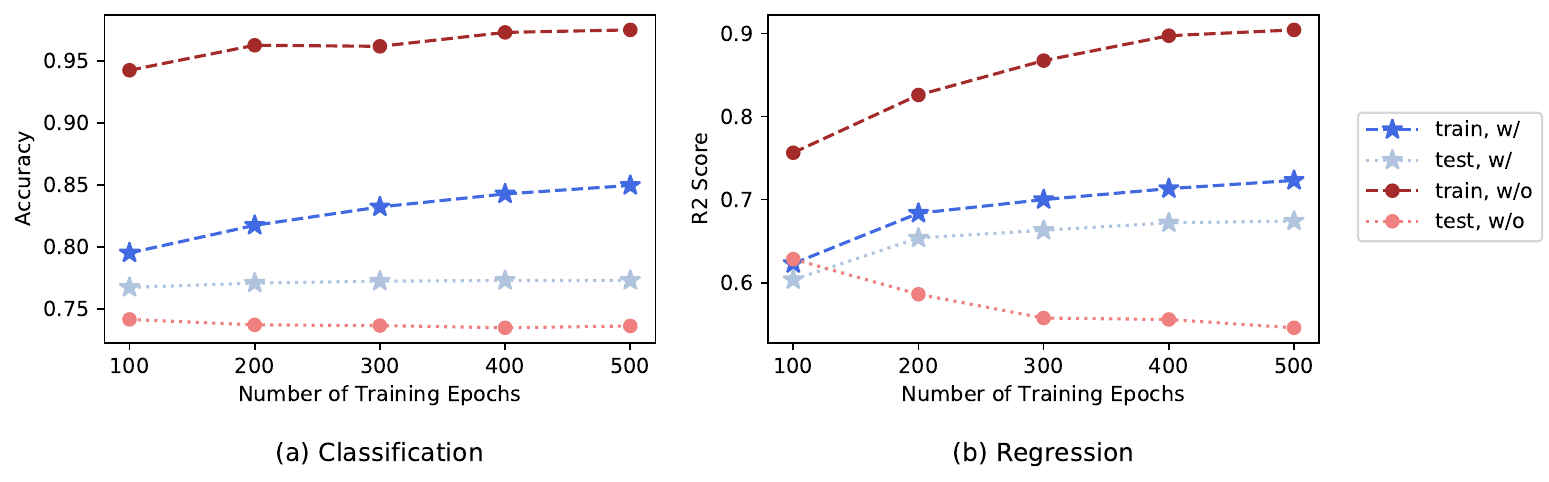}
        \caption{Overfitting arises when not applying sampling in the forest ensemble, affecting both (a) classification and (b) regression tasks. ``Train" refers to training performance, and ``Test" refers to testing performance.``w/" indicates the use of applying sampling to construct multiple forests, while ``w/o" indicates the use of all constructed rODTs to form a single forest.}
        \label{fig:overfitting}
\end{figure}

To further investigate the significant drop in performance without applying the sampling process during forest ensemble, we analyzed model performance at various training checkpoints.
As illustrated in \cref{fig:overfitting}, omitting sampling in the forest ensembles leads to better training performance but significantly worse testing performance, with the gap widening as training epochs increase, indicating the overfitting issue. 
In contrast, using an ensemble of multiple forests improves both training and testing performance, mitigating overfitting.

\subsection{Weights of Individual Relaxed ODT}
\label{sec:analysis-weight}

In DOFEN, large amount of rODTs are constructed by randomly combining a set of generated conditions for each column. This step aims for increasing the diversity of rODTs to benefit the later bagging process on the predictions of multiple rODT forests, as rODT forests are also constructed by randomly aggregating rODTs.

\begin{figure}[H]
    \centering
    \begin{subfigure}{.49\columnwidth}
        \centering
        \includegraphics[width=\linewidth]{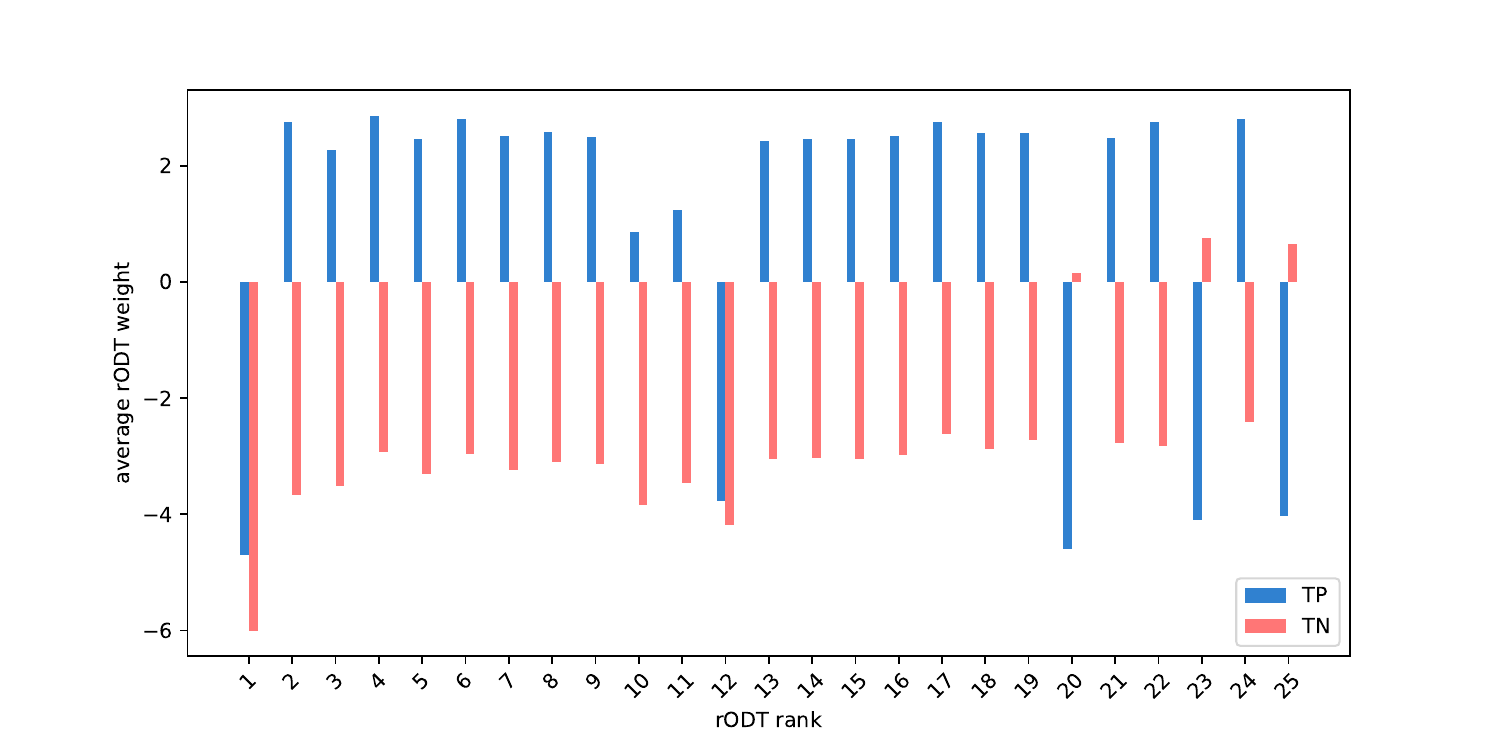}
        \caption{Relaxed ODTs with large weight variation.}
        \label{fig:weight-head-covertype}
    \end{subfigure}
    \begin{subfigure}{.49\columnwidth}
        \centering
        \includegraphics[width=\linewidth]{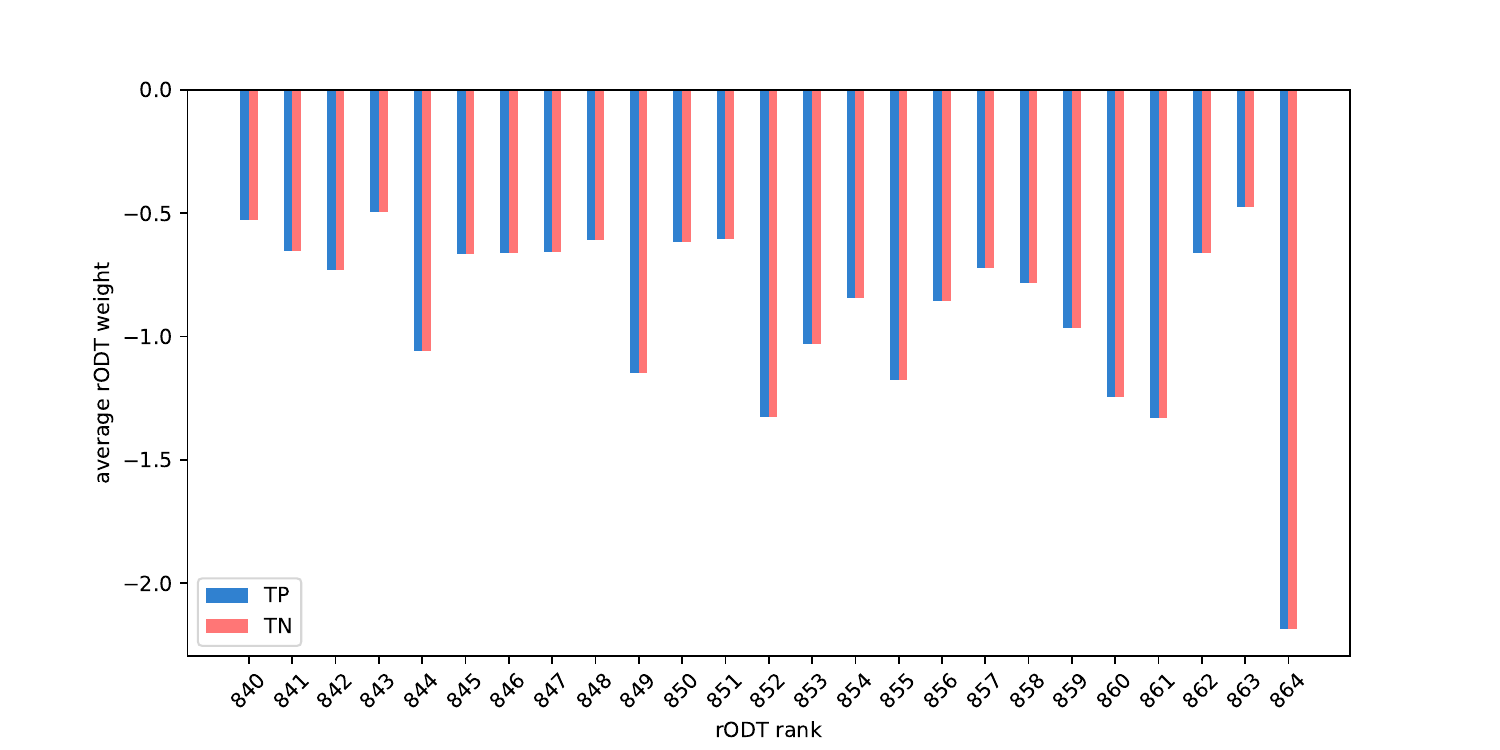}
        \caption{Relaxed ODTs with small weight variation.}
        \label{fig:weight-tail-covertype}
    \end{subfigure}
    \caption{In the covertype dataset, \cref{fig:weight-head-covertype} shows that the average weights of true positives differ significantly from those of true negatives.
    Conversely, \cref{fig:weight-tail-covertype} reveals a contrasting result for rODTs with small weight variation.
    }
    \label{fig:weight-covertype}
\end{figure}

\begin{figure}[H]
    \centering
    \begin{subfigure}[t]{.5\columnwidth}
        \centering
        \includegraphics[width=\linewidth]{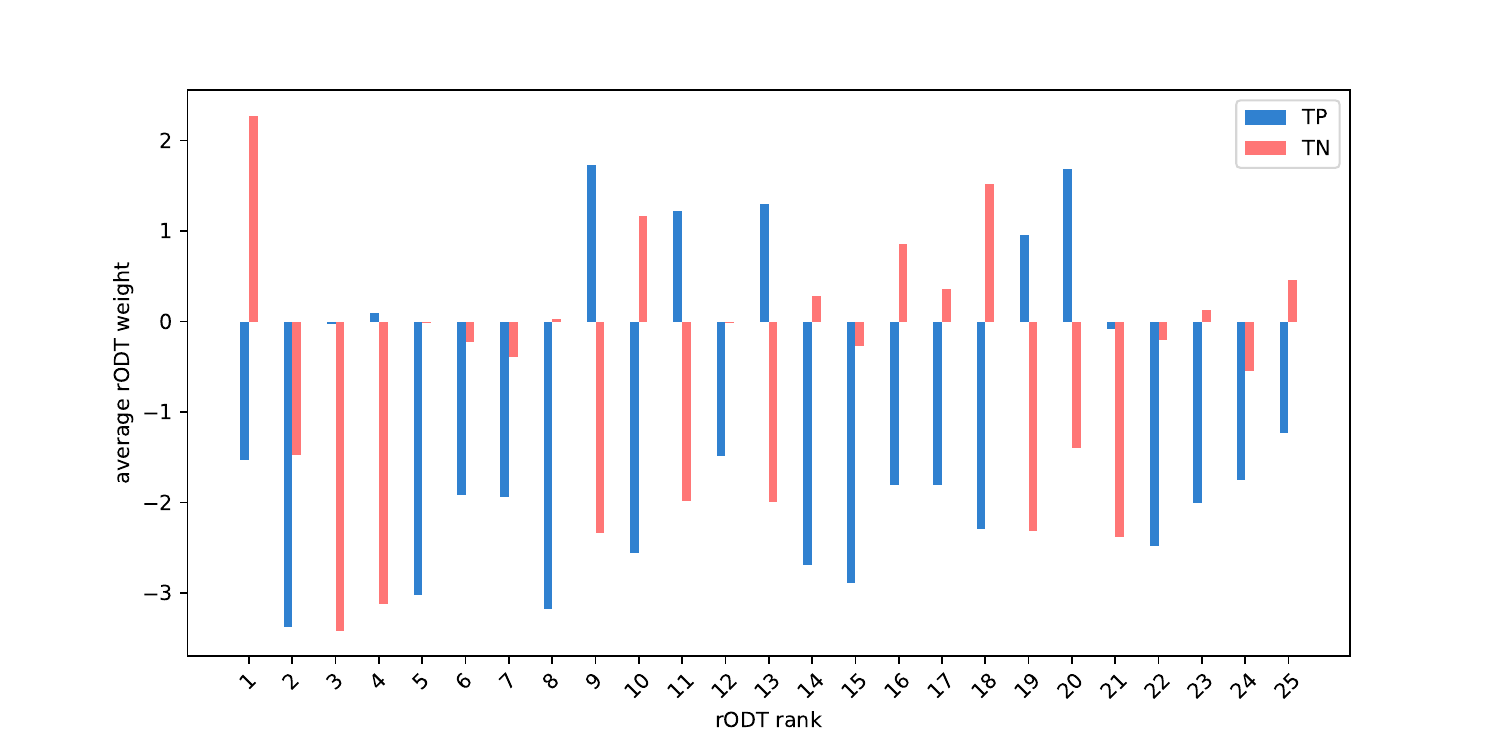}
        \caption{Relaxed ODTs with large weight variation.}
        \label{fig:weight-head-compass}
    \end{subfigure}%
    \begin{subfigure}[t]{.5\columnwidth}
        \centering
        \includegraphics[width=\linewidth]{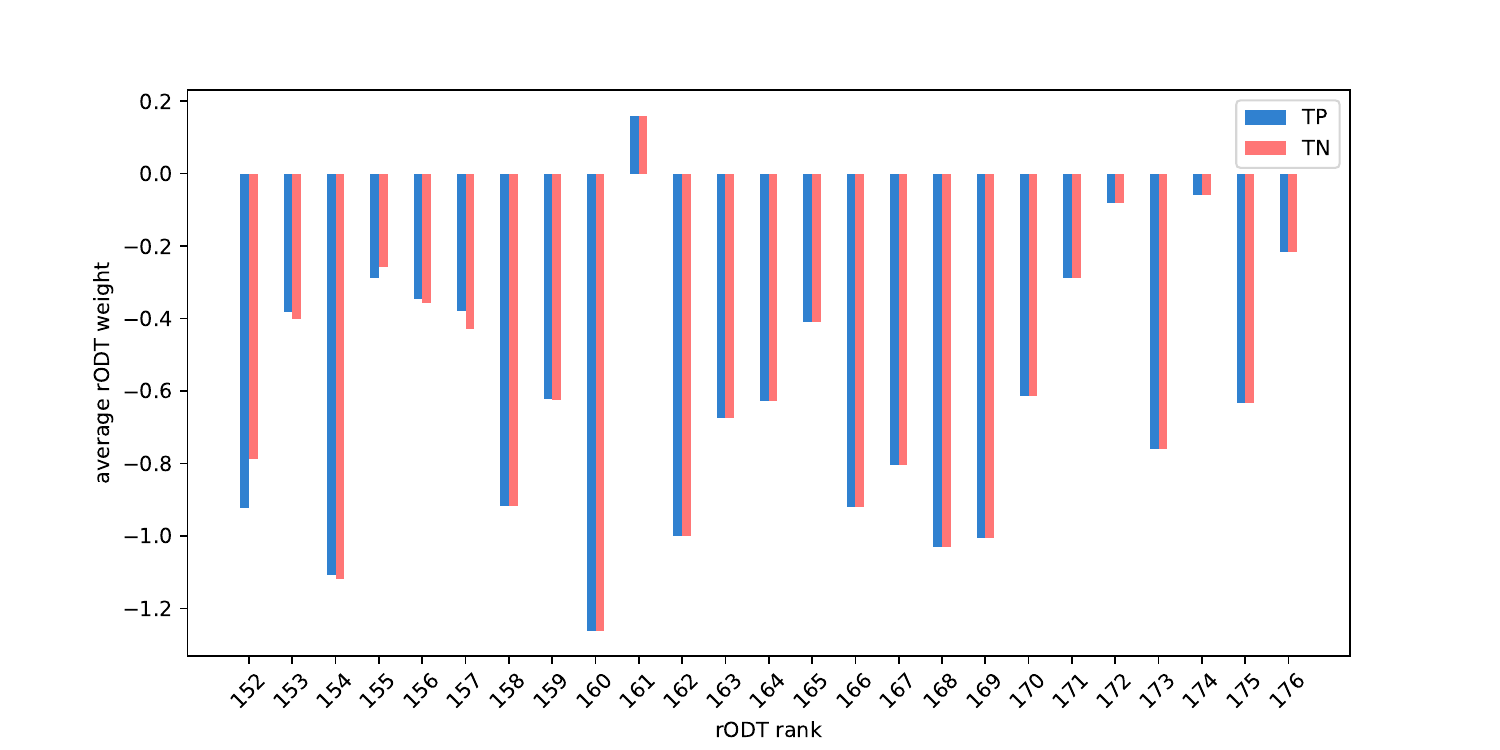}
        \caption{Relaxed ODTs with small weight variation.}
        \label{fig:weight-tail-compass}
    \end{subfigure}
    \caption{In the compass dataset, the weights $w_i$ of rODT are sorted based on the standard deviation calculated across true positive (TP) and true negative (TN) samples in the testing data.
    \cref{fig:weight-head-compass} shows that the weights of TP samples differ significantly from those of TN samples when the standard deviation of the weights is higher.
    Conversely, \cref{fig:weight-tail-compass} reveals contrasting results for weights with a lower standard deviation.
    }
    \label{fig:weight-compass}
\end{figure}

This means an rODT will affect the performance of an rODT forest, and as the process of constructing rODTs involves randomize operations, there might exist redundant rODTs that are not important for corresponding task.
To check whether if redundant rODTs exist, we analyze two binary classification dataset (covertype) to observe the variation in the weights assigned to individual rODTs across different samples, as shown in \cref{fig:weight-covertype,fig:weight-compass}.

\cref{fig:weight-head-covertype} shows that, for most rODTs ranked in the top 25 according to their weight standard deviations, there is a significant difference between the average weights of true positive samples and those of true negative samples.
Conversely, \cref{fig:weight-tail-covertype} shows an opposite trend for rODTs with the smallest weight standard deviations.
These trends are also observed in another dataset, as shown in \cref{fig:weight-head-compass,fig:weight-tail-compass}.
These observations imply that rODTs with larger weight standard deviations play a more crucial role in classifying samples, while rODTs with less weight standard deviations are not sensitive to samples with different label.

In addition, we come up with an idea to examine the performance change after pruning rODT weights with small standard deviations across samples and their corresponding rODT embeddings, seeing if these rODTs can be considered as redundant rODTs.
The results are provided in \cref{sec:more-analysis-prune} and suggest that the variation serves as a reliable indicator of the importance of rODTs. 
Moreover, pruning the less important rODTs not only enhances the model's efficiency but also its performance.


\subsection{Pruning of Relaxed ODT}
\label{sec:more-analysis-prune}

Following \cref{sec:analysis-weight}, in this section, we aim to examine the performance change after pruning rODT weights with small standard deviations and their corresponding rODT embeddings.

\cref{tab:prune-tail} shows the performance under different pruning ratios.
The column labeled 'by dataset' indicates that we tailored the pruning ratio for each dataset based on its validation data.
As shown in \cref{tab:prune-tail}, pruning these rODTs does not negatively affect performance.
In fact, a minor degree of pruning can actually enhance performance, with the optimal pruning ratio being 0.02 for classification datasets and 0.1 for regression datasets.
Notice that the 'by dataset' approach is better suited to real-world scenarios, even though it does not always yield the best performance.

\begin{table}[H]
\centering
\small
\caption{Pruning of rODT with varying ratio.
Weights $w_i$ with \emph{lower} standard deviation are pruned. \newline}
\begin{tabular}{l c c c c c c}
\toprule
Ratio & \textbf{\makecell{0.0 (default)}} & \textbf{0.02} & \textbf{0.1} & \textbf{0.2} & \textbf{by dataset} \\ \midrule
Classification & 0.7725 & \underline{0.7733} & 0.7726 & 0.7709 & 0.7732 \\
Regression & 0.6605 & 0.6629 & 0.6630 & 0.6621 & \underline{0.6657} \\
\bottomrule
\end{tabular}
\label{tab:prune-tail}
\end{table}


We then investigate the outcomes when weights with higher standard deviations are pruned. 
Consequently, we sort the weights and prune them from the higher end.
The results, presented in \cref{tab:prune-head}, show that the performance in both classification and regression tasks monotonically drops as the prune ratio increases.
This finding suggests that the standard deviation of weights is a good indicator of their importance in making predictions.
It further validates why pruning weights with lower standard deviation does not harm performance and, in some cases, even helps.

\begin{table}[H]
\centering
\small
\caption{rODT pruning with varying ratio.
Weights $w_i$ with \emph{higher} standard deviation are pruned.}
\begin{tabular}{l c c c c c c}
\toprule
Ratio & \textbf{0.0(default)} & \textbf{0.02} & \textbf{0.05} & \textbf{0.10} & \textbf{0.2} \\ \midrule
Classification & \underline{0.7725} & \underline{0.7725} & 0.7715 & 0.7667 & 0.763 \\
Regression & \underline{0.6605} & 0.6571 & 0.6484 & 0.6383 & 0.601 \\
\bottomrule
\end{tabular}
\label{tab:prune-head}
\end{table}

In addition, we discuss another, potentially more straightforward, pruning approach.
Specifically, we prune the weights $w_i$ based on their average value across samples.
Similar to the experiments that use standard deviation as the metric for pruning, this time we sort the weights by their average.
We then attempt to prune the weights from both the top and bottom ends.
The results are provided in \cref{tab:prune-tail-mean} and \cref{tab:prune-head-mean}, suggesting that the value of weights is not an effective indicator for pruning.
Although there is some improvement in performance at a low ratio, this approach generally diminishes performance with larger ratios, regardless of whether the weights are pruned from the higher or lower end.

\begin{table}[H]
\centering
\small
\caption{rODT pruning with varying ratio.
Weights $w_i$ with \emph{lower} average value are pruned.}
\begin{tabular}{l c c c c c c}
\toprule
Ratio & \textbf{0.0(default)} & \textbf{0.02} & \textbf{0.05} & \textbf{0.10} & \textbf{0.2} \\ \midrule
Classification & 0.7725 & \underline{0.773} & 0.7715 & 0.7722 & 0.7703 \\
Regression & 0.6605 & \underline{0.6611} & 0.6592 & 0.6575 & 0.6425 \\
\bottomrule
\end{tabular}
\label{tab:prune-tail-mean}
\end{table}

\begin{table}[H]
\centering
\small
\caption{rODT pruning with varying ratio.
Weights $w_i$ with \emph{higher} average value are pruned.}
\begin{tabular}{l c c c c c c}
\toprule
Ratio & \textbf{0.0(default)} & \textbf{0.02} & \textbf{0.05} & \textbf{0.10} & \textbf{0.2} \\ \midrule
Classification & 0.7725 & \underline{0.7731} & 0.7725 & 0.7704 & 0.7643 \\
Regression & 0.6605 & \underline{0.6619} & 0.6573 & 0.6352 & 0.4881 \\
\bottomrule
\end{tabular}
\label{tab:prune-head-mean}
\end{table}
\section{More Evaluation Results on Tabular Benchmark}

\subsection{Performance Evaluation on Large-sized Benchmark}
\label{sec:eval-large}


This section discusses the evaluation results on large-sized classification and regression tasks.
Overall, the results demonstrate a similar trend as the medium-sized tabular benchmark.
Notably, DOEFN achieves the top ranks in both tasks with numerical features. 


\begin{figure}[H]
    \centering
    \begin{subfigure}[t]{.5\columnwidth}
        \centering
        \includegraphics[width=.7\linewidth]{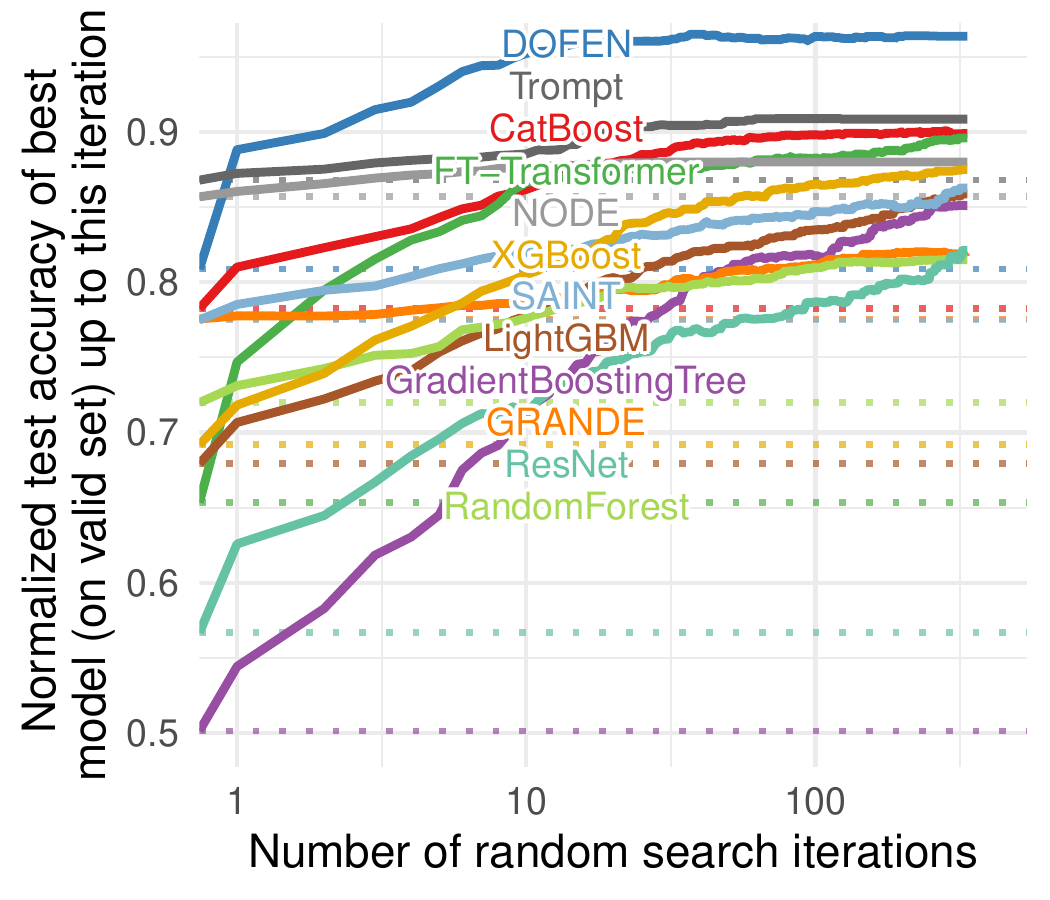}
        \caption{Exclusively Numerical}
        \label{fig:cls-eval-num-large}
    \end{subfigure}%
    \begin{subfigure}[t]{.5\columnwidth}
        \centering
        \includegraphics[width=.7\linewidth]{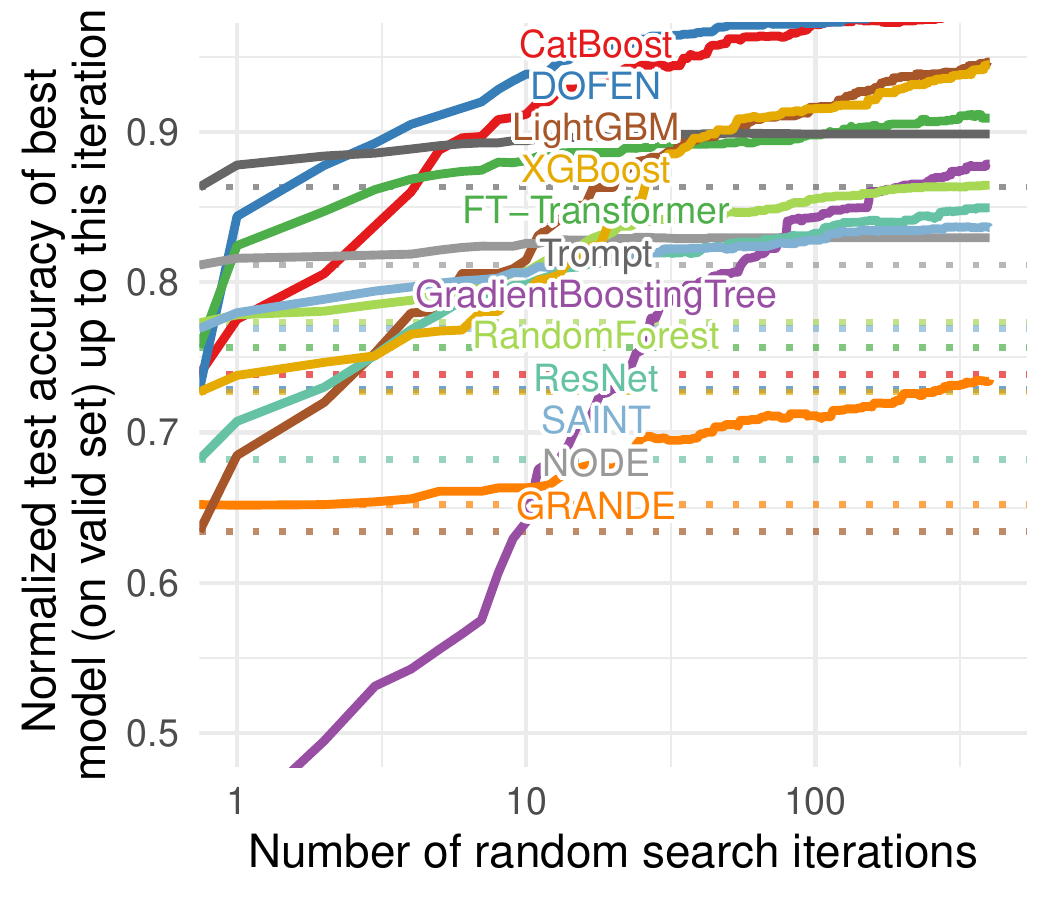}
        \caption{Heterogeneous}
        \label{fig:cls-eval-het-large}
    \end{subfigure}
    \caption{Results on \textbf{large-sized classification} datasets.}
    \label{fig:cls-eval-large}
\end{figure}

\textbf{Classification.}
In \cref{fig:cls-eval-num-large}, DOFEN even surpasses CatBoost to become the top performer. 
In \cref{fig:cls-eval-het-large}, DOFEN and CatBoost clearly outperforms other models.
In contrast to DOFEN, other tabular DNN models like FT-Transformer and Trompt rank in the middle among all models.
As a result, with the current development of tabular DNN models, their performance in processing numerical features is already on par with or even surpass that of tree-based models, and they are more advantageous for large-sized datasets.
However, DNN models are still less efficient in handling heterogeneous features.


\begin{figure}[H]
    \centering
    \begin{subfigure}[t]{.5\columnwidth}
        \centering
        \includegraphics[width=.7\linewidth]{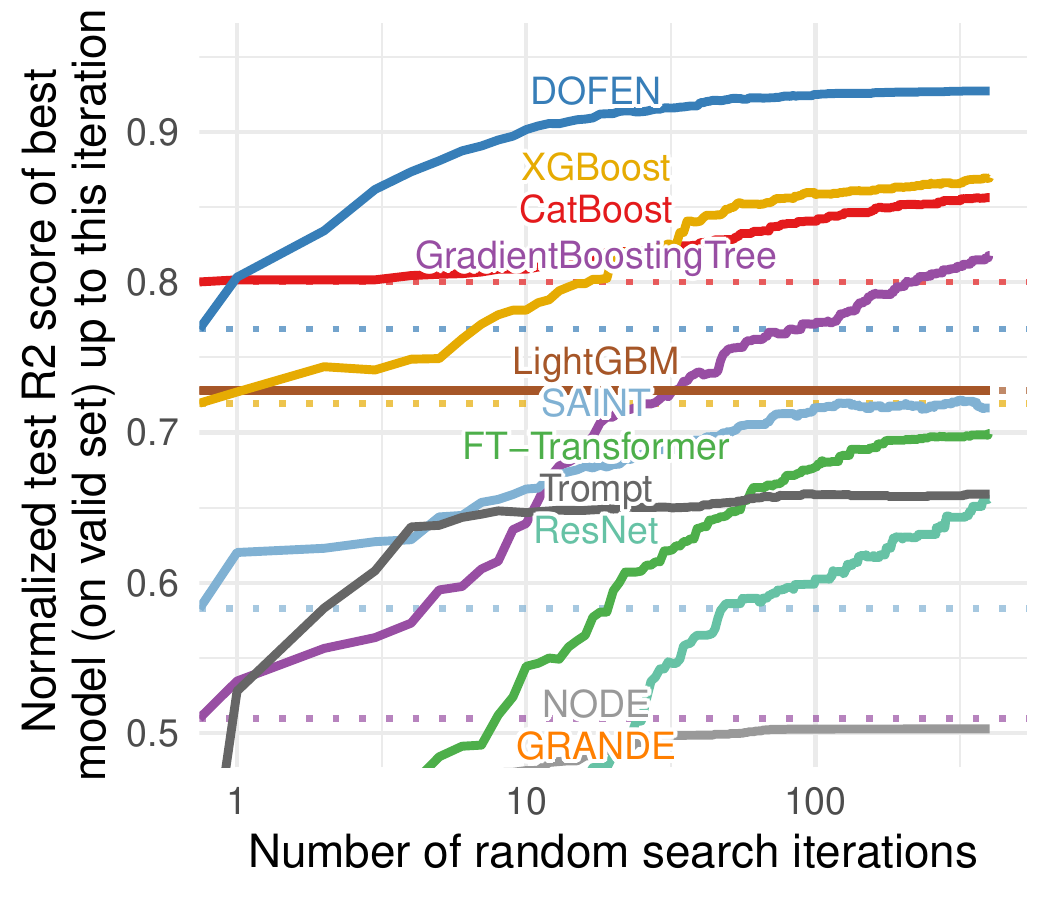}
        \caption{Large; Numerical}
        \label{fig:reg-eval-num-large}
    \end{subfigure}%
    \begin{subfigure}[t]{.5\columnwidth}
        \centering
        \includegraphics[width=.7\linewidth]{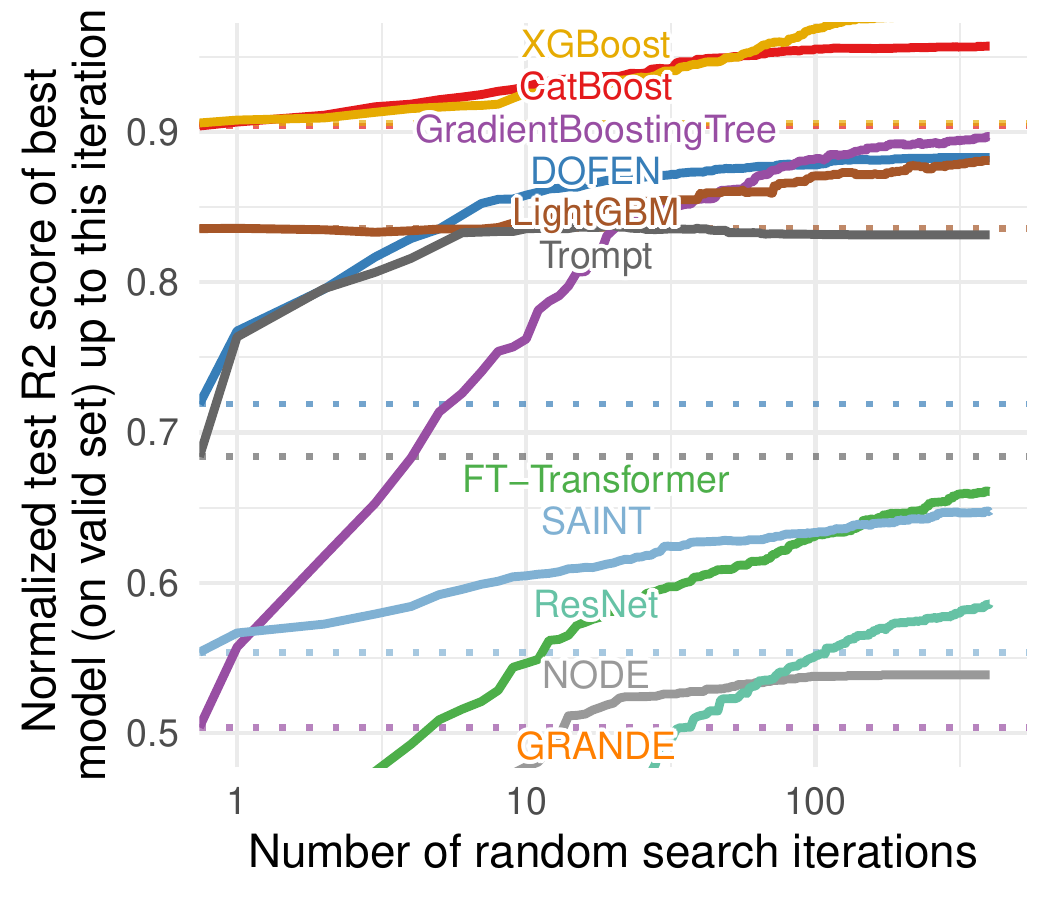}
        \caption{Large; Heterogeneous}
        \label{fig:reg-eval-het-large}
    \end{subfigure}
    \caption{Results on \textbf{large-sized regression} datasets.
    }
    \label{fig:reg-eval-large}
\end{figure}

\textbf{Regression.}
In \cref{fig:reg-eval-num-large}, the leading models remain DOFEN, XGBoost, and CatBoost.
DOFEN's proficiency in handling numerical features, further enhanced by the increased data volume, enables it to secure the top position once again.
In \cref{fig:reg-eval-het-large}, DOFEN and Trompt barely maintain their positions within the leading group, yet they still stand out from the other DNN models.

\subsection{Detailed Evaluation Results}
\label{sec:more-eval}
In the main paper, we have discussed the overall performance of DOFEN.
To simplify tables, we map dataset names with their OpenML ID, as described in \cref{sec:id-mapping}.
The evaluation results of each task are organized in \cref{tab:more-eval-organize}.
Please refer to the detailed figures and tables for each task of your interest.
The evaluation metrics are accuracy for classification tasks and $\mathrm{R}^2$ score for regression tasks, consistent with our main paper.
Furthermore, we calculate the mean and standard deviation of ranks across datasets to provide the rank for each model in the tables.

\begin{table}[H]
\centering
\small
\caption{Tables and figures for each task.}
\begin{tabular}{c c c c}
\toprule
\textbf{Task} & \textbf{Feature} & \textbf{Figure} & \textbf{Table} \\ 
\midrule
\multirow{2}{*}{medium-sized classification} & numerical & \cref{fig:medium-clf-num-dataset} & \cref{tab:performance-medium-num-clf-1,tab:performance-medium-num-clf-2} \\
& heterogeneous & \cref{fig:medium-clf-het-dataset} & \cref{tab:performance-medium-het-clf} \\
\midrule
\multirow{2}{*}{medium-sized regression} & numerical & \cref{fig:medium-reg-num-dataset} & \cref{tab:performance-medium-num-rgr-1,tab:performance-medium-num-rgr-2} \\
& heterogeneous & \cref{fig:medium-reg-het-dataset} & \cref{tab:performance-medium-het-rgr-1,tab:performance-medium-het-rgr-2} \\
\midrule
\multirow{2}{*}{large-sized classification} & numerical & \cref{fig:large-clf-num-dataset} & \cref{tab:performance-large-num-clf} \\
& heterogeneous & \cref{fig:large-clf-het-dataset} & \cref{tab:performance-large-het-clf} \\
\midrule
\multirow{2}{*}{large-sized regression} & numerical & \cref{fig:large-reg-num-dataset} & \cref{tab:performance-large-num-rgr} \\
& heterogeneous & \cref{fig:large-reg-het-dataset} & \cref{tab:performance-large-het-rgr} \\
\bottomrule
\end{tabular}
\label{tab:more-eval-organize}
\end{table}

\begin{figure}[H]
    \centering
    \includegraphics[width=\columnwidth]{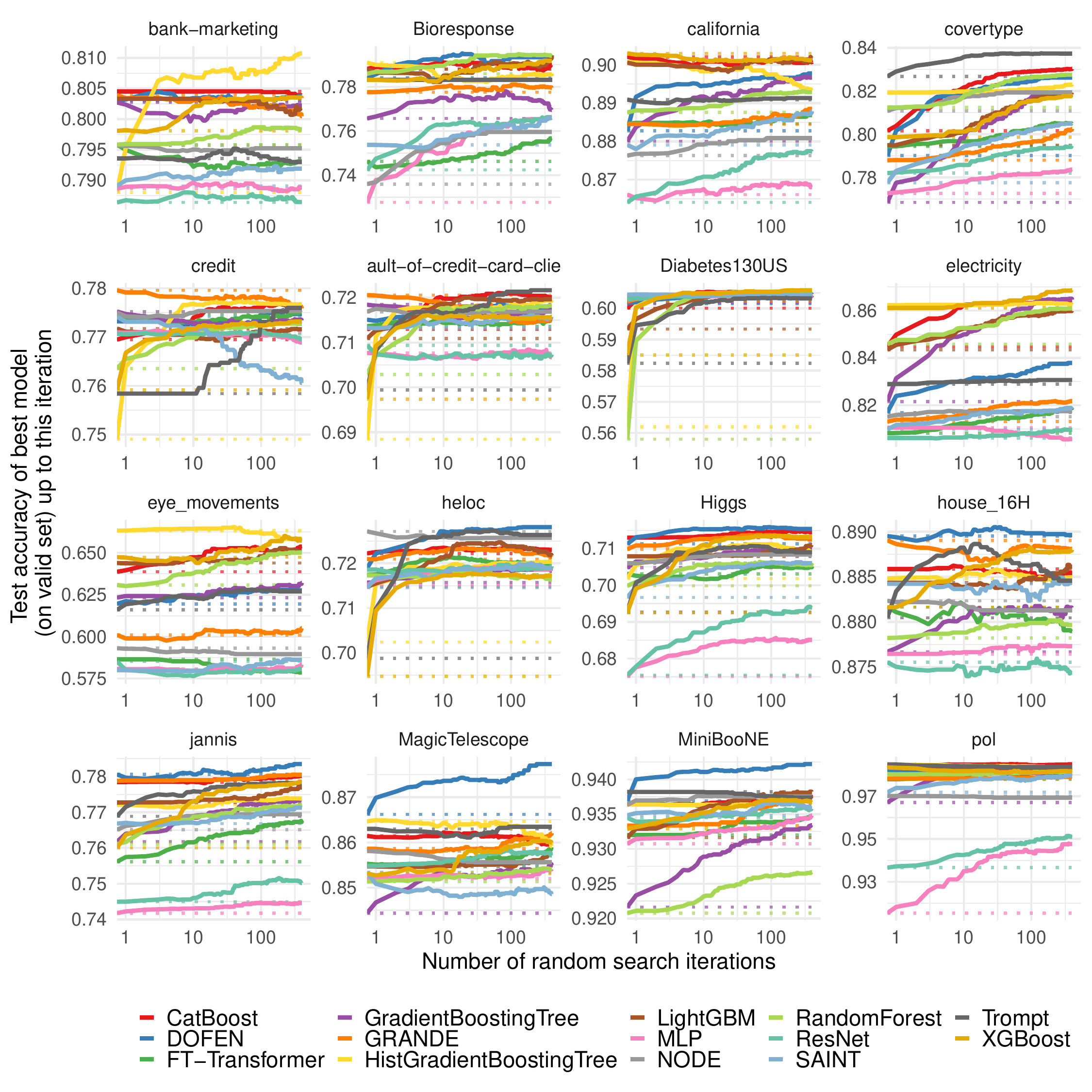}
    \caption{Results on \textit{each} \textbf{medium-sized classification} datasets with only \textbf{numerical} features.}
    \label{fig:medium-clf-num-dataset}
\end{figure}

\begin{figure}[H]
    \centering
    \includegraphics[width=\columnwidth]{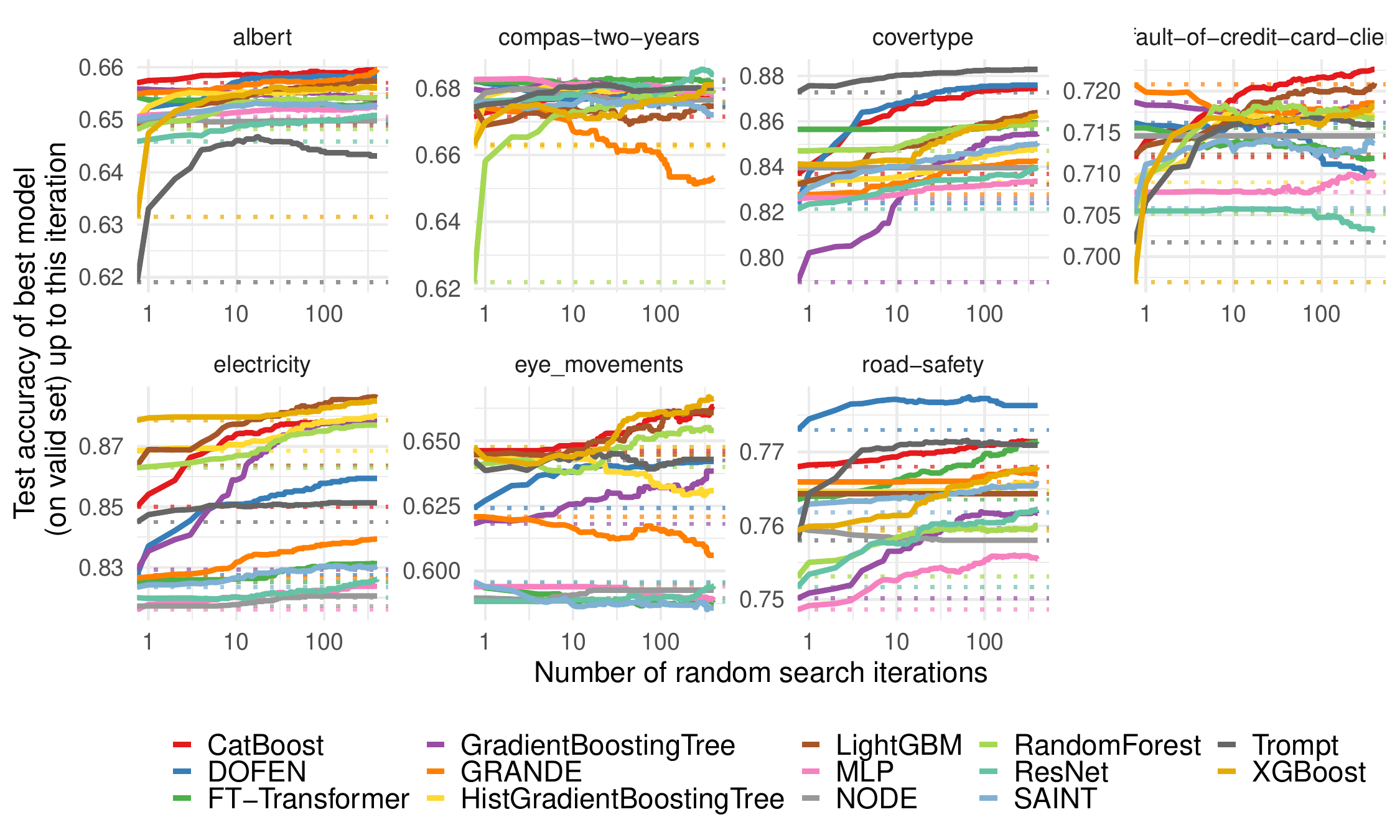}
    \caption{Results on \textit{each} \textbf{medium-sized classification} datasets with \textbf{heterogeneous} features.}
    \label{fig:medium-clf-het-dataset}
\end{figure}

\begin{figure}[H]
    \centering
    
    \includegraphics[width=\columnwidth]{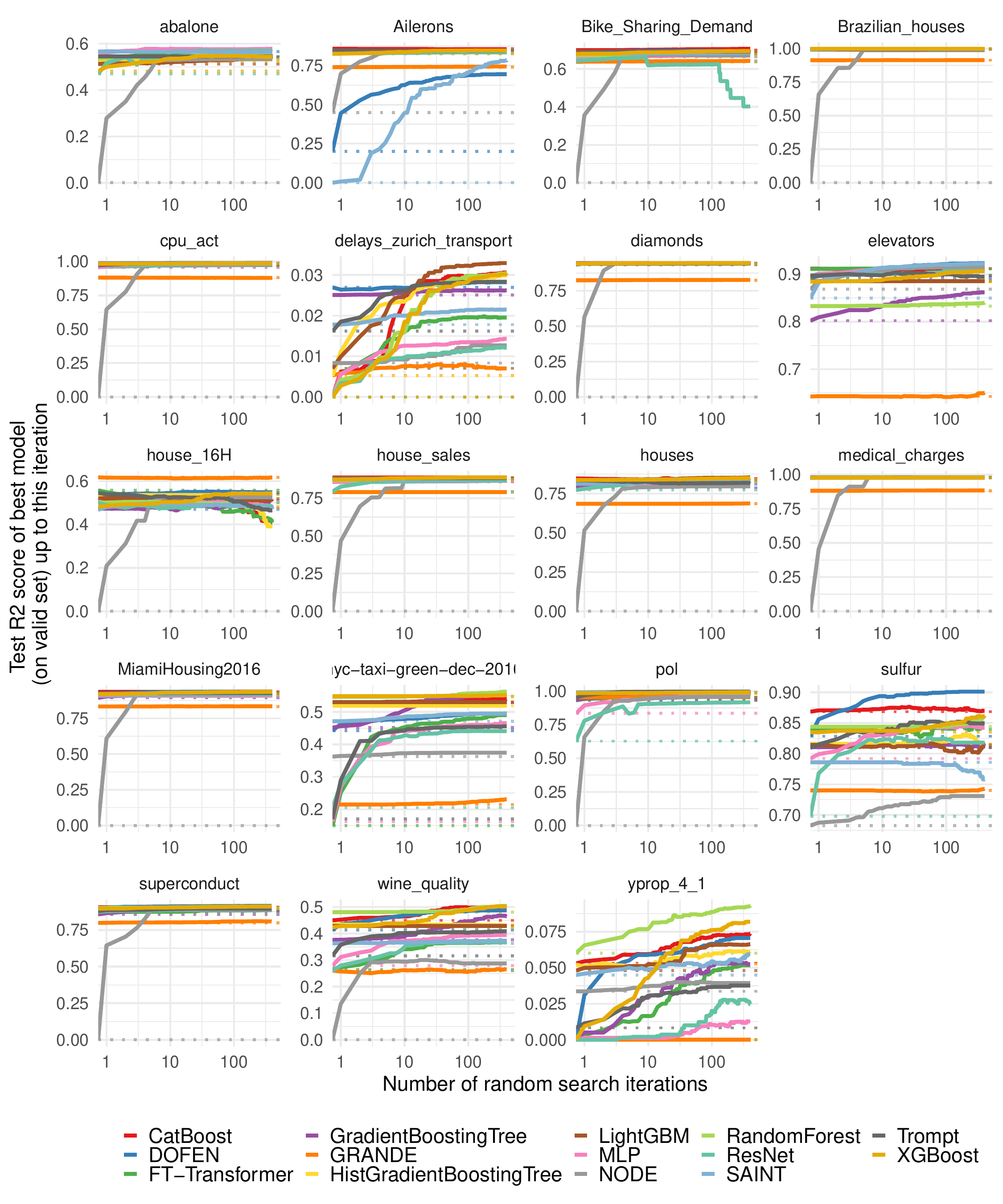}
    \caption{Results on \textit{each} \textbf{medium-sized regression} datasets with \textbf{numerical} features.}
    \label{fig:medium-reg-num-dataset}
\end{figure}

\begin{figure}[H]
    \centering
    \includegraphics[width=\columnwidth]{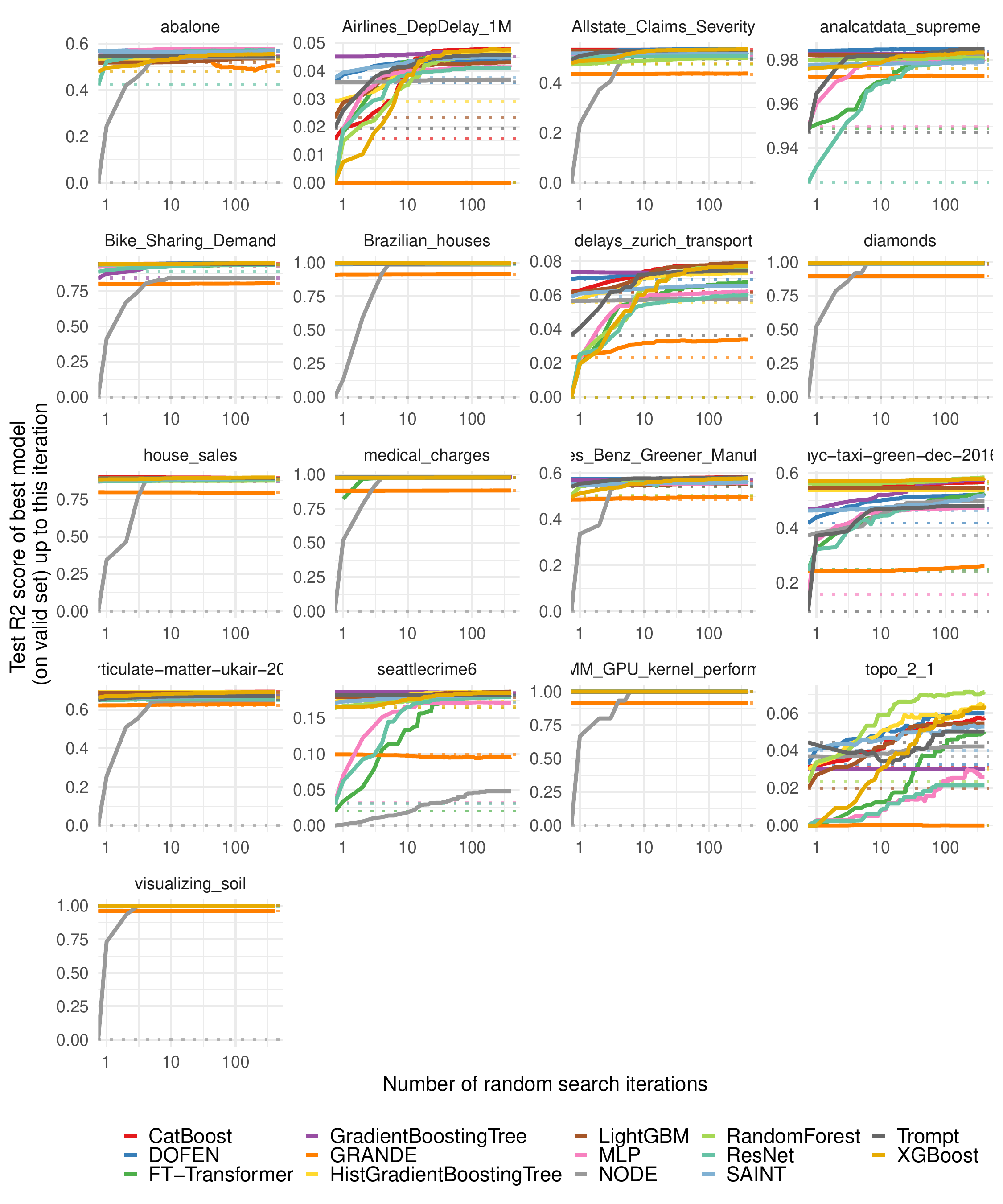}
    \caption{Results on \textit{each} \textbf{medium-sized regression} datasets with \textbf{heterogeneous} features.}
    \label{fig:medium-reg-het-dataset}
\end{figure}

\begin{figure}[H]
    \centering
    \includegraphics[width=\columnwidth]{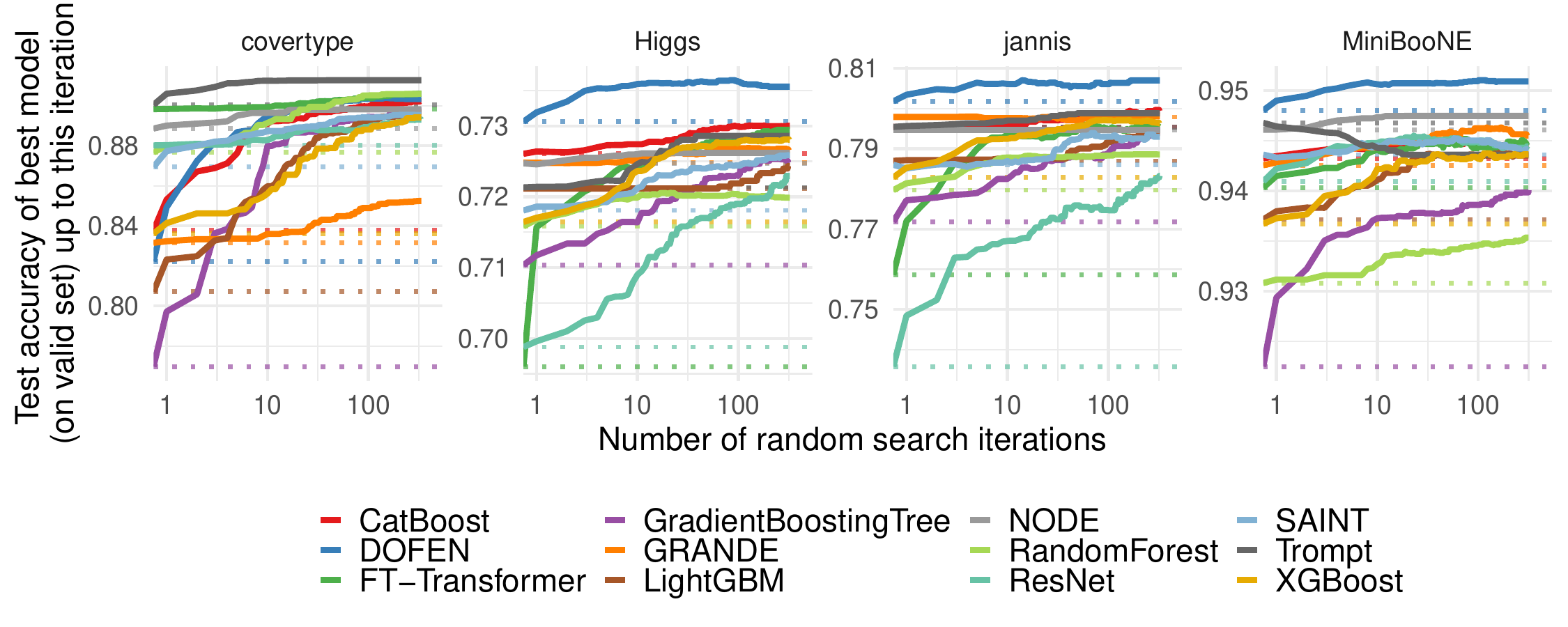}
    \caption{Results on \textit{each} \textbf{large-sized classification} datasets with only \textbf{numerical} features.}
    \label{fig:large-clf-num-dataset}
\end{figure}

\begin{figure}[H]
    \centering
    \includegraphics[width=0.6\columnwidth]{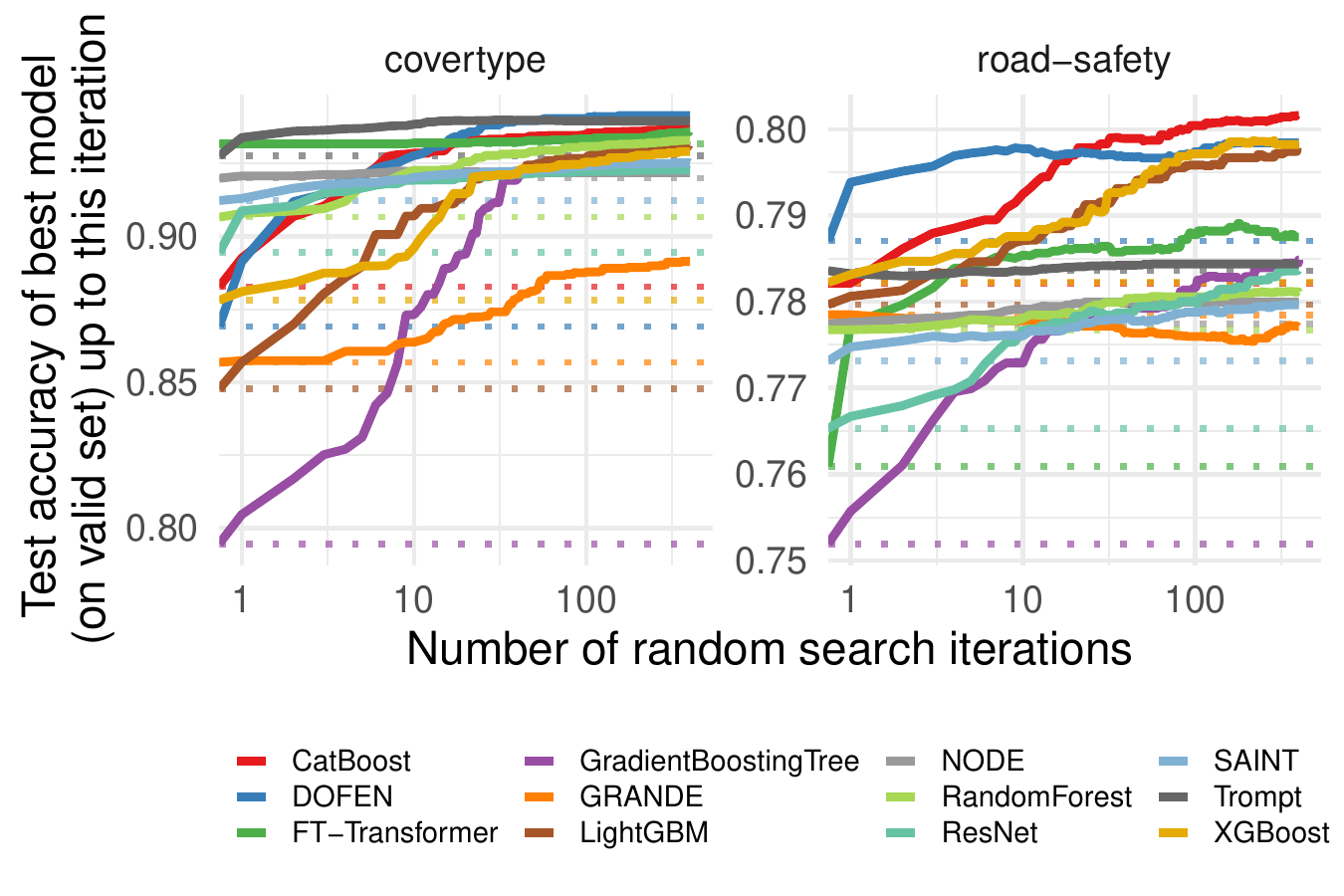}
    \caption{Results on \textit{each} \textbf{large-sized classification} datasets with \textbf{heterogeneous} features.}
    \label{fig:large-clf-het-dataset}
\end{figure}

\begin{figure}[H]
    \centering
    \includegraphics[width=\columnwidth]{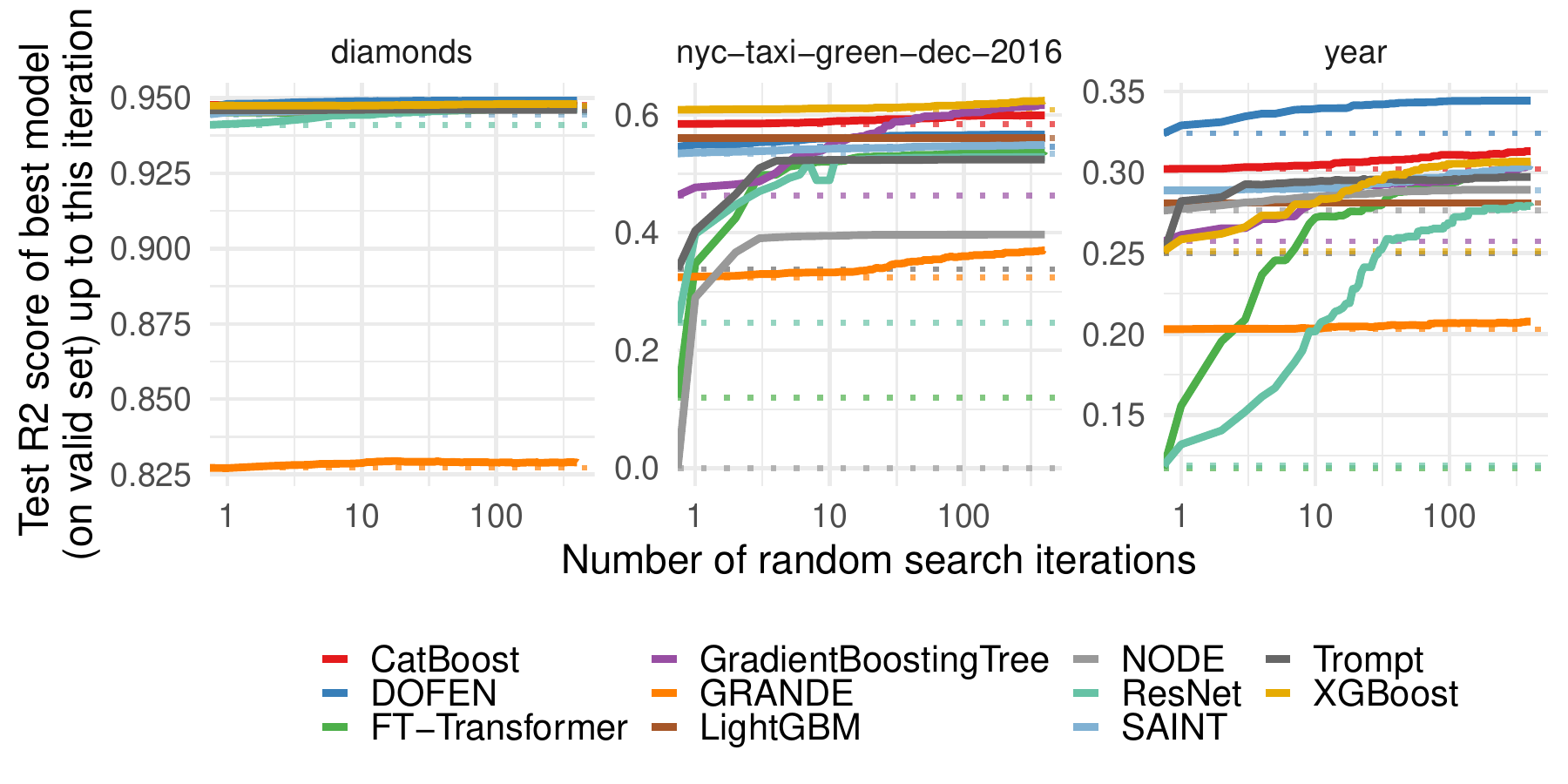}
    \caption{Results on \textit{each} \textbf{large-sized regression} datasets with \textbf{numerical} features.}
    \label{fig:large-reg-num-dataset}
\end{figure}

\begin{figure}[H]
    \centering
    \includegraphics[width=\columnwidth]{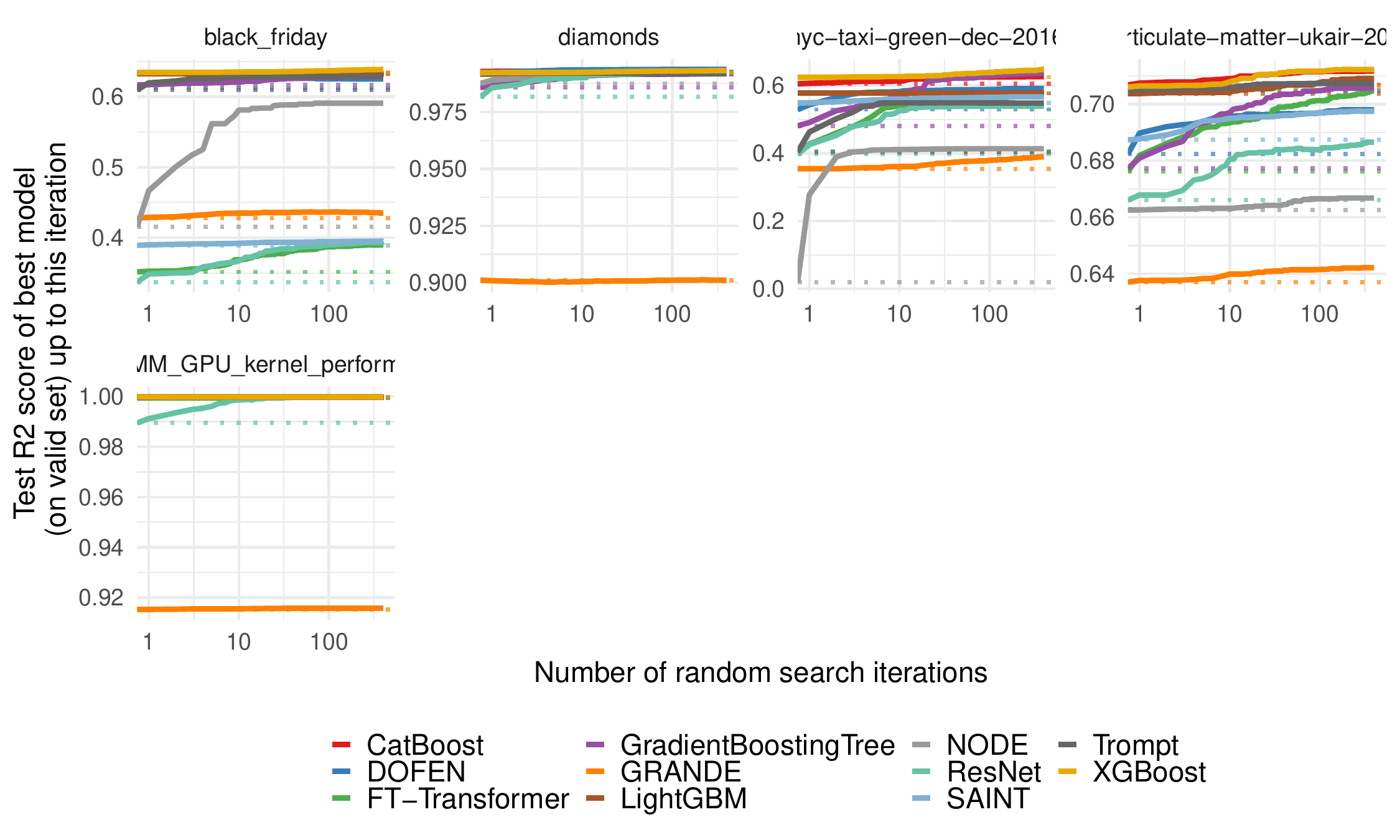}
    \caption{Results on \textit{each} \textbf{large-sized regression} datasets with \textbf{heterogeneous} features.}
    \label{fig:large-reg-het-dataset}
\end{figure}

\begin{table}[H]
\centering
\small
\setlength\tabcolsep{3.0pt}
\caption{The performance of \textbf{medium-sized classification} task (\emph{numerical features only}) (1).}
\label{tab:performance-medium-num-clf-1}
\begin{tabular}{l c c c c c c c c c c}
\toprule
 & \textbf{361276} & \textbf{361273} & \textbf{361069} & \textbf{361065} & \textbf{361068} & \textbf{361066} & \textbf{361277} & \textbf{361061} & \textbf{361055} & \textbf{361275} \\
\midrule
\multicolumn{11}{c}{Default} \\
\midrule
DOFEN (ours) & 0.7839 & 0.6016 & 0.7113 & \underline{0.8662} & 0.9369 & 0.8030 & 0.8827 & 0.7901 & 0.7732 & 0.7151 \\
Trompt & 0.7831 & 0.5823 & 0.6926 & 0.8630 & \underline{0.9382} & 0.7936 & 0.8909 & \underline{0.8268} & 0.7584 & 0.6994 \\
GRANDE & 0.7776 & 0.6023 & 0.7099 & 0.8586 & 0.9334 & 0.8039 & 0.8845 & 0.7880 & \underline{0.7796} & \underline{0.7206} \\
FT-Transformer & 0.7463 & 0.6025 & 0.7031 & 0.8553 & 0.9320 & 0.7958 & 0.8846 & 0.7944 & 0.7745 & 0.7137 \\
ResNet & 0.7424 & 0.6029 & 0.6755 & 0.8548 & 0.9345 & 0.7864 & 0.8641 & 0.7820 & 0.7706 & 0.7093 \\
MLP & 0.7277 & 0.6033 & 0.6752 & 0.8520 & 0.9307 & 0.7886 & 0.8661 & 0.7727 & 0.7710 & 0.7077 \\
SAINT & 0.7537 & \underline{0.6044} & 0.6967 & 0.8534 & 0.9348 & 0.7891 & 0.8791 & 0.7775 & 0.7741 & 0.7133 \\
NODE & 0.7360 & 0.6039 & 0.7060 & 0.8581 & 0.9363 & 0.7957 & 0.8763 & 0.8108 & 0.7750 & 0.7169 \\
CatBoost & 0.7881 & 0.6001 & \underline{0.7130} & 0.8614 & 0.9364 & \underline{0.8045} & 0.9021 & 0.8016 & 0.7695 & 0.7129 \\
LightGBM & 0.7878 & 0.5934 & 0.7079 & 0.8547 & 0.9316 & 0.8033 & 0.9006 & 0.7950 & 0.7717 & 0.7109 \\
XGBoost & 0.7831 & 0.5850 & 0.6925 & 0.8531 & 0.9329 & 0.7981 & \underline{0.9030} & 0.7987 & 0.7591 & 0.6974 \\
HistGradientBoostingTree & \underline{0.7909} & 0.5619 & 0.7018 & 0.8647 & 0.9364 & 0.7880 & 0.9007 & 0.8193 & 0.7490 & 0.6884 \\
GradientBoostingTree & 0.7657 & 0.6018 & 0.7048 & 0.8444 & 0.9216 & 0.8027 & 0.8800 & 0.7685 & 0.7752 & 0.7184 \\
RandomForest & 0.7859 & 0.5579 & 0.6998 & 0.8514 & 0.9208 & 0.7958 & 0.8876 & 0.8124 & 0.7635 & 0.7029 \\
\midrule
\multicolumn{11}{c}{Searched} \\
\midrule
DOFEN (ours) & 0.7934 & 0.6043 & \underline{0.7154} & \underline{0.8773} & \underline{0.9422} & 0.8038 & 0.8978 & 0.8262 & 0.7747 & 0.7175 \\
Trompt & 0.7831 & 0.6032 & 0.7090 & 0.8635 & 0.9374 & 0.7930 & 0.8913 & \underline{0.8373} & 0.7760 & \underline{0.7217} \\
GRANDE & 0.7795 & 0.6042 & 0.7125 & 0.8620 & 0.9368 & 0.8005 & 0.8885 & 0.8025 & 0.7759 & 0.7150 \\
FT-Transformer & 0.7567 & 0.6044 & 0.7051 & 0.8589 & 0.9345 & 0.7931 & 0.8869 & 0.8049 & 0.7751 & 0.7152 \\
ResNet & 0.7663 & 0.6045 & 0.6939 & 0.8571 & 0.9358 & 0.7864 & 0.8774 & 0.7945 & 0.7696 & 0.7073 \\
MLP & 0.7659 & 0.6035 & 0.6851 & 0.8541 & 0.9348 & 0.7888 & 0.8680 & 0.7836 & 0.7689 & 0.7068 \\
SAINT & 0.7660 & 0.6043 & 0.7060 & 0.8486 & 0.9359 & 0.7919 & 0.8879 & 0.8051 & 0.7606 & 0.7157 \\
NODE & 0.7596 & 0.6043 & 0.7084 & 0.8556 & 0.9375 & 0.7952 & 0.8808 & 0.8194 & 0.7753 & 0.7169 \\
CatBoost & 0.7898 & 0.6052 & 0.7145 & 0.8597 & 0.9375 & 0.8039 & 0.9007 & 0.8303 & 0.7762 & 0.7201 \\
LightGBM & \underline{0.7942} & 0.6050 & 0.7110 & 0.8574 & 0.9382 & 0.8017 & \underline{0.9019} & 0.8193 & 0.7716 & 0.7197 \\
XGBoost & 0.7917 & \underline{0.6059} & 0.7132 & 0.8600 & 0.9367 & 0.8034 & 0.9012 & 0.8177 & 0.7732 & 0.7156 \\
HistGradientBoostingTree & 0.7859 & 0.6050 & 0.7092 & 0.8592 & 0.9372 & \underline{0.8108} & 0.8934 & 0.8231 & \underline{0.7767} & 0.7194 \\
GradientBoostingTree & 0.7695 & 0.6042 & 0.7100 & 0.8550 & 0.9334 & 0.8022 & 0.8967 & 0.8182 & 0.7734 & 0.7172 \\
RandomForest & 0.7940 & 0.6047 & 0.7055 & 0.8541 & 0.9266 & 0.7983 & 0.8929 & 0.8275 & 0.7727 & 0.7182 \\
\bottomrule
\end{tabular}
\end{table}

\begin{table}[H]
\centering
\small
\setlength\tabcolsep{3.0pt}
\caption{The performance of \textbf{medium-sized classification} task (\emph{numerical features only}) (2).}
\label{tab:performance-medium-num-clf-2}
\begin{tabular}{l c c c c c c c}
\toprule
 & \textbf{361060} & \textbf{361070} & \textbf{361278} & \textbf{361063} & \textbf{361274} & \textbf{361062} & \textbf{Ranking} \\
\midrule
\multicolumn{8}{c}{Default} \\
\midrule
DOFEN (ours) & 0.8169 & 0.6196 & 0.7189 & \underline{0.8895} & \underline{0.7806} & 0.9822 & $4.81 \pm 3.42$ \\
Trompt & 0.8289 & 0.6160 & 0.6987 & 0.8805 & 0.7689 & \underline{0.9849} & $7.41 \pm 4.22$ \\
GRANDE & 0.8131 & 0.6010 & 0.7210 & 0.8890 & 0.7788 & 0.9783 & $5.38 \pm 3.61$ \\
FT-Transformer & 0.8082 & 0.5864 & 0.7175 & 0.8816 & 0.7562 & 0.9780 & $8.19 \pm 2.62$ \\
ResNet & 0.8062 & 0.5852 & 0.7186 & 0.8755 & 0.7449 & 0.9366 & $10.75 \pm 3.87$ \\
MLP & 0.8105 & 0.5808 & 0.7151 & 0.8765 & 0.7418 & 0.9153 & $11.56 \pm 3.82$ \\
SAINT & 0.8098 & 0.5799 & 0.7146 & 0.8842 & 0.7668 & 0.9718 & $8.88 \pm 3.56$ \\
NODE & 0.8151 & 0.5931 & \underline{0.7271} & 0.8823 & 0.7651 & 0.9701 & $6.75 \pm 3.62$ \\
CatBoost & 0.8448 & 0.6387 & 0.7222 & 0.8859 & 0.7785 & 0.9846 & $3.91 \pm 3.55$ \\
LightGBM & 0.8434 & 0.6439 & 0.7148 & 0.8843 & 0.7727 & 0.9838 & $6.06 \pm 2.96$ \\
XGBoost & 0.8611 & 0.6475 & 0.6948 & 0.8816 & 0.7600 & 0.9835 & $8.12 \pm 4.05$ \\
HistGradientBoostingTree & \underline{0.8623} & \underline{0.6633} & 0.7024 & 0.8848 & 0.7721 & 0.9846 & $6.22 \pm 5.10$ \\
GradientBoostingTree & 0.8216 & 0.6233 & 0.7157 & 0.8767 & 0.7618 & 0.9671 & $8.50 \pm 3.88$ \\
RandomForest & 0.8458 & 0.6308 & 0.7173 & 0.8782 & 0.7611 & 0.9803 & $8.47 \pm 3.69$ \\
\midrule
\multicolumn{8}{c}{Searched} \\
\midrule
DOFEN (ours) & 0.8378 & 0.6277 & \underline{0.7281} & \underline{0.8896} & \underline{0.7835} & 0.9828 & $3.84 \pm 3.51$ \\
Trompt & 0.8307 & 0.6271 & 0.7263 & 0.8846 & 0.7782 & 0.9837 & $5.75 \pm 3.75$ \\
GRANDE & 0.8217 & 0.6050 & 0.7214 & 0.8881 & 0.7805 & 0.9792 & $7.25 \pm 3.35$ \\
FT-Transformer & 0.8189 & 0.5787 & 0.7194 & 0.8790 & 0.7676 & 0.9834 & $10.06 \pm 3.16$ \\
ResNet & 0.8097 & 0.5807 & 0.7193 & 0.8744 & 0.7502 & 0.9510 & $11.75 \pm 3.23$ \\
MLP & 0.8057 & 0.5833 & 0.7187 & 0.8773 & 0.7446 & 0.9476 & $13.03 \pm 3.10$ \\
SAINT & 0.8185 & 0.5862 & 0.7190 & 0.8843 & 0.7713 & 0.9796 & $10.44 \pm 2.47$ \\
NODE & 0.8175 & 0.5895 & 0.7257 & 0.8813 & 0.7694 & 0.9693 & $8.91 \pm 3.19$ \\
CatBoost & 0.8631 & 0.6538 & 0.7230 & 0.8854 & 0.7802 & \underline{0.9849} & $3.00 \pm 2.94$ \\
LightGBM & 0.8599 & 0.6508 & 0.7217 & 0.8861 & 0.7769 & 0.9822 & $4.94 \pm 2.91$ \\
XGBoost & \underline{0.8686} & 0.6573 & 0.7172 & 0.8880 & 0.7785 & 0.9817 & $5.25 \pm 3.68$ \\
HistGradientBoostingTree & 0.8625 & \underline{0.6584} & 0.7203 & 0.8849 & 0.7739 & 0.9834 & $4.56 \pm 2.78$ \\
GradientBoostingTree & 0.8646 & 0.6321 & 0.7183 & 0.8816 & 0.7737 & 0.9808 & $8.12 \pm 2.92$ \\
RandomForest & 0.8609 & 0.6503 & 0.7164 & 0.8796 & 0.7720 & 0.9816 & $8.09 \pm 3.66$ \\
\bottomrule
\end{tabular}
\end{table}

\begin{table}[H]
\centering
\small
\setlength\tabcolsep{3.0pt}
\caption{The performance of \textbf{medium-sized classification} task (\emph{heterogeneous features}).}
\label{tab:performance-medium-het-clf}
\begin{tabular}{l c c c c c c c c}
\toprule
 & \textbf{361282} & \textbf{361286} & \textbf{361113} & \textbf{361283} & \textbf{361110} & \textbf{361111} & \textbf{361285} & \textbf{Ranking} \\
\midrule
\multicolumn{9}{c}{Default} \\
\midrule
DOFEN (ours) & 0.6495 & 0.6823 & 0.8240 & 0.7162 & 0.8275 & 0.6241 & \underline{0.7730} & $6.00 \pm 4.17$ \\
Trompt & 0.6191 & 0.6743 & \underline{0.8729} & 0.7017 & 0.8450 & 0.6425 & 0.7580 & $8.43 \pm 4.43$ \\
GRANDE & 0.6549 & 0.6759 & 0.8278 & \underline{0.7208} & 0.8266 & 0.6208 & 0.7659 & $5.43 \pm 4.03$ \\
FT-Transformer & 0.6543 & 0.6820 & 0.8565 & 0.7156 & 0.8252 & 0.5952 & 0.7635 & $5.71 \pm 3.93$ \\
ResNet & 0.6459 & 0.6756 & 0.8214 & 0.7055 & 0.8200 & 0.5883 & 0.7517 & $11.43 \pm 4.39$ \\
MLP & 0.6506 & \underline{0.6826} & 0.8259 & 0.7078 & 0.8161 & 0.5939 & 0.7486 & $9.43 \pm 5.21$ \\
SAINT & 0.6501 & 0.6750 & 0.8261 & 0.7059 & 0.8234 & 0.5958 & 0.7618 & $8.86 \pm 2.70$ \\
NODE & 0.6497 & 0.6753 & 0.8397 & 0.7146 & 0.8172 & 0.5895 & 0.7597 & $8.43 \pm 3.28$ \\
CatBoost & \underline{0.6570} & 0.6715 & 0.8369 & 0.7120 & 0.8501 & 0.6462 & 0.7680 & $4.86 \pm 4.60$ \\
LightGBM & 0.6489 & 0.6747 & 0.8323 & 0.7123 & 0.8637 & 0.6448 & 0.7643 & $6.29 \pm 2.92$ \\
XGBoost & 0.6315 & 0.6632 & 0.8413 & 0.6969 & \underline{0.8786} & \underline{0.6477} & 0.7594 & $7.71 \pm 5.24$ \\
HistGradientBoostingTree & 0.6500 & 0.6625 & 0.8334 & 0.7090 & 0.8685 & 0.6446 & 0.7647 & $6.43 \pm 3.45$ \\
GradientBoostingTree & 0.6559 & 0.6798 & 0.7892 & 0.7187 & 0.8293 & 0.6181 & 0.7501 & $7.29 \pm 4.60$ \\
RandomForest & 0.6482 & 0.6219 & 0.8471 & 0.7052 & 0.8629 & 0.6400 & 0.7531 & $8.71 \pm 4.32$ \\
\midrule
\multicolumn{9}{c}{Searched} \\
\midrule
DOFEN (ours) & 0.6583 & 0.6791 & 0.8760 & 0.7102 & 0.8594 & 0.6421 & \underline{0.7763} & $5.43 \pm 4.02$ \\
Trompt & 0.6431 & 0.6801 & \underline{0.8829} & 0.7159 & 0.8513 & 0.6429 & 0.7709 & $6.43 \pm 4.02$ \\
GRANDE & 0.6596 & 0.6531 & 0.8425 & 0.7186 & 0.8394 & 0.6060 & 0.7670 & $7.71 \pm 4.10$ \\
FT-Transformer & 0.6529 & 0.6818 & 0.8594 & 0.7119 & 0.8312 & 0.5877 & 0.7714 & $7.57 \pm 4.07$ \\
ResNet & 0.6508 & \underline{0.6842} & 0.8396 & 0.7032 & 0.8261 & 0.5941 & 0.7622 & $10.29 \pm 4.60$ \\
MLP & 0.6524 & 0.6782 & 0.8336 & 0.7101 & 0.8238 & 0.5888 & 0.7555 & $12.29 \pm 4.32$ \\
SAINT & 0.6527 & 0.6721 & 0.8505 & 0.7137 & 0.8302 & 0.5857 & 0.7657 & $10.57 \pm 3.46$ \\
NODE & 0.6498 & 0.6766 & 0.8397 & 0.7146 & 0.8205 & 0.5926 & 0.7580 & $11.86 \pm 3.81$ \\
CatBoost & \underline{0.6596} & 0.6775 & 0.8745 & \underline{0.7226} & 0.8772 & 0.6632 & 0.7714 & $3.57 \pm 4.70$ \\
LightGBM & 0.6574 & 0.6747 & 0.8639 & 0.7209 & \underline{0.8863} & 0.6609 & 0.7643 & $5.00 \pm 4.45$ \\
XGBoost & 0.6561 & 0.6810 & 0.8626 & 0.7182 & 0.8850 & \underline{0.6662} & 0.7679 & $3.86 \pm 3.66$ \\
HistGradientBoostingTree & 0.6563 & 0.6789 & 0.8478 & 0.7171 & 0.8801 & 0.6308 & 0.7654 & $6.86 \pm 2.20$ \\
GradientBoostingTree & 0.6538 & 0.6807 & 0.8546 & 0.7184 & 0.8789 & 0.6384 & 0.7617 & $6.57 \pm 2.64$ \\
RandomForest & 0.6542 & 0.6795 & 0.8587 & 0.7167 & 0.8770 & 0.6540 & 0.7600 & $7.00 \pm 2.27$ \\
\bottomrule
\end{tabular}
\end{table}

\begin{table}[H]
\centering
\small
\setlength\tabcolsep{3.0pt}
\caption{The performance of \textbf{medium-sized regression} task (\emph{numerical features only}) (1).}
\label{tab:performance-medium-num-rgr-1}
\begin{tabular}{l c c c c c c c c c c}
\toprule
 & \textbf{361077} & \textbf{361082} & \textbf{361081} & \textbf{361087} & \textbf{361280} & \textbf{361072} & \textbf{361281} & \textbf{361080} & \textbf{361074} & \textbf{361079} \\
\midrule
\multicolumn{11}{c}{Default} \\
\midrule
DOFEN (ours) & 0.2011 & 0.6874 & 0.9931 & 0.9194 & 0.5651 & 0.9837 & \underline{0.0269} & 0.9352 & 0.8921 & 0.5398 \\
Trompt & 0.8480 & 0.6829 & 0.9970 & 0.9275 & 0.5443 & 0.9723 & 0.0162 & 0.9415 & 0.8969 & 0.5453 \\
GRANDE & 0.7399 & 0.6376 & 0.9126 & 0.8340 & 0.5281 & 0.8812 & 0.0053 & 0.8239 & 0.6423 & \underline{0.6173} \\
FT-Transformer & 0.8436 & 0.6691 & 0.9958 & 0.9205 & 0.5308 & 0.9594 & 0.0000 & 0.9419 & 0.9115 & 0.5571 \\
ResNet & 0.8331 & 0.6423 & 0.9923 & 0.9145 & 0.4696 & 0.9747 & 0.0000 & 0.9404 & 0.8979 & 0.4995 \\
MLP & 0.8299 & 0.6634 & 0.9939 & 0.9091 & 0.5435 & 0.9570 & 0.0000 & 0.9411 & 0.8958 & 0.5057 \\
SAINT & 0.0000 & 0.6816 & 0.9938 & 0.9158 & \underline{0.5658} & 0.9835 & 0.0178 & 0.9422 & 0.8500 & 0.4679 \\
NODE & 0.4500 & 0.0000 & 0.0000 & 0.0000 & 0.0000 & 0.0000 & 0.0083 & 0.0000 & 0.8684 & 0.0000 \\
CatBoost & \underline{0.8576} & \underline{0.6993} & 0.9960 & \underline{0.9356} & 0.5279 & \underline{0.9856} & 0.0000 & \underline{0.9457} & \underline{0.9117} & 0.5101 \\
LightGBM & 0.8468 & 0.6928 & 0.9938 & 0.9225 & 0.5124 & 0.9846 & 0.0070 & 0.9449 & 0.8859 & 0.5195 \\
XGBoost & 0.8258 & 0.6793 & \underline{0.9976} & 0.9203 & 0.4817 & 0.9825 & 0.0000 & 0.9409 & 0.8848 & 0.4814 \\
HistGradientBoostingTree & 0.8464 & 0.6932 & 0.9938 & 0.9233 & 0.5259 & 0.9828 & 0.0052 & 0.9448 & 0.8855 & 0.5361 \\
GradientBoostingTree & 0.8397 & 0.6758 & 0.9962 & 0.8942 & 0.5399 & 0.9835 & 0.0251 & 0.9441 & 0.8022 & 0.4733 \\
RandomForest & 0.8372 & 0.6720 & 0.9931 & 0.9141 & 0.5359 & 0.9826 & 0.0000 & 0.9394 & 0.8330 & 0.5016 \\
\midrule
\multicolumn{11}{c}{Searched} \\
\midrule
DOFEN (ours) & 0.6955 & 0.6943 & 0.9945 & 0.9374 & 0.5735 & \underline{0.9877} & 0.0282 & 0.9448 & \underline{0.9236} & 0.5488 \\
Trompt & 0.8457 & 0.6916 & 0.9956 & 0.9280 & 0.5443 & 0.9873 & 0.0283 & 0.9427 & 0.8948 & 0.4650 \\
GRANDE & 0.7449 & 0.6426 & 0.9136 & 0.8352 & 0.5457 & 0.8786 & 0.0070 & 0.8270 & 0.6498 & \underline{0.6165} \\
FT-Transformer & 0.8457 & 0.6798 & 0.9975 & 0.9222 & 0.5620 & 0.9843 & 0.0195 & 0.9435 & 0.9163 & 0.4091 \\
ResNet & 0.8354 & 0.4023 & 0.9966 & 0.9167 & 0.5703 & 0.9823 & 0.0121 & nan & 0.9066 & 0.4850 \\
MLP & 0.8367 & 0.6753 & 0.9929 & 0.9094 & \underline{0.5776} & 0.9789 & 0.0143 & 0.9439 & 0.9181 & 0.4804 \\
SAINT & 0.7840 & 0.6858 & 0.9935 & 0.9245 & 0.5640 & 0.9849 & 0.0215 & 0.9443 & 0.9225 & 0.4719 \\
NODE & 0.8365 & 0.6704 & 0.9877 & 0.9260 & 0.5332 & 0.9730 & 0.0127 & 0.9427 & 0.9148 & 0.5257 \\
CatBoost & \underline{0.8554} & \underline{0.7061} & 0.9921 & \underline{0.9376} & 0.5353 & 0.9865 & 0.0306 & 0.9450 & 0.9103 & 0.4156 \\
LightGBM & 0.8468 & 0.6928 & 0.9927 & 0.9337 & 0.5399 & 0.9799 & \underline{0.0329} & 0.9449 & 0.8859 & 0.5163 \\
XGBoost & 0.8467 & 0.6947 & \underline{0.9976} & 0.9360 & 0.5426 & 0.9862 & 0.0303 & \underline{0.9456} & 0.9069 & 0.5445 \\
HistGradientBoostingTree & 0.8464 & 0.6932 & 0.9929 & 0.9267 & 0.5322 & 0.9745 & 0.0299 & 0.9449 & 0.8863 & 0.3912 \\
GradientBoostingTree & 0.8422 & 0.6892 & 0.9960 & 0.9250 & 0.5445 & 0.9853 & 0.0262 & 0.9448 & 0.8619 & 0.5135 \\
RandomForest & 0.8386 & 0.6870 & 0.9931 & 0.9242 & 0.5518 & 0.9829 & 0.0305 & 0.9453 & 0.8402 & 0.4841 \\
\bottomrule
\end{tabular}
\end{table}

\begin{table}[H]
\centering
\small
\setlength\tabcolsep{3.0pt}
\caption{The performance of \textbf{medium-sized regression} task (\emph{numerical features only}) (2).}
\label{tab:performance-medium-num-rgr-2}
\begin{tabular}{l c c c c c c c c c c}
\toprule
 & \textbf{361084} & \textbf{361078} & \textbf{361086} & \textbf{361083} & \textbf{361073} & \textbf{361085} & \textbf{361088} & \textbf{361076} & \textbf{361279} & \textbf{Ranking} \\
\midrule
\multicolumn{11}{c}{Default} \\
\midrule
DOFEN (ours) & 0.8723 & 0.8099 & 0.9756 & 0.4427 & 0.9885 & 0.8285 & 0.8950 & 0.4139 & 0.0000 & $6.84 \pm 3.68$ \\
Trompt & 0.8804 & 0.8352 & 0.9788 & 0.1699 & 0.9513 & 0.8096 & 0.8791 & 0.3168 & 0.0083 & $5.68 \pm 3.30$ \\
GRANDE & 0.7918 & 0.6854 & 0.8822 & 0.2149 & 0.9536 & 0.7400 & 0.7986 & 0.2609 & 0.0000 & $11.11 \pm 3.70$ \\
FT-Transformer & 0.8766 & 0.8235 & \underline{0.9794} & 0.1499 & 0.9313 & 0.8400 & 0.8751 & 0.2648 & 0.0000 & $7.24 \pm 3.77$ \\
ResNet & 0.7948 & 0.7729 & 0.9772 & 0.2050 & 0.6279 & 0.6979 & 0.8739 & 0.2598 & 0.0000 & $10.71 \pm 2.89$ \\
MLP & 0.8575 & 0.8133 & 0.9789 & 0.1615 & 0.8343 & 0.7922 & 0.8842 & 0.2792 & 0.0000 & $9.08 \pm 3.01$ \\
SAINT & 0.8731 & 0.8139 & 0.9788 & 0.4713 & \underline{0.9904} & 0.7859 & 0.8909 & 0.3632 & 0.0449 & $7.00 \pm 3.49$ \\
NODE & 0.0000 & 0.0000 & 0.0000 & 0.3622 & 0.0000 & 0.6828 & 0.0000 & 0.0000 & 0.0336 & $12.53 \pm 3.80$ \\
CatBoost & \underline{0.8873} & \underline{0.8472} & 0.9782 & 0.5291 & 0.9863 & \underline{0.8685} & \underline{0.9051} & 0.4500 & 0.0530 & $3.34 \pm 4.03$ \\
LightGBM & 0.8812 & 0.8351 & 0.9785 & 0.5306 & 0.9870 & 0.8143 & 0.8979 & 0.4286 & 0.0480 & $4.74 \pm 2.98$ \\
XGBoost & 0.8743 & 0.8374 & 0.9773 & \underline{0.5487} & 0.9850 & 0.8349 & 0.8955 & 0.4237 & 0.0000 & $7.08 \pm 3.36$ \\
HistGradientBoostingTree & 0.8816 & 0.8325 & 0.9785 & 0.5186 & 0.9865 & 0.8161 & 0.8964 & 0.4336 & 0.0522 & $5.00 \pm 2.64$ \\
GradientBoostingTree & 0.8617 & 0.7874 & 0.9794 & 0.4516 & 0.9349 & 0.8106 & 0.8563 & 0.3763 & 0.0000 & $7.74 \pm 3.62$ \\
RandomForest & 0.8689 & 0.8270 & 0.9768 & 0.5460 & 0.9867 & 0.8439 & 0.9011 & \underline{0.4807} & \underline{0.0601} & $6.92 \pm 3.81$ \\
\midrule
\multicolumn{11}{c}{Searched} \\
\midrule
DOFEN (ours) & 0.8844 & 0.8453 & 0.9788 & 0.4950 & 0.9943 & \underline{0.9012} & \underline{0.9145} & 0.4871 & 0.0706 & $4.47 \pm 3.79$ \\
Trompt & 0.8832 & 0.8187 & 0.9792 & 0.4557 & \underline{0.9958} & 0.8495 & 0.8949 & 0.4091 & 0.0376 & $7.32 \pm 3.13$ \\
GRANDE & 0.7918 & 0.6877 & 0.8840 & 0.2305 & 0.9541 & 0.7425 & 0.8083 & 0.2654 & 0.0000 & $12.68 \pm 4.10$ \\
FT-Transformer & 0.8823 & 0.8350 & 0.9795 & 0.4901 & 0.9949 & 0.8496 & 0.8874 & 0.3683 & 0.0517 & $7.47 \pm 3.38$ \\
ResNet & 0.8666 & 0.8266 & 0.9793 & 0.4410 & 0.9161 & 0.8181 & 0.8948 & 0.3669 & 0.0248 & $9.78 \pm 3.67$ \\
MLP & 0.8669 & 0.8190 & \underline{0.9796} & 0.4654 & 0.9701 & 0.8422 & 0.8930 & 0.3946 & 0.0128 & $9.00 \pm 3.68$ \\
SAINT & 0.8811 & 0.8275 & 0.9795 & 0.4951 & 0.9949 & 0.7580 & 0.8937 & 0.3729 & 0.0593 & $7.68 \pm 2.89$ \\
NODE & 0.8762 & 0.7969 & 0.9782 & 0.3743 & 0.9580 & 0.7309 & 0.8857 & 0.2874 & 0.0393 & $11.11 \pm 3.45$ \\
CatBoost & 0.8870 & 0.8489 & 0.9793 & 0.5399 & 0.9907 & 0.8701 & 0.9100 & 0.5007 & 0.0736 & $4.42 \pm 4.06$ \\
LightGBM & 0.8860 & \underline{0.8539} & 0.9785 & 0.5306 & 0.9870 & 0.8140 & 0.9048 & 0.4286 & 0.0663 & $6.53 \pm 3.66$ \\
XGBoost & \underline{0.8884} & 0.8497 & 0.9787 & 0.5517 & 0.9908 & 0.8617 & 0.9106 & 0.5020 & 0.0819 & $3.63 \pm 3.60$ \\
HistGradientBoostingTree & 0.8822 & 0.8378 & 0.9791 & 0.5186 & 0.9870 & 0.8144 & 0.9009 & 0.4313 & 0.0612 & $7.84 \pm 3.13$ \\
GradientBoostingTree & 0.8827 & 0.8405 & 0.9794 & 0.5532 & 0.9896 & 0.8162 & 0.9028 & 0.4660 & 0.0525 & $6.47 \pm 2.28$ \\
RandomForest & 0.8711 & 0.8291 & 0.9789 & \underline{0.5626} & 0.9891 & 0.8595 & 0.9087 & \underline{0.5042} & \underline{0.0929} & $6.37 \pm 3.65$ \\
\bottomrule
\end{tabular}
\end{table}

\begin{table}[H]
\centering
\small
\setlength\tabcolsep{3.0pt}
\caption{The performance of \textbf{medium-sized regression} task (\emph{heterogeneous features}) (1).}
\label{tab:performance-medium-het-rgr-1}
\begin{tabular}{l c c c c c c c c c c}
\toprule
 & \textbf{361293} & \textbf{361292} & \textbf{361099} & \textbf{361098} & \textbf{361097} & \textbf{361104} & \textbf{361288} & \textbf{361093} & \textbf{361291} & \textbf{361096} \\
\midrule
\multicolumn{11}{c}{Default} \\
\midrule
DOFEN (ours) & 0.0359 & 0.5258 & 0.9341 & 0.9932 & \underline{0.5750} & 0.9997 & \underline{0.5686} & \underline{0.9837} & 0.0694 & 0.9869 \\
Trompt & 0.0195 & 0.4939 & 0.9393 & 0.9963 & 0.5409 & 0.9996 & 0.5459 & 0.9470 & 0.0364 & 0.9888 \\
GRANDE & 0.0000 & 0.4356 & 0.8014 & 0.9101 & 0.4853 & 0.9151 & 0.5354 & 0.9723 & 0.0231 & 0.8957 \\
FT-Transformer & nan & 0.5160 & 0.9280 & 0.9960 & 0.5540 & 0.9997 & 0.5480 & 0.9490 & 0.0000 & 0.9872 \\
ResNet & 0.0000 & 0.4993 & 0.8861 & 0.9883 & 0.5470 & 0.9975 & 0.4229 & 0.9244 & 0.0000 & 0.9857 \\
MLP & 0.0000 & 0.5105 & 0.9213 & 0.9942 & 0.5546 & \underline{0.9998} & 0.5486 & 0.9497 & 0.0000 & 0.9861 \\
SAINT & 0.0375 & 0.5191 & 0.9375 & 0.9930 & 0.5522 & 0.9990 & 0.5676 & 0.9777 & 0.0591 & 0.9867 \\
NODE & 0.0361 & 0.0000 & 0.0000 & 0.0000 & 0.0000 & 0.0000 & 0.0000 & 0.9797 & 0.0566 & 0.0000 \\
CatBoost & 0.0156 & \underline{0.5347} & \underline{0.9421} & 0.9959 & 0.5633 & 0.9997 & 0.5375 & 0.9801 & 0.0621 & \underline{0.9911} \\
LightGBM & 0.0234 & 0.5275 & 0.9402 & 0.9938 & 0.5477 & 0.9997 & 0.5183 & 0.9823 & 0.0618 & 0.9901 \\
XGBoost & 0.0000 & 0.4807 & 0.9393 & \underline{0.9976} & 0.4968 & 0.9997 & 0.4797 & 0.9759 & 0.0000 & 0.9896 \\
HistGradientBoostingTree & 0.0290 & 0.5226 & 0.9410 & 0.9939 & 0.5421 & 0.9997 & 0.5289 & 0.9823 & 0.0557 & 0.9908 \\
GradientBoostingTree & \underline{0.0451} & 0.5066 & 0.8415 & 0.9962 & 0.5717 & 0.9997 & 0.5470 & 0.9827 & \underline{0.0736} & 0.9842 \\
RandomForest & 0.0000 & 0.4748 & 0.9369 & 0.9929 & 0.5034 & 0.9998 & 0.5407 & 0.9799 & 0.0000 & 0.9878 \\
\midrule
\multicolumn{11}{c}{Searched} \\
\midrule
DOFEN (ours) & 0.0440 & \underline{0.5377} & \underline{0.9477} & 0.9945 & 0.5769 & 0.9998 & 0.5760 & \underline{0.9849} & 0.0738 & 0.9914 \\
Trompt & 0.0457 & 0.5330 & 0.9397 & 0.9957 & 0.5816 & 0.9997 & 0.5465 & 0.9848 & 0.0744 & 0.9899 \\
GRANDE & 0.0000 & 0.4386 & 0.8041 & 0.9137 & 0.4947 & 0.9166 & 0.5061 & 0.9725 & 0.0340 & 0.8956 \\
FT-Transformer & 0.0452 & 0.5217 & 0.9388 & \underline{0.9979} & 0.5663 & 0.9998 & 0.5645 & 0.9796 & 0.0679 & 0.9899 \\
ResNet & 0.0411 & 0.5093 & 0.9356 & 0.9967 & 0.5693 & 0.9997 & 0.5718 & 0.9796 & 0.0599 & nan \\
MLP & 0.0413 & 0.5157 & 0.9341 & 0.9948 & 0.5571 & \underline{0.9998} & \underline{0.5775} & 0.9801 & 0.0623 & 0.9876 \\
SAINT & 0.0450 & 0.5256 & 0.9409 & 0.9958 & 0.5610 & 0.9997 & 0.5670 & 0.9789 & 0.0656 & 0.9893 \\
NODE & 0.0369 & 0.5169 & 0.8428 & 0.9878 & 0.5735 & 0.9998 & 0.5409 & 0.9823 & 0.0580 & 0.9860 \\
CatBoost & \underline{0.0479} & 0.5350 & 0.9465 & 0.9922 & \underline{0.5816} & 0.9997 & 0.5449 & 0.9839 & 0.0778 & \underline{0.9917} \\
LightGBM & 0.0431 & 0.5336 & 0.9432 & 0.9941 & 0.5546 & 0.9997 & 0.5370 & 0.9823 & \underline{0.0789} & 0.9907 \\
XGBoost & 0.0476 & 0.5357 & 0.9465 & 0.9977 & 0.5777 & 0.9998 & 0.5556 & 0.9832 & 0.0773 & 0.9911 \\
HistGradientBoostingTree & 0.0470 & 0.5268 & 0.9416 & 0.9933 & 0.5773 & 0.9997 & 0.5453 & 0.9815 & 0.0730 & 0.9909 \\
GradientBoostingTree & 0.0475 & 0.5298 & 0.9414 & 0.9958 & 0.5760 & 0.9998 & 0.5498 & 0.9817 & 0.0746 & 0.9897 \\
RandomForest & 0.0456 & 0.5004 & 0.9366 & 0.9935 & 0.5755 & 0.9998 & 0.5561 & 0.9809 & 0.0766 & 0.9881 \\
\bottomrule
\end{tabular}
\end{table}

\begin{table}[H]
\centering
\small
\setlength\tabcolsep{3.0pt}
\caption{The performance of \textbf{medium-sized regression} task (\emph{heterogeneous features}) (2).}
\label{tab:performance-medium-het-rgr-2}
\begin{tabular}{l c c c c c c c c}
\toprule
 & \textbf{361102} & \textbf{361294} & \textbf{361101} & \textbf{361103} & \textbf{361289} & \textbf{361287} & \textbf{361094} & \textbf{Ranking} \\
\midrule
\multicolumn{9}{c}{Default} \\
\midrule
DOFEN (ours) & 0.8838 & 0.9756 & 0.4178 & 0.6647 & 0.1835 & 0.0329 & 0.9996 & $5.47 \pm 3.39$ \\
Trompt & 0.8902 & 0.9782 & 0.0961 & 0.6494 & 0.1817 & \underline{0.0445} & 0.9995 & $7.29 \pm 3.71$ \\
GRANDE & 0.7977 & 0.8822 & 0.2421 & 0.6217 & 0.0992 & 0.0000 & 0.9615 & $11.88 \pm 2.95$ \\
FT-Transformer & 0.8883 & nan & 0.2472 & 0.6710 & 0.0200 & 0.0000 & 0.9998 & $8.00 \pm 3.17$ \\
ResNet & 0.8736 & 0.9782 & 0.2434 & 0.6487 & 0.0305 & 0.0000 & 0.9958 & $11.00 \pm 2.89$ \\
MLP & 0.8751 & 0.9792 & 0.1580 & 0.6555 & 0.0321 & 0.0000 & 0.9999 & $8.06 \pm 3.79$ \\
SAINT & 0.8836 & 0.9777 & 0.4631 & 0.6602 & 0.1712 & 0.0401 & 0.9998 & $6.41 \pm 2.62$ \\
NODE & 0.0000 & 0.0000 & 0.3719 & 0.0000 & 0.0000 & 0.0370 & 0.0000 & $11.47 \pm 4.58$ \\
CatBoost & \underline{0.8975} & 0.9776 & 0.5463 & \underline{0.6916} & 0.1843 & 0.0313 & 0.9999 & $3.76 \pm 3.45$ \\
LightGBM & 0.8905 & 0.9779 & 0.5448 & 0.6874 & 0.1792 & 0.0199 & 0.9999 & $5.18 \pm 2.89$ \\
XGBoost & 0.8834 & 0.9773 & \underline{0.5699} & 0.6619 & 0.1653 & 0.0000 & 1.0000 & $7.47 \pm 3.94$ \\
HistGradientBoostingTree & 0.8914 & 0.9785 & 0.5389 & 0.6904 & 0.1727 & 0.0302 & 0.9999 & $5.00 \pm 3.11$ \\
GradientBoostingTree & 0.8693 & \underline{0.9794} & 0.4694 & 0.6717 & \underline{0.1861} & 0.0305 & 0.9994 & $5.35 \pm 4.27$ \\
RandomForest & 0.8747 & 0.9767 & 0.5619 & 0.6551 & 0.1639 & 0.0233 & \underline{1.0000} & $7.94 \pm 3.49$ \\
\midrule
\multicolumn{9}{c}{Searched} \\
\midrule
DOFEN (ours) & 0.8929 & 0.9788 & 0.5202 & 0.6689 & 0.1865 & 0.0600 & 0.9996 & $5.24 \pm 3.65$ \\
Trompt & 0.8916 & 0.9787 & 0.4804 & 0.6690 & 0.1821 & 0.0503 & 0.9999 & $7.24 \pm 2.80$ \\
GRANDE & 0.7943 & 0.8837 & 0.2621 & 0.6290 & 0.0960 & 0.0000 & 0.9622 & $13.88 \pm 3.05$ \\
FT-Transformer & 0.8932 & 0.9796 & 0.5281 & 0.6730 & 0.1797 & 0.0501 & 0.9999 & $7.24 \pm 2.70$ \\
ResNet & 0.8845 & 0.9793 & 0.4788 & 0.6562 & 0.1795 & 0.0215 & 0.9982 & $9.88 \pm 3.62$ \\
MLP & 0.8845 & \underline{0.9796} & 0.4739 & 0.6590 & 0.1716 & 0.0261 & 0.9999 & $9.12 \pm 4.19$ \\
SAINT & 0.8912 & 0.9796 & 0.5195 & 0.6706 & 0.1820 & 0.0530 & 0.9999 & $8.18 \pm 2.97$ \\
NODE & 0.8842 & 0.9782 & 0.4972 & 0.6477 & 0.0477 & 0.0423 & 0.9984 & $11.00 \pm 3.40$ \\
CatBoost & 0.8941 & 0.9787 & 0.5665 & \underline{0.6933} & 0.1854 & 0.0574 & 1.0000 & $4.47 \pm 4.05$ \\
LightGBM & 0.8935 & 0.9781 & 0.5448 & 0.6874 & \underline{0.1868} & 0.0546 & 1.0000 & $6.53 \pm 3.94$ \\
XGBoost & \underline{0.8967} & 0.9788 & 0.5804 & 0.6907 & 0.1848 & 0.0639 & 1.0000 & $3.24 \pm 3.07$ \\
HistGradientBoostingTree & 0.8907 & 0.9791 & 0.5383 & 0.6904 & 0.1858 & 0.0646 & 1.0000 & $6.24 \pm 2.85$ \\
GradientBoostingTree & 0.8892 & 0.9794 & 0.5756 & 0.6843 & 0.1857 & 0.0305 & \underline{1.0000} & $5.71 \pm 2.68$ \\
RandomForest & 0.8750 & 0.9788 & \underline{0.5842} & 0.6742 & 0.1827 & \underline{0.0713} & 1.0000 & $6.82 \pm 3.77$ \\
\bottomrule
\end{tabular}
\end{table}

\begin{table}[H]
\centering
\small
\setlength\tabcolsep{3.0pt}
\caption{The performance of \textbf{large-sized classification} task (\emph{numerical features only}).}
\label{tab:performance-large-num-clf}
\begin{tabular}{l c c c c c}
\toprule
 & \textbf{361069} & \textbf{361068} & \textbf{361061} & \textbf{361274} & \textbf{Ranking} \\
\midrule
\multicolumn{6}{c}{Default} \\
\midrule
DOFEN (ours) & \underline{0.7306} & \underline{0.9480} & 0.8222 & \underline{0.8018} & $3.25 \pm 5.22$ \\
Trompt & 0.7213 & 0.9468 & \underline{0.9004} & 0.7954 & $2.88 \pm 4.35$ \\
GRANDE & 0.7248 & 0.9425 & 0.8315 & 0.7979 & $5.00 \pm 3.05$ \\
FT-Transformer & 0.6960 & 0.9403 & 0.8983 & 0.7586 & $8.25 \pm 4.16$ \\
ResNet & 0.6988 & 0.9409 & 0.8801 & 0.7358 & $8.50 \pm 4.04$ \\
MLP & nan & nan & nan & nan & $nan \pm nan$ \\
SAINT & 0.7181 & 0.9436 & 0.8694 & 0.7860 & $6.00 \pm 1.30$ \\
NODE & 0.7247 & 0.9461 & 0.8886 & 0.7946 & $3.75 \pm 2.92$ \\
CatBoost & 0.7261 & 0.9432 & 0.8377 & 0.7954 & $4.38 \pm 2.77$ \\
LightGBM & 0.7212 & 0.9371 & 0.8071 & 0.7870 & $8.00 \pm 2.30$ \\
XGBoost & 0.7164 & 0.9367 & 0.8361 & 0.7828 & $8.50 \pm 2.61$ \\
HistGradientBoostingTree & nan & nan & nan & nan & $nan \pm nan$ \\
GradientBoostingTree & 0.7103 & 0.9225 & 0.7698 & 0.7718 & $11.00 \pm 4.58$ \\
RandomForest & 0.7158 & 0.9308 & 0.8767 & 0.7797 & $8.50 \pm 3.29$ \\
\midrule
\multicolumn{6}{c}{Searched} \\
\midrule
DOFEN (ours) & \underline{0.7355} & \underline{0.9509} & 0.9033 & \underline{0.8070} & $1.75 \pm 4.76$ \\
Trompt & 0.7286 & 0.9436 & \underline{0.9127} & 0.7988 & $4.25 \pm 3.63$ \\
GRANDE & 0.7266 & 0.9455 & 0.8524 & 0.7981 & $6.25 \pm 3.51$ \\
FT-Transformer & 0.7294 & 0.9445 & 0.9060 & 0.7962 & $4.00 \pm 3.11$ \\
ResNet & 0.7229 & 0.9444 & 0.8930 & 0.7832 & $10.00 \pm 4.66$ \\
MLP & nan & nan & nan & nan & $nan \pm nan$ \\
SAINT & 0.7259 & 0.9441 & 0.8954 & 0.7930 & $8.50 \pm 1.92$ \\
NODE & 0.7262 & 0.9475 & 0.8982 & 0.7946 & $5.50 \pm 2.35$ \\
CatBoost & 0.7300 & 0.9446 & 0.9021 & 0.7994 & $3.25 \pm 3.70$ \\
LightGBM & 0.7243 & 0.9435 & 0.8968 & 0.7939 & $9.25 \pm 2.92$ \\
XGBoost & 0.7280 & 0.9436 & 0.8942 & 0.7963 & $7.00 \pm 2.17$ \\
HistGradientBoostingTree & nan & nan & nan & nan & $nan \pm nan$ \\
GradientBoostingTree & 0.7249 & 0.9401 & 0.8976 & 0.7943 & $8.75 \pm 2.59$ \\
RandomForest & 0.7199 & 0.9353 & 0.9059 & 0.7885 & $9.50 \pm 5.05$ \\
\bottomrule
\end{tabular}
\end{table}

\begin{table}[H]
\centering
\small
\setlength\tabcolsep{3.0pt}
\caption{The performance of \textbf{large-sized classification} task (\emph{heterogeneous features}).}
\label{tab:performance-large-het-clf}
\begin{tabular}{l c c c}
\toprule
 & \textbf{361113} & \textbf{361285} & \textbf{Ranking} \\
\midrule
\multicolumn{4}{c}{Default} \\
\midrule
DOFEN (ours) & 0.8691 & \underline{0.7870} & $5.00 \pm 5.11$ \\
Trompt & 0.9276 & 0.7836 & $2.00 \pm 5.77$ \\
GRANDE & 0.8568 & 0.7785 & $8.00 \pm 3.51$ \\
FT-Transformer & \underline{0.9317} & 0.7609 & $6.00 \pm 5.03$ \\
ResNet & 0.8945 & 0.7653 & $8.00 \pm 3.51$ \\
MLP & nan & nan & $nan \pm nan$ \\
SAINT & 0.9123 & 0.7731 & $6.50 \pm 2.57$ \\
NODE & 0.9199 & 0.7774 & $5.00 \pm 3.75$ \\
CatBoost & 0.8827 & 0.7821 & $5.50 \pm 2.29$ \\
LightGBM & 0.8476 & 0.7797 & $8.00 \pm 4.16$ \\
XGBoost & 0.8781 & 0.7822 & $5.50 \pm 3.04$ \\
HistGradientBoostingTree & nan & nan & $nan \pm nan$ \\
GradientBoostingTree & 0.7946 & 0.7519 & $12.00 \pm 6.35$ \\
RandomForest & 0.9066 & 0.7767 & $6.50 \pm 1.61$ \\
\midrule
\multicolumn{4}{c}{Searched} \\
\midrule
DOFEN (ours) & \underline{0.9414} & 0.7984 & $1.50 \pm 6.08$ \\
Trompt & 0.9395 & 0.7844 & $4.50 \pm 3.82$ \\
GRANDE & 0.8914 & 0.7771 & $12.00 \pm 6.35$ \\
FT-Transformer & 0.9359 & 0.7875 & $4.50 \pm 2.93$ \\
ResNet & 0.9227 & 0.7836 & $9.00 \pm 3.06$ \\
MLP & nan & nan & $nan \pm nan$ \\
SAINT & 0.9252 & 0.7796 & $10.00 \pm 4.16$ \\
NODE & 0.9219 & 0.7800 & $10.50 \pm 4.93$ \\
CatBoost & 0.9366 & \underline{0.8016} & $2.00 \pm 5.29$ \\
LightGBM & 0.9312 & 0.7979 & $5.00 \pm 2.00$ \\
XGBoost & 0.9291 & 0.7982 & $5.50 \pm 2.65$ \\
HistGradientBoostingTree & nan & nan & $nan \pm nan$ \\
GradientBoostingTree & 0.9307 & 0.7849 & $6.50 \pm 0.58$ \\
RandomForest & 0.9329 & 0.7811 & $7.00 \pm 2.31$ \\
\bottomrule
\end{tabular}
\end{table}

\begin{table}[H]
\centering
\small
\setlength\tabcolsep{3.0pt}
\caption{The performance of \textbf{large-sized regression} task (\emph{numerical features only}).}
\label{tab:performance-large-num-rgr}
\begin{tabular}{l c c c c}
\toprule
 & \textbf{361080} & \textbf{361083} & \textbf{361091} & \textbf{Ranking} \\
\midrule
\multicolumn{5}{c}{Default} \\
\midrule
DOFEN (ours) & 0.9469 & 0.5459 & \underline{0.3240} & $3.00 \pm 3.54$ \\
Trompt & 0.9458 & 0.3379 & 0.2498 & $7.00 \pm 1.29$ \\
GRANDE & 0.8272 & 0.3243 & 0.2031 & $9.33 \pm 3.40$ \\
FT-Transformer & 0.9452 & 0.1198 & 0.1172 & $9.67 \pm 4.27$ \\
ResNet & 0.9410 & 0.2469 & 0.1188 & $9.67 \pm 4.11$ \\
MLP & nan & nan & nan & $nan \pm nan$ \\
SAINT & 0.9445 & 0.5344 & 0.2887 & $5.67 \pm 2.53$ \\
NODE & 0.9453 & 0.0000 & 0.2763 & $7.67 \pm 3.10$ \\
CatBoost & \underline{0.9476} & 0.5847 & 0.3020 & $1.67 \pm 4.69$ \\
LightGBM & 0.9475 & 0.5607 & 0.2810 & $3.00 \pm 3.35$ \\
XGBoost & 0.9474 & \underline{0.6087} & 0.2512 & $3.67 \pm 3.30$ \\
HistGradientBoostingTree & nan & nan & nan & $nan \pm nan$ \\
GradientBoostingTree & 0.9459 & 0.4635 & 0.2574 & $5.67 \pm 0.63$ \\
RandomForest & nan & nan & nan & $nan \pm nan$ \\
\midrule
\multicolumn{5}{c}{Searched} \\
\midrule
DOFEN (ours) & \underline{0.9491} & 0.5665 & \underline{0.3442} & $2.00 \pm 4.72$ \\
Trompt & 0.9461 & 0.5242 & 0.2971 & $8.33 \pm 2.59$ \\
GRANDE & 0.8289 & 0.3707 & 0.2078 & $11.00 \pm 5.00$ \\
FT-Transformer & 0.9463 & 0.5382 & 0.3042 & $6.67 \pm 1.50$ \\
ResNet & 0.9464 & 0.5282 & 0.2789 & $8.33 \pm 2.72$ \\
MLP & nan & nan & nan & $nan \pm nan$ \\
SAINT & 0.9465 & 0.5491 & 0.3041 & $6.00 \pm 0.25$ \\
NODE & 0.9460 & 0.3967 & 0.2892 & $9.33 \pm 3.79$ \\
CatBoost & 0.9480 & 0.5994 & 0.3130 & $2.33 \pm 3.61$ \\
LightGBM & 0.9475 & 0.5607 & 0.2810 & $6.00 \pm 2.17$ \\
XGBoost & 0.9479 & \underline{0.6241} & 0.3065 & $2.33 \pm 3.71$ \\
HistGradientBoostingTree & nan & nan & nan & $nan \pm nan$ \\
GradientBoostingTree & 0.9471 & 0.6167 & 0.3048 & $3.67 \pm 2.50$ \\
RandomForest & nan & nan & nan & $nan \pm nan$ \\
\bottomrule
\end{tabular}
\end{table}

\begin{table}[H]
\centering
\small
\setlength\tabcolsep{3.0pt}
\caption{The performance of \textbf{large-sized regression} task (\emph{heterogeneous features}).}
\label{tab:performance-large-het-rgr}
\begin{tabular}{l c c c c c c}
\toprule
 & \textbf{361104} & \textbf{361095} & \textbf{361096} & \textbf{361101} & \textbf{361103} & \textbf{Ranking} \\
\midrule
\multicolumn{7}{c}{Default} \\
\midrule
DOFEN (ours) & 0.9998 & 0.6120 & 0.9923 & 0.5288 & 0.6823 & $4.00 \pm 2.48$ \\
Trompt & 0.9996 & 0.6097 & 0.9917 & 0.4035 & 0.7048 & $6.20 \pm 1.86$ \\
GRANDE & 0.9153 & 0.4275 & 0.9011 & 0.3544 & 0.6370 & $10.00 \pm 3.99$ \\
FT-Transformer & 0.9994 & 0.3514 & 0.9923 & 0.4061 & 0.6761 & $7.40 \pm 2.79$ \\
ResNet & 0.9895 & 0.3370 & 0.9816 & 0.3971 & 0.6660 & $9.80 \pm 3.27$ \\
MLP & nan & nan & nan & nan & nan & $nan \pm nan$ \\
SAINT & 0.9997 & 0.3891 & 0.9918 & 0.5480 & 0.6874 & $5.80 \pm 1.79$ \\
NODE & 0.9997 & 0.4156 & 0.9875 & 0.0198 & 0.6626 & $8.80 \pm 2.79$ \\
CatBoost & 0.9998 & 0.6332 & \underline{0.9928} & 0.6050 & \underline{0.7068} & $2.00 \pm 3.44$ \\
LightGBM & 0.9998 & 0.6324 & 0.9916 & 0.5769 & 0.7037 & $4.00 \pm 2.40$ \\
XGBoost & \underline{0.9998} & \underline{0.6345} & 0.9922 & \underline{0.6244} & 0.7060 & $1.80 \pm 3.93$ \\
HistGradientBoostingTree & nan & nan & nan & nan & nan & $nan \pm nan$ \\
GradientBoostingTree & 0.9998 & 0.6165 & 0.9857 & 0.4809 & 0.6773 & $6.20 \pm 1.74$ \\
RandomForest & nan & nan & nan & nan & nan & $nan \pm nan$ \\
\midrule
\multicolumn{7}{c}{Searched} \\
\midrule
DOFEN (ours) & 0.9998 & 0.6251 & \underline{0.9937} & 0.5912 & 0.6980 & $4.00 \pm 3.06$ \\
Trompt & 0.9998 & 0.6286 & 0.9918 & 0.5479 & 0.7073 & $7.00 \pm 2.34$ \\
GRANDE & 0.9158 & 0.4348 & 0.9010 & 0.3899 & 0.6421 & $10.40 \pm 4.02$ \\
FT-Transformer & 0.9998 & 0.3899 & 0.9925 & 0.5694 & 0.7047 & $7.40 \pm 2.64$ \\
ResNet & 0.9998 & 0.3936 & 0.9922 & 0.5390 & 0.6865 & $8.00 \pm 2.71$ \\
MLP & nan & nan & nan & nan & nan & $nan \pm nan$ \\
SAINT & 0.9998 & 0.3951 & 0.9925 & 0.5662 & 0.6975 & $6.40 \pm 2.07$ \\
NODE & 0.9998 & 0.5908 & 0.9918 & 0.4135 & 0.6668 & $8.20 \pm 2.99$ \\
CatBoost & 0.9998 & 0.6363 & 0.9932 & 0.6262 & 0.7117 & $3.20 \pm 3.39$ \\
LightGBM & 0.9998 & 0.6324 & 0.9924 & 0.5769 & 0.7091 & $5.40 \pm 2.66$ \\
XGBoost & \underline{0.9998} & \underline{0.6387} & 0.9932 & \underline{0.6472} & \underline{0.7122} & $1.40 \pm 4.00$ \\
HistGradientBoostingTree & nan & nan & nan & nan & nan & $nan \pm nan$ \\
GradientBoostingTree & 0.9998 & 0.6301 & 0.9918 & 0.6356 & 0.7057 & $4.60 \pm 2.79$ \\
RandomForest & nan & nan & nan & nan & nan & $nan \pm nan$ \\
\bottomrule
\end{tabular}
\end{table}
\section{Evaluation Results on other Benchmarks}

This section aims to evaluate DOFEN's effectiveness and generalizability under different scenarios of benchmark settings.

\subsection{On Datasets used in FT-Transformer paper}
\label{sec:eval-ft-result}

To have a more comprehensive comparison between methods on larger size datasets, we choose to evaluate DOFEN on datasets used in FT-Transformer paper, which are significantly larger than the datasets used in the large-sized benchmark of Tabular Benchmark. For the experiment settings, we mainly follow the ones acknowledged in the FT-Transformer paper but with a few adjustments. Here we outlined two changes to the experiment settings we have made:

\begin{enumerate}
    \item{First, due to the lack of computational resources and time for these large size datasets, we only report the result of 5 different seeds, instead of the original setting that averages the result across 15 seeds from FT-Transformer paper.}

    \item{Second, for model comparison, aside from models included in FT-Transformer paper itself (i.e. FT-Transformer, Catboost, and XGBoost), we additionally include two state-of-the-art deep learning models, Trompt and GRANDE, to show the effectiveness of DOFEN.}
\end{enumerate}

The performance of FT-Transformer, Catboost, and XGBoost are acquired from the official GitHub repository of FT-Transformer, and the performance of Trompt is obtained from Trompt paper. It is worth noting that we only report performance with default hyperparameter settings for Trompt, GRANDE, and DOFEN. For Trompt, this is simply due to the searched results are not provided; For DOFEN and GRANDE, this is due to the lack of time and resources. Additionally, we adjust the default setting of DOFEN to better accommodate with large size dataset by setting $N_\text{head}$ to 4, other hyperparameter settings have remained as default.

The experiment results are provided in \cref{tab:performance-ft}. Although DOFEN only reports results using default performance, we are impressed that the default DOFEN already exceeds the searched performance of FT-Transformer on 7 out of 11 datasets and ranks first on 3 out of 11 datasets. This indicates the potential of DOFEN to achieve even better performance after utilizing hyperparameter search techniques. 

\begin{table}[H]
\centering
\small
\setlength\tabcolsep{3.0pt}
\caption{The performance of datasets used in FT-Transformer paper, averaged within 5 seeds. As the tree-based models act as an upper bound of performance, we highlight the best performing "deep learning model" in \textbf{\underline{Bold with underline}}, and the "deep learning model" with second performance in \textbf{Bold}.}
\begin{threeparttable}
\label{tab:performance-ft}
\begin{tabular}{l c c c c c c c c c c c}
\toprule
 & \textbf{CA ↓} & \textbf{AD ↑} & \textbf{HE ↑} & \textbf{JA ↑} & \textbf{HI ↑} & \textbf{AL ↑} & \textbf{EP ↑} & \textbf{YE ↓} & \textbf{CO ↑} & \textbf{YA ↓} & \textbf{MI ↓} \\
\midrule
\multicolumn{12}{c}{Default} \\
\midrule
DOFEN (ours) & \textbf{\underline{0.4584}} & \textbf{\underline{0.8667}} & \textbf{0.3858} & \textbf{\underline{0.7332}} & \textbf{\underline{0.7311}} & \textbf{0.9567} & OOM & \textbf{\underline{8.7572}} & 0.9104 & \textbf{\underline{0.7470}} & \textbf{\underline{0.7438}} \\
GRANDE & 0.481 & 0.859 & 0.350 & 0.723 & 0.721 & 0.785 & 0.870 & 8.943 & 0.792 & 0.779 & 0.754 \\
Trompt & 0.4793 & \textbf{0.8622} & 0.3665 & \textbf{0.7287} & \textbf{0.7299} & 0.9316 & 0.8941 & \textbf{8.8427} & 0.9035 & \textbf{0.7535} & \textbf{0.7465} \\
FT-Transformer & 0.4677 & 0.8580 & 0.3803 & 0.7237 & 0.7237 & 0.9522 & \textbf{0.8960} & 8.9173 & \textbf{0.9667} & 0.7560 & 0.7471 \\
CatBoost & 0.4303 & 0.8730 & 0.3820 & 0.7207 & 0.7255 & 0.9464 & 0.8882 & 8.9140 & 0.9077 & 0.7510 & 0.7454 \\
XGBoost & 0.4622 & 0.8741 & 0.3479 & 0.7110 & 0.7165 & 0.9242 & 0.8799 & 9.1922 & 0.9640 & 0.7607 & 0.7514 \\
\midrule
\multicolumn{12}{c}{Searched} \\
\midrule
FT-Transformer & \textbf{0.4597} & 0.8583 & \textbf{\underline{0.3916}} & \textbf{\underline{0.7332}} & 0.7292 & \textbf{\underline{0.9608}} & \textbf{\underline{0.8981}} & 8.8589 & \textbf{\underline{0.9695}} & N/A & 0.7462 \\
CatBoost & 0.4317 & 0.8722 & 0.3846 & 0.7231 & 0.7243 & OOM & 0.8881 & 8.8765 & 0.9658 & 0.7429 & 0.7429 \\
XGBoost & 0.4338 & 0.8722 & 0.3749 & 0.7218 & 0.7264 & OOM & 0.8835 & 8.953 & 0.9689 & 0.7357 & 0.7424 \\
\bottomrule
\end{tabular}
\end{threeparttable}
\end{table}

\subsection{On Datasets used in GRANDE paper}
\label{sec:eval-grande-result}

The datasets used in GRANDE paper are from the OpenML-CC18 benchmark \citep{bischl2017openml}, which are different from both Tabular Benchmark and datasets used in FT-Transformer paper. Hence, we decide to evaluate DOFEN on these datasets for a more comprehensive comparison. It is worth noting that, these datasets only cover binary classification tasks and include many small datasets (data size < 1000) according to the definition of Tabular Benchmark. For experimental settings, we strictly follow the settings mentioned in GRANDE paper. 

The experiment result is provided in \cref{tab:performance-grande}. The performance of GRANDE, XGBoost, CatBoost, and NODE for both searched and default hyperparameters are taken from the GRANDE paper, while we only report the performance of default hyperparameters for DOFEN, due to the lack of computational resources for a complete hyperparameter search. Additionally, DOFEN has made some minor adjustments to accommodate the wide range of dataset size of this benchmark, these adjustments are applied evenly on all datasets, with the details provided as follows:

\begin{enumerate}
    \item DOFEN originally set batch size as 256. We choose to use an adaptive batch size for each dataset, which is set to $\min\{256, 2^{\floor{\log_2(\frac{N}{10})}}\}$, where $N$ represents dataset size. This adjustment is made mainly because using too large batch size will have negative effect when training DOFEN on small datasets.
    \item We set $N_\text{head}$ to 4 for DOFEN. This is because there exist several large-sized datasets in this benchmark, this adjustment is made to better accommodate with large-sized datasets. This adjustment has also been made in \cref{sec:eval-ft-result} for the same reason.
\end{enumerate}

Based on the experiment result, the average ranking of DOFEN using the default hyperparameter ranks 3rd, only ranks behind GRANDE and CatBoost using searched hyperparameters, and exceeds XGBoost using searched hyperparameters. 
Moreover, we found that the performance of DOFEN is competitive with GRANDE and tree-based models on medium-to-large datasets, while sometimes falling behind other models on medium-to-small datasets. 
These findings are similar to those found on Tabular Benchmark and datasets used in FT-Transformer paper, where DOFEN's performance benefits from a larger dataset size. 
For the slightly worse performance on medium-to-small datasets, this can be explained that DOFEN tends to overfit on smaller datasets, hence including hyperparameter search with regularization techniques (e.g. dropout rate) or lower the capacity of a model (e.g. try smaller $m$ and $d$ in DOFEN) will help to improve the performance.

\begin{table}[H]
\centering
\small
\setlength\tabcolsep{2.25pt}
\caption{The performance of datasets used in GRANDE paper. We report the test macro F1-score (mean for a 5-fold
CV) with default (subscript d) and searched parameters (subscript s), and an average performance ranking across datasets is calculated for an overall comparison. The best performance is highlighted in \textbf{\underline{bold with underline}}, while the second best performance is highlighted in \textbf{bold}. The datasets are sorted based on the data size, here we only show the names in the first three letters for a more compact table. The letter in the bracket represents the size of the dataset, where "L" indicates large datasets (data size > 10000), "M" indicates medium datasets (10000 > data size > 1000), and "S" indicates small datasets (data size < 1000), according to the definition of Tabular Benchmark.}
\begin{threeparttable}
\label{tab:performance-grande}

\begin{tabular}{l c c c c c c c c c}
\toprule
 & \begin{tabular}[c]{@{}c@{}}$\text{DOFEN}_d$\\(ours)\end{tabular} & $\text{GRANDE}_s$ & $\text{XGB}_s$ & $\text{CatBoost}_s$ & $\text{NODE}_s$ & $\text{GRANDE}_d$ & $\text{XGB}_d$ & $\text{CatBoost}_d$ & $\text{NODE}_d$ \\
\midrule
NUM (L) & \textbf{\underline{0.520}} & \textbf{0.519} & 0.518 & \textbf{0.519} & 0.503 & 0.503 & 0.516 & \textbf{0.519} & 0.506 \\
ADU (L) & \textbf{\underline{0.801}} & 0.790 & 0.798 & 0.796 & 0.794 & 0.785 & 0.796 & 0.796 & \textbf{0.799} \\
NOM (L) & 0.956 & 0.958 & \textbf{\underline{0.965}} & \textbf{0.964} & 0.956 & 0.955 & \textbf{\underline{0.965}} & 0.962 & 0.955 \\
AMA (L) & \textbf{\underline{0.674}} & 0.665 & 0.621 & \textbf{0.671} & 0.649 & 0.602 & 0.608 & 0.652 & 0.621 \\
PHI (L) & 0.953 & \textbf{\underline{0.969}} & \textbf{0.968} & 0.965 & \textbf{0.968} & \textbf{\underline{0.969}} & \textbf{\underline{0.969}} & 0.963 & 0.961 \\
SPE (M) & \textbf{0.723} & \textbf{0.723} & 0.704 & 0.718 & 0.707 & \textbf{\underline{0.725}} & 0.686 & 0.693 & 0.703 \\
PHO (M) & 0.847 & 0.846 & \textbf{0.872} & \textbf{\underline{0.876}} & 0.862 & 0.86 & 0.864 & 0.861 & 0.842 \\
CHU (M) & \textbf{0.926} & 0.914 & 0.900 & 0.869 & \textbf{\underline{0.930}} & 0.896 & 0.897 & 0.862 & 0.925 \\
WIL (M) & \textbf{0.939} & 0.936 & 0.911 & 0.919 & 0.937 & 0.933 & 0.897 & \textbf{\underline{0.962}} & 0.925 \\
BIO (M) & 0.773 & 0.794 & \textbf{0.799} & \textbf{\underline{0.801}} & 0.780 & 0.789 & 0.789 & 0.792 & 0.786 \\
MAD (M) & 0.580 & 0.803 & 0.833 & \textbf{\underline{0.861}} & 0.571 & 0.768 & 0.811 & \textbf{0.851} & 0.650 \\
OZO (M) & 0.721 & \textbf{0.726} & 0.688 & 0.703 & 0.721 & \textbf{\underline{0.735}} & 0.686 & 0.702 & 0.662 \\
QSA (M) & 0.847 & \textbf{\underline{0.854}} & \textbf{0.853} & 0.844 & 0.836 & 0.851 & 0.844 & 0.843 & 0.838 \\
TOK (S) & \textbf{0.922} & 0.921 & 0.915 & \textbf{\underline{0.927}} & 0.921 & \textbf{0.922} & 0.917 & 0.917 & 0.921 \\
ILP (S) & \textbf{0.647} & \textbf{\underline{0.657}} & 0.632 & 0.643 & 0.526 & 0.646 & 0.629 & 0.643 & 0.501 \\
WDB (S) & \textbf{0.970} & \textbf{\underline{0.975}} & 0.953 & 0.963 & 0.966 & 0.962 & 0.966 & 0.955 & 0.964 \\
CYL (S) & 0.758 & \textbf{\underline{0.819}} & 0.773 & 0.801 & 0.754 & \textbf{0.813} & 0.770 & 0.795 & 0.696 \\
CLI (S) & 0.712 & \underline{0.853} & 0.763 & 0.778 & \textbf{0.802} & 0.758 & 0.781 & 0.781 & 0.766 \\
DRE (S) & 0.557 & \textbf{\underline{0.612}} & 0.581 & 0.588 & 0.564 & \textbf{0.596} & 0.570 & 0.573 & 0.559 \\
\midrule
Avg. Rank & 4.53 & \textbf{\underline{3.00}} & 4.95 & \textbf{3.74} & 5.47 & 4.89 & 5.32 & 4.95 & 6.74 \\
\midrule
Avg. Rank (L) & \textbf{3.60} & 3.80 & 3.80 & \textbf{\underline{3.40}} & 6.00 & 7.00 & 4.00 & 4.20 & 6.20 \\
Avg. Rank (M) & 4.63 & \textbf{\underline{3.63}} & 4.38 & \textbf{4.00} & 5.25 & 4.25 & 6.13 & 5.25 & 7.00 \\
Avg. Rank (S) & 5.17 & \textbf{\underline{1.50}} & 6.67 & \textbf{3.67} & 5.33 & 4.00 & 5.33 & 5.17 & 6.83 \\
\bottomrule
\end{tabular}
\end{threeparttable}
\end{table}
\section{More Experiment Settings}

\subsection{Hardware Used}
\label{sec:hardware-used}
The following hardware configuration was used for all of our experiments. The hardware selection was based on availability, with neural networks consistently run on GPUs and tree-based models executed on CPUs.

GPUs: NVIDIA GeForce RTX 2080 Ti, NVIDIA DGX1, NVIDIA A100

CPUs:  Intel(R) Xeon(R) Silver 4210 CPU, Intel(R) Xeon(R) CPU E5-2698 v4, AMD EPYC605 7742 64-core Processor

\subsection{Hyperparameter Search Space}
\label{sec:search-space}

This section details the hyperparameter search space adopted for each model, as referenced in various tables (\cref{tab:search-DOFEN,tab:search-xgboost,tab:search-catboost,tab:search-lightgbm,tab:search-gbt,tab:search-forest,tab:search-node,tab:search-trompt,tab:search-ft-transformer,tab:search-saint,tab:search-resnet,tab:search-mlp}).
We have employed search spaces consistent with those presented in the Tabular Benchmark \cite{grinsztajn2022tree} for models including XGBoost, GradientBoostingTree, RandomForest, FT-Transformer, SAINT, ResNet, and MLP.

Additionally, we have defined specific search spaces for newer baselines such as CatBoost, LightGBM, Trompt, NODE, and GRANDE.
For CatBoost, our search space aligns with the parameters specified by the FT-Transformer study \cite{gorishniy2021revisiting}.
In the case of LightGBM, we have derived the search space based on recommendations from field practitioners, as cited in \cite{kaggle2019lightgbm, neptune.ai2022lightgbm}.
For NODE, our approach follows the guidelines provided in TabZilla \cite{mcelfresh2023neural}.
For GRANDE, we follow the settings provided in the notebook example from the official github of GRANDE.

In the context of our model, DOFEN, we have focused our search on the number of $m$ and $d$, which relate to the varied number of $N_\text{cond}$ and the conditions per rODT.
Additionally, $N_\text{head}$ is another important parameter since it increase the capacity for the model to evaluate how well a sample aligns with the conditions of an rODT.
Lastly, we have explored the \textit{drop\_rate} parameter to fine-tune the degree of regularization in our model. 
It is important to note that the overall search space for DOFEN is relatively compact when compared to the other baseline models while achieving competitive performance.

\begin{table}[H]
\centering
\small
\caption{Hyperparameter search space of DOFEN.}
\begin{tabular}{c c}
\toprule
\textbf{Hyperparameter} & \textbf{Distribution}     \\ \midrule
$d$                       & $[3,4,6,8]$                          \\ 
$m$                       & $[16,32,64]$                          \\
$N_\text{head}$                      &                       $[1,4,8]$\\
\textit{drop\_rate}               & $[0.0,0.1,0.2]$         \\
\bottomrule
\end{tabular}
\label{tab:search-DOFEN}
\end{table}

\begin{table}[H]
\centering
\small
\caption{Hyperparameter search space of XGBoost.}
\begin{tabular}{c c}
\toprule
\textbf{Hyperparameter} & \textbf{Distribution} \\ \midrule
max\_depth & $\text{uniform\_int}[1,11]$\\
num\_estimators & $1000$\\
min\_child\_weight & $\text{log\_uniform\_int}[1,1\mathrm{e}2]$\\
subsample & $\text{unifrom}[0.5,1]$\\
learning\_rate & $\text{log\_unifrom}[1\mathrm{e}{-5},0.7]$\\
col\_sample\_by\_level & $\text{uniform}[0.5,1]$\\
col\_sample\_by\_tree & $\text{uniform}[0.5,1]$\\
gamma & $\text{log\_uniform}[1\mathrm{e}{-8},7]$\\
lambda & $\text{log\_uniform}[1,4]$\\
alpha & $\text{log\_uniform}[1\mathrm{e}{-8},1\mathrm{e}2]$\\
\bottomrule
\end{tabular}
\label{tab:search-xgboost}
\end{table}

\begin{table}[H]
\centering
\small
\caption{Hyperparameter search space of CatBoost.}
\begin{tabular}{c c}
\toprule
\textbf{Hyperparameter} & \textbf{Distribution} \\ \midrule
max\_depth & $[3,4,5,6,7,8,9,10]$\\
learning\_rate & $\text{log\_uniform}[1\mathrm{e}{-5},1]$\\
iterations & $\text{quantile\_uniform}[100,6000]$\\
bagging\_temperature & $\text{uniform}[0,1]$\\
l2\_leaf\_reg & $\text{log\_uniform}[1,10]$\\
leaf\_estimation\_iteration & $[1,2,3,4,5,6,7,8,9,10]$\\
\bottomrule
\end{tabular}
\label{tab:search-catboost}
\end{table}

\begin{table}[H]
\centering
\small
\caption{Hyperparameter search space of LightGBM.}
\begin{tabular}{c c}
\toprule
\textbf{Hyperparameter} & \textbf{Distribution} \\ \midrule
learning\_rate & $\text{uniform}[0.001,1]$\\
max\_depth & $[1,2,3,4,5,6,7,8,9,10,11]$\\
bagging\_fraction & $\text{uniform}[0.1,1.0]$\\
bagging\_frequency & $[1,2,3,4,5]$\\
num\_leaves & $\text{quantile\_uniform}[30,150]$\\
feature\_fraction & $\text{uniform}[0.1,1.0]$\\
num\_estimators & $1000$\\
boosting & $[\text{gbdt},\text{rf},\text{dart}]$\\
\bottomrule
\end{tabular}
\label{tab:search-lightgbm}
\end{table}

\begin{table}[H]
\centering
\small
\caption{Hyperparameter space of GradientBoostingTree.}
\begin{tabular}{c c}
\toprule
\textbf{Hyperparameter} & \textbf{Distribution} \\ \midrule
loss & 

\begin{tabular}[c]{@{}c@{}}$[\text{deviance},\text{exponential}](classification),$\\$[\text{squared\_error},\text{absolute\_error},\text{huber}](regression)$\end{tabular}\\
learning\_rate & $\text{log\_normal}[\text{log}(0.01),\text{log}(10)]$\\
subsample & $\text{uniform}[0.5,1]$\\
num\_estimators & $1000$\\
criterion & $[\text{friedman\_mse},\text{squared\_error}]$\\
max\_depth & $[\text{none},2,3,4,5]$\\
min\_samples\_split & $[2.3]$\\
min\_impurity\_decrease & $[0.0,0.01,0.02,0.05]$\\
max\_leaf\_nodes & $[\text{none},5,10,15]$\\
\bottomrule
\end{tabular}
\label{tab:search-gbt}
\end{table}

\begin{table}[H]
\centering
\small
\caption{Hyperparameter search space of RandomForest.}
\begin{tabular}{c c}
\toprule
\textbf{Hyperparameter} & \textbf{Distribution}\\
\midrule
max\_depth & $[\text{none},2,3,4]$\\
num\_estimators & $250$\\
criterion & $[\text{gini},\text{entropy}]$\\
max\_features & $[\text{sqrt},\text{log2},\text{none},0.1,0.2,0.3,0.4,0.5,0.6,0.7,0.8,0.9]$\\
min\_samples\_split & $[2,3]$\\
min\_samples\_leaf & $\text{log\_uniform\_int}[1.5,50.5]$\\
bootstrap & $[\text{true},\text{false}]$\\
min\_impurity\_decrease & $[0.0,0.01,0.02,0.05]$\\
\bottomrule
\end{tabular}
\label{tab:search-forest}
\end{table}

\begin{table}[H]
\centering
\small
\caption{Hyperparameter search space of NODE.}
\begin{tabular}{c c}
\toprule
\textbf{Hyperparameter} & \textbf{Distribution} \\ \midrule
num\_layers & $[2,4,8]$\\
total\_tree\_count & $[1024,2048]$\\
tree\_depth & $[6,8]$\\
tree\_output\_dimension & $[2,3](regression),[\text{num\_classes}](classification)$\\
\bottomrule
\end{tabular}
\label{tab:search-node}
\end{table}

\begin{table}[H]
\centering
\small
\caption{Hyperparameter search space of Trompt.} 
\begin{tabular}{c c}
\toprule
\textbf{Hyperparameter} & \textbf{Distribution} \\ \midrule
hidden\_dimension & $[18,128]$\\
feature\_importances\_type & $[\text{concat},\text{add}]$\\
feature\_importances\_dense & $[\text{true},\text{false}]$\\
feature\_importances\_residual\_connection & $[\text{true},\text{false}]$\\
feature\_importances\_sharing\_dense & $[\text{true},\text{false}]$\\
feature\_embeddings\_residual\_connection & $[\text{true},\text{false}]$\\
minimal\_batch\_ratio & $[0.1,0.01]$\\
\bottomrule
\end{tabular}
\label{tab:search-trompt}
\end{table}

\begin{table}[H]
\centering
\small
\caption{Hyperparameter search space of FT-Transformer.}
\begin{tabular}{c c}
\toprule
\textbf{Hyperparameter} & \textbf{Distribution} \\ \midrule
mum\_layers & $\text{uniform\_int}[1,6]$\\
feature\_embedding\_size & $\text{uniform\_int}[64,512]$\\
residual\_dropout  & $\text{uniform}[0,0.5]$\\
attention\_dropout & $\text{uniform}[0,0.5]$\\
FFN\_dropout & $\text{uniform}[0,0.5]$\\
FFN\_factor & $\text{uniform}[2/3,8/3]$\\
learning\_rate & $\text{log\_uniform}[1\mathrm{e}{-5},1\mathrm{e}{-3}]$\\
weight\_decay & $\text{log\_uniform}[1\mathrm{e}-6,1\mathrm{e}-3]$\\
KV\_compression & $[\text{true},\text{false}]$\\
LKV\_compression\_sharing & $[\text{headwise},\text{key\_value}]$\\
learning\_rate\_scheduler & $[\text{true},\text{false}]$\\
batch\_size & $[256, 512, 1024]$\\
\bottomrule
\end{tabular}
\label{tab:search-ft-transformer}
\end{table}

\begin{table}[H]
\centering
\small
\caption{Hyperparameter search space of SAINT.}
\begin{tabular}{c c}
\toprule
\textbf{Hyperparameter} & \textbf{Distribution}\\
\midrule
num\_layers & $\text{uniform\_int}[1,2,3,6,12]$\\
num\_heads & $[2,4,8]$\\
layer\_size & $\text{uniform\_int}[32,64,128]$\\
dropout & $[0,0.1,0.2,0.3,0.4,0.5,0.6,0.7,0.8]$\\
learning\_rate & $\text{log\_uniform}[1\mathrm{e}{-5},1\mathrm{e}{-3}]$\\
batch\_size & $[128,256]$\\
\bottomrule
\end{tabular}
\label{tab:search-saint}
\end{table}

\begin{table}[H]
\centering
\small
\caption{Hyperparameter search space of ResNet.}
\begin{tabular}{c c}
\toprule
\textbf{Hyperparameter} & \textbf{Distribution} \\ \midrule
num\_layers & $\text{uniform\_int}[1,16]$\\
layer\_size & $\text{uniform\_int}[64,1024]$\\
hidden\_factor & $\text{uniform}[1,4]$\\
hidden\_dropout & $[0,0.5]$\\
residual\_dropout & $\text{uniform}[0,0.5]$\\
learning\_rate & $\text{log\_uniform}[1\mathrm{e}{-5},1\mathrm{e}{-2}]$\\
weight\_decay & $\text{log\_uniform}[1\mathrm{e}{-8},1\mathrm{e}{-3}]$\\
category\_embedding\_size & $\text{uniform\_int}[64,512]$\\
normalization & $[\text{batch\_norm},\text{layer\_norm}]$\\
learning\_rate\_scheduler & $[\text{true},\text{false}]$\\
batch\_size & $[256,512,1024]$\\
\bottomrule
\end{tabular}
\label{tab:search-resnet}
\end{table}

\begin{table}[H]
\centering
\small
\caption{Hyperparameter search space of MLP.}
\begin{tabular}{c c}
\toprule
\textbf{Hyperparameter} & \textbf{Distribution} \\ \midrule
num\_layers & $\text{uniform\_int}[1,8]$\\
layer\_size & $\text{uniform\_int}[16,1024]$\\
dropout & $[0,0.5]$\\
learning\_rate & $\text{log\_uniform}[1\mathrm{e}{-5},1\mathrm{e}{-2}]$\\
category\_embedding\_size & $\text{uniform\_int}[64,512]$\\
learning\_rate\_scheduler & $[\text{true},\text{false}]$\\
batch\_size & $[256,512,1024]$\\
\bottomrule
\end{tabular}
\label{tab:search-mlp}
\end{table}

\begin{table}
\centering
\small
\caption{Hyperparameter search space of GRANDE.}
\begin{tabular}{c c}
\toprule
\textbf{Hyperparameter} & \textbf{Distribution} \\ \midrule
depth & $[4, 6]$\\
n\_estimators & $[512, 1024, 2048]$\\
learning\_rate\_weights & $\text{log\_uniform}[1\mathrm{e}{-4},1\mathrm{e}{-1}]$\\
learning\_rate\_index & $\text{log\_uniform}[5\mathrm{e}{-3},2\mathrm{e}{-1}]$\\
learning\_rate\_values & $\text{log\_uniform}[5\mathrm{e}{-3},2\mathrm{e}{-1}]$\\
learning\_rate\_leaf & $\text{log\_uniform}[5\mathrm{e}{-3},2\mathrm{e}{-1}]$\\
cosine\_decay\_steps & $[0, 100, 1000]$\\
loss & $[\text{crossentropy},\text{focal\_crossentropy}](classification),[\text{mse}](regression)$\\
dropout & $[0.0, 0.25, 0.5]$ \\
selected\_variables & $\text{uniform}[0.5, 1.0]$\\
data\_subset\_fraction & $\text{uniform}[0.8, 1.0]$\\
\bottomrule
\end{tabular}
\label{tab:search-grande}
\end{table}


\end{document}